%% file: main.tex
\documentclass[nonblindrev]{informs3noheader}

\OneAndAHalfSpacedXI

\usepackage{mathtools}
\usepackage[inline]{enumitem}
\usepackage{subfig}
\usepackage{booktabs}
\usepackage{array}
\usepackage{multirow}
\usepackage{booktabs,caption} %,fixltx2e}
\usepackage[flushleft]{threeparttable}
\usepackage{lscape}
\usepackage{makecell}
\usepackage{amssymb}
\usepackage[mathscr]{euscript}
\usepackage{algorithm}
\usepackage{algpseudocode}
\usepackage{pifont}
\usepackage{caption}
\usepackage{algorithmicx}
\usepackage{hyperref}
\usepackage{mathrsfs}
\usepackage{tabularx}
\usepackage{bbm}
\usepackage{soul}
\usepackage{blindtext}
\usepackage{anyfontsize}
\usepackage{longtable}
\hypersetup{
colorlinks=true,
linkcolor=black,
citecolor=black,
filecolor=black,
urlcolor=black,
}

\usepackage{comment}

\usepackage{tikz}
\usetikzlibrary{shapes.geometric}
\usetikzlibrary{positioning}
\usetikzlibrary{decorations.pathreplacing}
\definecolor{mygreen}    {RGB}{0,90,0}
\definecolor{myblue}     {RGB}{0,51,140}
\definecolor{myorange}   {RGB}{238,118,0}
\definecolor{myred}      {RGB}{126,0,0}
\definecolor{mygray}     {RGB}{100,100,105}
\definecolor{mygrayblue} {RGB}{0,128,128}
\definecolor{mygraygreen}{RGB}{128,128,0}
\definecolor{DarkPurple}     {RGB}{142, 36, 170}
\definecolor{LightPurple}    {RGB}{57, 130, 7}

\definecolor{mylightblue}{RGB}{183,200,242}
\definecolor{mymidblue}{RGB}{103,141,242}
\definecolor{mydarkblue}{RGB}{13,43,117}
\definecolor{mylightorange}{RGB}{242,182,145}
\definecolor{mymidorange}{RGB}{241,100,13}
\definecolor{mydarkorange}{RGB}{143,42,15}
\definecolor{myyellow}{RGB}{255,176,0}
\definecolor{mypink}{RGB}{220,38,127}
\definecolor{mypurple}{RGB}{120,94,240}
\definecolor{mainblue}{RGB}{0,114,178}
\definecolor{mainorange}{RGB}{213,94,0}

\usepackage{natbib}
\bibpunct[, ]{(}{)}{,}{a}{}{,}%
\def\bibfont{\small}%

\usepackage[utf8]{inputenc}
\usepackage[english]{babel}
\usepackage{graphicx}

\newcolumntype{T}[1]{%
    >{\centering\arraybackslash\hspace{0pt}}p{#1}}%

\TheoremsNumberedThrough

\EquationsNumberedThrough
{\theoremstyle{EXkey}}

\MANUSCRIPTNO{}

\input{abbrev}

\begin{document}

\ARTICLEAUTHORS{
	\AUTHOR{Cynthia Barnhart, Alexandre Jacquillat, Alexandria Schmid}
	\AFF{Sloan School of Management and Operations Research Center, MIT, Cambridge, MA}
}	
\RUNAUTHOR{Barnhart, Jacquillat, Schmid}

\RUNTITLE{Robotic warehousing operations}

\TITLE{\Large Robotic warehousing operations: a learn-then-optimize approach to large-scale neighborhood search}

\ABSTRACT{The rapid deployment of robotics technologies requires dedicated optimization algorithms to manage large fleets of autonomous agents. This paper supports robotic parts-to-picker operations in warehousing by optimizing order-workstation assignments, item-pod assignments and the schedule of order fulfillment at workstations. The model maximizes throughput, while managing human workload at the workstations and congestion in the facility. We solve it via large-scale neighborhood search, with a novel learn-then-optimize approach to subproblem generation. The algorithm relies on an offline machine learning procedure to predict objective improvements based on subproblem features, and an online optimization model to generate a new subproblem at each iteration. In collaboration with Amazon Robotics, we show that our model and algorithm generate much stronger solutions for practical problems than state-of-the-art approaches. In particular, our solution enhances the utilization of robotic fleets by coordinating robotic tasks for human operators to pick multiple items at once, and by coordinating robotic routes to avoid congestion in the facility.}


\maketitle

\vspace{-18pt}

\section{Introduction}

\begin{figure} %%
  \centering
  \subfloat[Bird's-eye view of a fulfillment center]{\includegraphics[width=0.45\textwidth]{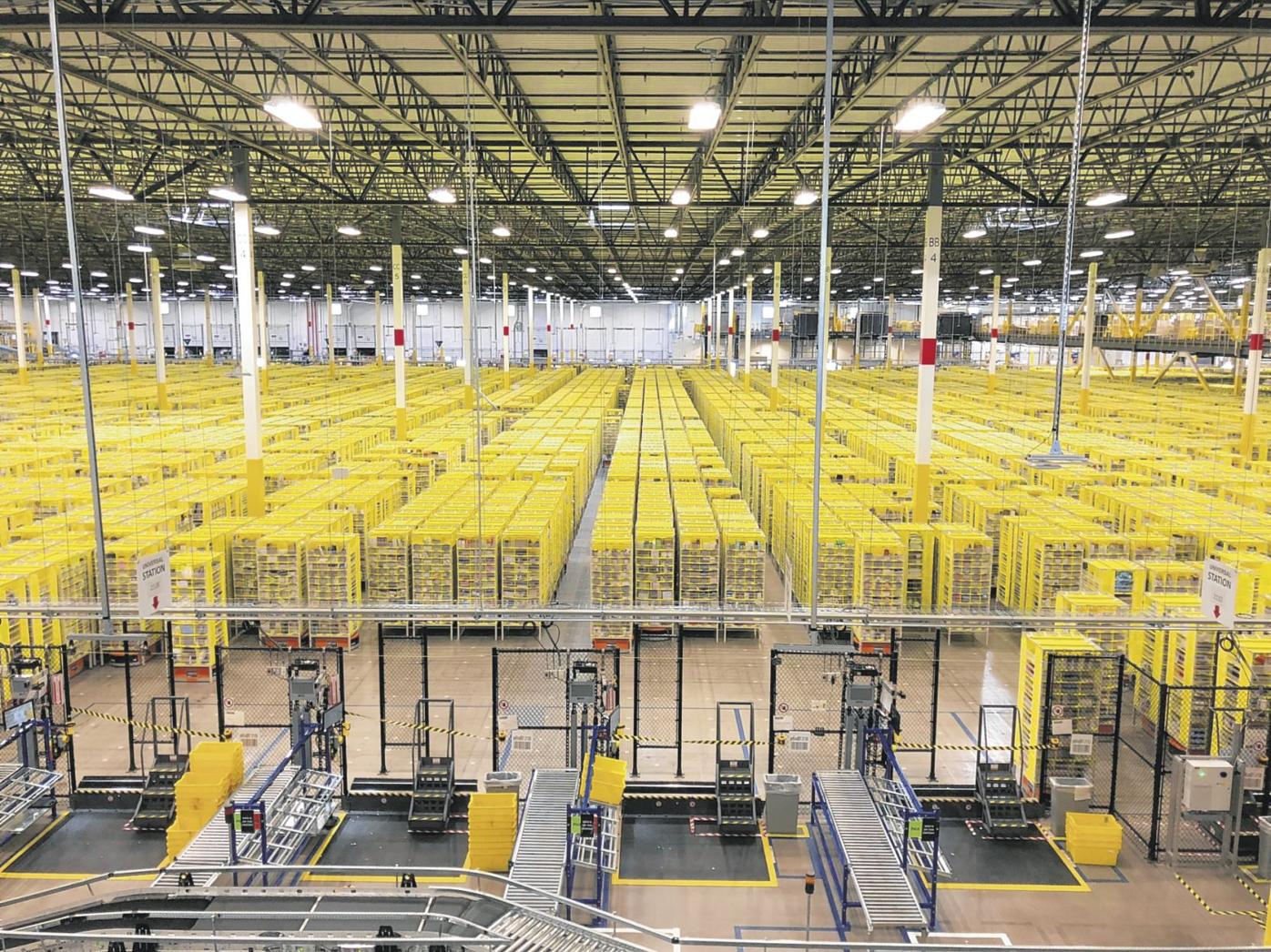}\label{fig:warehouse}}
  \hfill
  \subfloat[Robot-human interactions]{\includegraphics[width=0.45\textwidth]{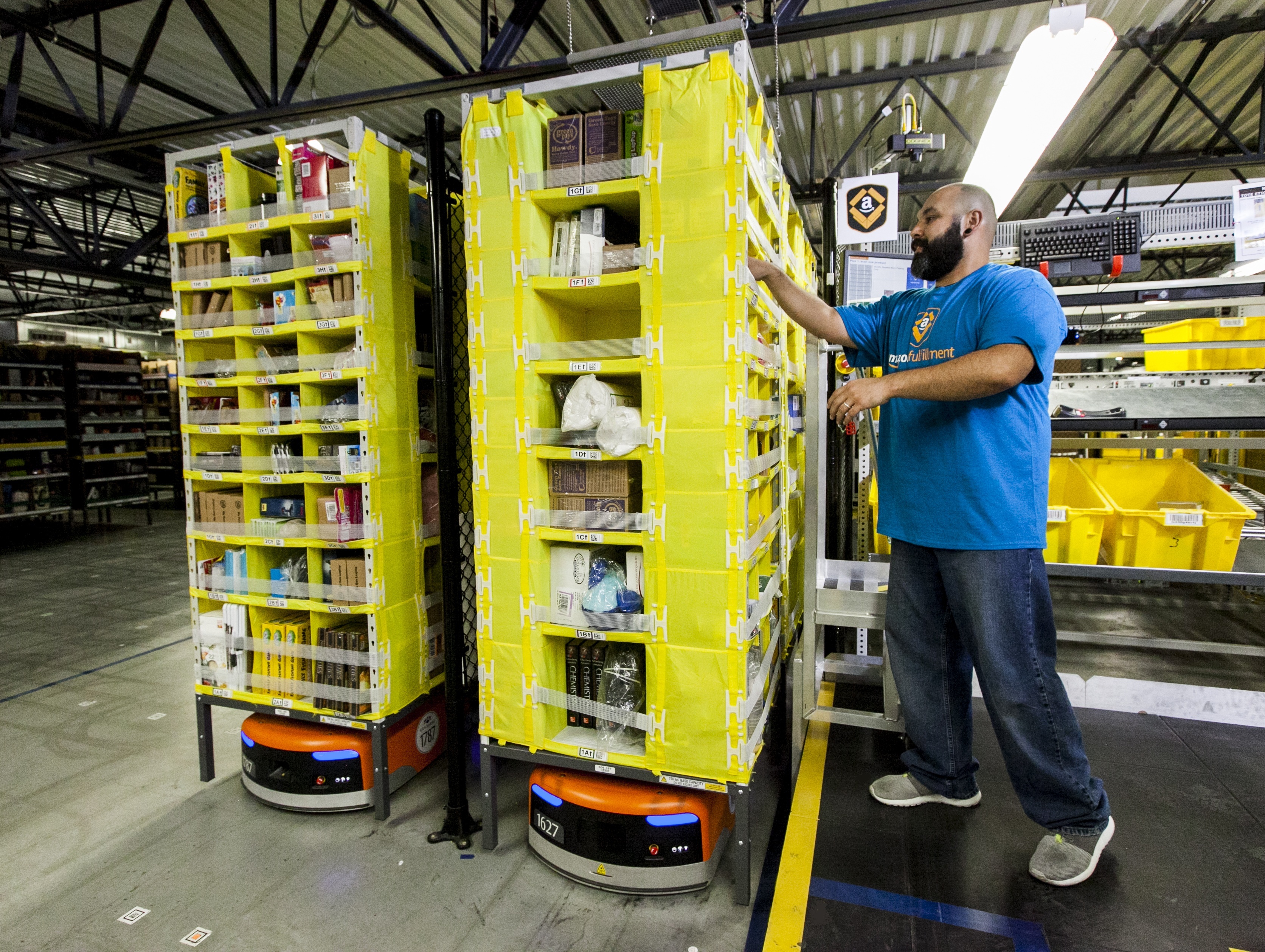}\label{fig:picker}}
  \caption{(a) Full-scale fulfillment center, with workstations in the foreground and inventory shelves stored behind.\\(b) Robotic agents bring shelves to the workstations in part-to-picker operations.}
  \label{fig:amazon}
  \vspace{-12pt}
\end{figure}

Fueled by advances in artificial intelligence, robotic process automation is impacting virtually every sector of the economy \citep{mck2017future}. The logistics sector lies at the core of this transformation: autonomous mobile robots are being deployed in tens of thousands of manufacturing and distribution facilities with a near-term \$10-50 billion market potential \citep{GVR,ABI}. A predominant operating model, shown in Figure~\ref{fig:amazon}, involves \textit{part-to-picker} warehousing operations, which relies on robotic agents transporting shelves of inventory from a storage location to a workstation for a human operator to fulfill orders and back to a storage location. Robotic operations can improve throughput and working conditions by letting human workers focus on the more productive tasks, while improving system reliability.

Yet, to truly take advantage of automation opportunities, modern warehousing systems require dedicated decision support tools to manage large robotic fleets and human-robot interactions in high-density operations. At the core of robotic process automation lies the computer vision, sensing, mapping and robotic technologies to empower autonomous agents---in our case, robots capable to move shelves of inventory. A subsequent problem involves control mechanisms to coordinate multi-agent systems---in our case, to avoid conflicts and collisions between robots. But then, a third layer involves determining task design and assignment---in our case, which robot should move which shelf from where to where at what time. This last layer is becoming increasingly prevalent in high-density warehousing environments, where the joint actions of hundreds to thousands of robotic agents may result in unnecessary congestion, excessive human workload, and reduced throughput. In response to these challenges, the core robotic algorithms need to be augmented with large-scale optimization algorithms to coordinate large-scale fleets in robotic warehousing.

As part of this agenda, we partner with Amazon Robotics to optimize part-to-picker operations in automated fulfillment centers. Our first contribution is formulating a \textit{task design and scheduling problem with congestion and restricted workload (TDS-CW)}. The TDS-CW considers customer orders, each comprising one or several items. To create operating flexibility, each item is stored in several shelves (or ``pods'') and each order can be processed at any workstation---yet, all items in an order need to be delivered to the same workstation. The facility has access to a large robotic fleet to move pods between storage locations and workstations. The main bottlenecks lie in human workload restrictions and in congestion in the facility. The TDS-CW optimizes order-workstation assignments, item-pod assignments and the schedule of order fulfillments at workstations to maximize throughput (a proxy for revenue and delivery times) while restricting pod flows (a proxy for congestion) and the number of open orders at a workstation (a proxy for human workload).

The TDS-CW exhibits several features that separate it from traditional pickup-and-delivery problems. One is the dichotomy of items and pods: the origin of each movement is subject to optimization because each item can be found on multiple pods \citep{weidinger2018scattered}. Another one is the dichotomy of orders and workstations: the destination of each movement is also subject to optimization because each order can be processed at any workstation. Yet another one is the dichotomy of demand and supply: each order can comprise multiple items that need to be routed to the same workstation. These degrees of freedom create strong spatial-temporal coordination requirements across the facility. Accordingly, we propose an integer optimization formulation of the TDS-CW using a time-space network representation of fulfillment operations.

At the same time, the TDS-CW model is highly challenging to solve due to the multi-dimensional assignment variables required to capture coordination requirements. Full-scale instances involve tens of millions of binary variables and over a hundred million constraints. This sheer size severely hinders tractability---off-the-shelf methods cannot even scale to small-scale examples. We therefore solve it via large-scale neighborhood search (LSNS), using a ``destroy and repair'' approach to iteratively re-optimize the solution within a large neighborhood while maintaining global feasibility. The success of this approach relies on its ability to find neighborhoods where the solution can be improved locally---in our case, neighborhoods with ``synergistic'' orders, items, pods, workstations and time blocks to facilitate assignments . This paper proposes a new method to guide subproblem generation in the LSNS algorithm, using machine learning and optimization.

Specifically, our second contribution is developing a learn-then-optimize algorithm to generate LSNS subproblems at each iteration, by combining supervised learning and optimization. Several learning-enhanced approaches to LSNS \emph{select} a neighborhood with the highest predicted throughput, among pre-defined candidate neighborhoods. Our setting, however, involves an exponential number of candidate neighborhoods, so baseline approaches can lead to poor-quality solutions with few candidate neighborhoods, and to extensive computational requirements. Instead, our learn-then-optimize approach \emph{generates} a new neighborhood at each iteration, using offline machine learning and online optimization. Our algorithm first trains a linear machine learning model---using the holistic regression approach from \cite{bertsimas2016or}---to predict the local objective as a function of features of low-dimensional components of the neighborhood (e.g., characteristics of orders, pods, workstations, time periods, and their pair-wise interactions, in our case). Our algorithm then uses integer optimization to construct a new neighborhood at each LSNS iteration, in order to maximize the expected solution improvement subject to neighborhood constraints (e.g., continuity in time and space) and tractability constraints (e.g., size of the subproblem). In other words, the proposed approach attributes local performance to low-dimensional neighborhood components to generate new neighborhoods iteratively. In particular, the key component of the learn-then-optimize LSNS algorithm is that it can visit new neighborhoods in the online optimization phase, beyond those that were visited in the offline training phase.

Our third contribution is demonstrating, via extensive computational experiments, the practical benefits of our modeling and algorithmic approach. Notably, our learn-then-optimize LSNS algorithm scales to medium-sized, yet otherwise-intractable TDS-CW instances, with 4-14\% throughput improvements over LSNS benchmarks (namely, domain-based heuristics, random sampling, and baseline learning-enhanced approaches). In particular, these results underscore the impact of generating a new ``synergistic'' neighborhood at each iteration (via our learn-then-optimize algorithm), as opposed to restricting the algorithm to a subset of pre-defined neighborhoods (via baseline learning-enhanced algorithms). We embed the algorithm into a partitioning scheme that uses the learn-then-optimize outputs to decompose large-scale instances into separate regions, which can then be solved in parallel. Results suggest that the optimization methodology developed in this paper can generate high-quality solutions in full-sized instances representative of warehouses encountered in practice, in reasonable computational times consistent with practical requirements. These performance improvements are driven by two main mechanisms. First, our solution coordinates operations to consolidate order fulfillments by picking multiple items from each pod, thus increasing utilization and reducing pod movements. Second, our solution avoids overcrowding some areas of the warehouse, resulting in lower delays and higher utilization---with resulting throughput increases of over 10\% as compared to a congestion-blind benchmark. Ultimately, the methodology developed in this paper contributes new decision tools to enhance fleet management and support high-density operations in emerging robotic warehousing environments.

\section{Literature review}

\subsubsection*{Multi-agent robotic coordination.} The Multi-Agent Path Finding (MAPF) problem optimizes the movements of several robotic agents, each traveling from a fixed origin to a fixed destination, to maximize throughput and avoid conflicts. Applications include search-and-rescue operations \citep{jennings1997cooperative}, air traffic control \citep{vsivslak2010agent}, and robotic warehousing \citep{wurman2008coordinating}. In the Lifelong Multi-Agent Path Finding extension, each agent needs to perform a sequence of tasks, which is typically solved via temporal decomposition in a rolling horizon \citep{grenouilleau2019multi,li2021lifelong}. The related Multi-Agent Pickup and Delivery (MAPD) problem also optimizes the assignment of agents to pre-defined tasks, each with a fixed origin and a fixed destination \citep{ma2017lifelong,liu2019task,chen2021integrated}. The MAPF and MAPD problems are known to be highly challenging, and typically solved via heuristics \citep{salzman2020research}. In particular, \cite{gharehgozli2020robot} formulate the problem as a generalized asymmetric traveling salesman problem and solve it via adaptive large neighborhood search. Importantly, these problems treat the set of tasks as given, whereas the TDS-CW \textit{designs} and \textit{schedules} robotic tasks by optimizing the origin (pod) and destination (workstation) of each movement. This flexibility adds complexity to the already-complicated MAPF and MAPD problems. To retain tractability, we consider a lower level of granularity by focusing on fleet management operations without controlling the microscopic movements of robotic agents within the facility.

This distinction connects our paper to the operations research literature on fulfillment center operations \citep{boysen2019warehousing,azadeh2019robotized}. Sample problems include zone picking design \citep{van2020capacity}, inventory storage \citep{ang2012robust,weidinger2018scattered}, inventory restocking \citep{cezik2022velocity}, picking operations \citep{zhang2009modeling,goeke2021modeling}, etc. Several studies rely on sequential heuristics to decompose the problem into smaller components. \cite{boysen2017parts} optimize item processing at a single workstation by separating order sequencing and pod sequencing; similarly, \cite{valle2021order} decompose a multi-workstation problem into an assignment module and a sequencing module. \cite{wang2022robot} solve a problem with fluctuations in processing rates at the workstations via approximate dynamic programming and branch-and-price. Using data from JD.com, \cite{qin2022jd} formulate a tripartite matching problem between robots, pods, and workstations. Using data from Amazon, \cite{allgor2023algorithm} optimize item-pod assignments, item-workstation assignments, as well as temporal dynamics, using a multi-step decomposition approach. Our paper contributes an integrated model to jointly optimize order-workstation assignments, item-pod assignments, and the schedule of order fulfillments, while capturing congestion within the warehouse and human workload at the workstations.

Finally, the structure of the TDS-CW falls into vehicle routing problems with flexibility regarding each trip's origin and destination, such as roaming delivery locations \citep{ozbaygin2017branch}, vehicle-drone coordination \citep{carlsson2017coordinated,poikonen2019mothership}, and vehicle-customer coordination \citep{zhang2021routing}. The TDS-CW involves new mechanisms for endogenous origins and destinations in pickup-and-delivery problems---item-pod assignments for origins and order-workstation assignments for destinations. Moreover, the TDS-CW is not primarily constrained by vehicle availability, but rather by workload restrictions and congestion. Thus, it relates to routing problems with congestion \citep{dabia2013branch,florio2021routing,liu2023branch}; however, this literature models congestion via exogenous variations in travel times, whereas congestion depends endogenously on pod movements in the TDS-CW. These complexities render the TDS-CW highly challenging in large-scale instances, motivating our solution algorithm.

\subsubsection*{Machine learning and optimization.} Our algorithm falls into the recent literature using machine learning for optimization problems \citep{bengio2021machine}. \cite{bertsimas2022online} use classification trees to learn selected variables in binary optimization and binding constraints in linear optimization. \cite{larsen2022predicting} develop an aggregation-prediction approach for two-stage stochastic optimization with a complex second-stage problem. Several studies used machine learning to guide optimization algorithms, such as fixed-point optimization \citep{sambharya2023learning}, branching \citep{khalil2016learning,gupta2020hybrid}, local branching \citep{liu2022learning}, cutting planes \citep{tang2020reinforcement,dey2022cutting}, and column generation \citep{morabit2021machine}. Our paper follows a similar approach by leveraging machine learning to accelerate large-scale neighborhood search.

Within the realm of LSNS, \cite{chen2019learning} train a deep reinforcement learning model to select a (small) neighborhood and an update rule in local search. \cite{hottung2019neural} train a deep neural network to learn the ``repair'' procedure in LSNS; in contrast, our objective is to construct a subproblem (the ``destroy'' operator) and we then use integer optimization to improve the solution (the ``repair'' operator). \cite{moll2000machine} train a machine learning model to mimic the ``repair'' function in local search, which they use as a hill-climbing heuristic to select a subproblem. \cite{song2020general} propose imitation learning and reinforcement learning to decompose an optimization problem into sequential subproblems---a simpler decomposition structure than iterative ``destroy and repair' LSNS algorithms. In vehicle routing, \cite{li2021learning} train a transformer model to find the subproblem (i.e., set of cities) with the strongest objective improvement, by exploiting the routing structure to avoid searching over an exponential number of subproblems.

Our paper contributes a new learn-then-optimize approach to generate a subproblem at each LSNS iteration, using machine learning to predict the local objective as a function of low-dimensional neighborhood features and integer optimization to subsequently generate a subproblem. This approach optimizes over an exponential number of candidate subproblems, as opposed to restricting the search to a small subset of candidates visited in the training data. Extensive research has developed procedures to construct effective neighborhoods in large-scale neighborhood search \citep[see][for a comprehensive review]{ahuja2002survey}. \cite{ghiani2015model} use unsupervised learning to define neighborhoods around homogeneous clusters. \cite{adamo2017mip,adamo2017automatic} extract semantic features of optimization problem and the incumbent solutions to automate the construction of new neighborhoods. In contrast, our approach uses supervised learning to predict solution improvements as a function of the entities constituting a neighborhood. \cite{liu2022machine} adopt a similar objective, by training a graph neural network to predict which variables can improve the incumbent solution, and subsequently define a new neighborhood. Our approach relies on a simpler learning structure but leverages integer optimization to construct attractive neighborhoods for LSNS re-optimization---for instance, neighborhoods with spatial-temporal continuity, with synergistic components, and of small enough size to retain computational efficiency.

\section{Task design and scheduling with congestion and restricted workload}\label{sec:model}

\subsection{Problem statement}

The TDS-CW optimizes the spatial-temporal assignment of pods to pick up items and assemble orders at the workstations. The scope is tactical, over a moderate planning horizon (15 minutes, in our experiments). As such, the problem lies between operational problems (e.g., multi-agent path-finding) and strategic problems (e.g., facility design). The TDS-CW takes the following inputs: 
\begin{itemize}[itemsep=0pt,topsep=0pt]
    \item[--] \textit{Facility layout:} sets of storage locations $\calL$, warehouse intersections $\calJ$, and workstations $\calW$. 
    \item[--] \textit{Pods:} pod starting locations, pod storage locations, and item inventories on each pod. Each pod needs to be returned to its storage location after each movement. By design, and to create operating flexibility, each item can be located on multiple pods across the facility.
    \item[--] \textit{Orders:} incoming orders with corresponding items. All items need to be delivered to the same workstation. The problem maximizes throughput, without applying order deadlines.
\end{itemize}

Spatial-temporal fulfillment operations include pod movements from their storage locations to the workstations and back (the task design layer) as well as the timing and sequencing of those movements (the scheduling layer). We propose a multi-objective formulation to maximize throughput while managing the two main operating bottlenecks---human workload and congestion.

Specifically, the throughput objective aims to maximize the number of items picked across the planning horizon. Increasing throughput can be achieved by (i) avoiding idle time at the workstations; and by (ii) picking multiple items from each pod due to changeover constraints whenever a new pod is brought to a workstation. We assume that it takes a time $\delta^{\text{item}}$ for a worker to pick an item from one pod, and a time $\delta^{\text{pod}}$ to process each pod at a workstation. Recall that each item is stored on multiple pods across the warehouse, thus creating a trade-off between bringing an item to a workstation from a nearby pod (hence, reducing lead times and congestion) versus finding another pod that contains other items (thus, reducing processing times). As such, the problem involves coordinating order, item, and pod assignments to leverage the full flexibility of the facility.

Next, the human workload objective is formalized by limiting the number of open orders at any workstation at any time. A multi-item order is open from the time the first item is picked at the workstation until the last item is processed. By definition, a single-item order is never open. Let $C_w$ denote the maximum number of open orders at workstation $w\in\calW$ ($C_w=4$ in our experiments).

Finally, the growth of robotic operations has come into conflict with narrow warehouse corridors that can only accommodate a handful of pods at a time, giving rise to our congestion management objective. This objective is highly complex to formulate due to the interdependencies between upstream task design and downstream pod movements. Ideally, we would integrate a multi-agent path finding model into the TDS-CW to reflect pod routes, but this would come at great cost in terms of scalability. Instead, we use a proxy for congestion to avoid overcrowding at intersections. This proxy captures the first-order interactions between task design and route planning decisions---notably, a pod located far from a workstation is more likely to create congestion than a nearby pod---while retaining tractability. We provide more details on our proxy for congestion below.

\subsection{Time-space network representation of pod movements}

We define a time-space network over a planning horizon discretized into a set $\calT$ of time periods, each of length $\Delta$. Each time-space node specifies a physical location $\ell \in \calW \cup \calL$ (workstation or pod storage), and a time $t \in \calT$. Time-space arc $a\in\calA$ include (i) traveling arcs in $\calA_{\text{tr}}$ comprising all movements from node $(\ell_1,t_1)$ to node $(\ell_2,t_2)$, where $\ell_1 \neq \ell_2$ and $t_2 - t_1$ is the travel time rounded up to the nearest time period; and (ii) idle arcs in $\calA_{\text{id}}$ from node $(\ell,t)$ to node $(\ell,t+1)$.

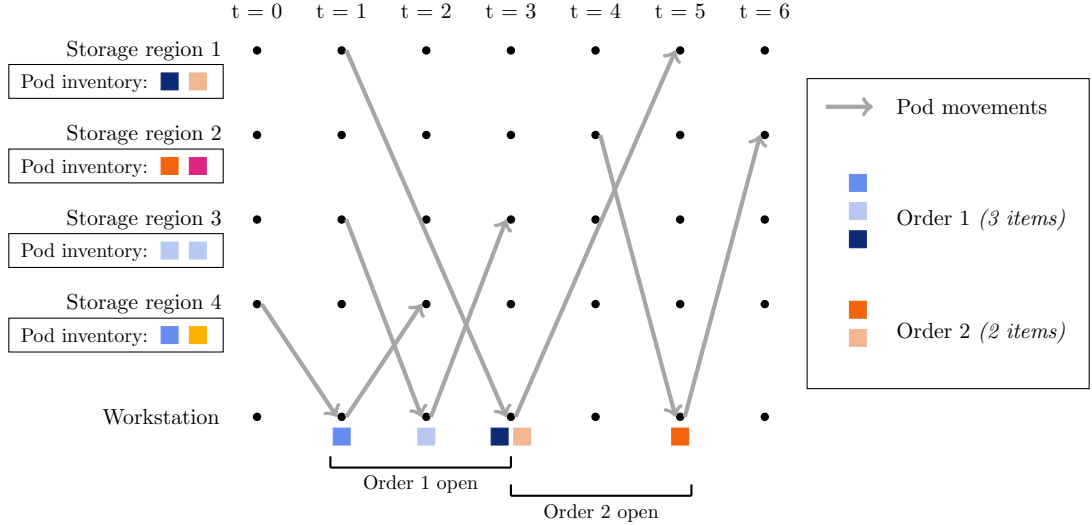
\begin{figure} %%
     \centering
        \begin{tikzpicture}[scale=0.75,transform shape,
        roundnode/.style={circle,fill,inner sep=1.5pt},
        roundnode2/.style={circle,color=white,fill,inner sep=1.5pt}
        ]
        %Auxiliary nodes
        \node[roundnode2] at (0,-1.9)   (l3t1_up) {};

        %Nodes
        \node[roundnode] at (0,0)   (l1t0) {};
        \node[roundnode] at (0,-1.5)   (l2t0) {};
        \node[roundnode] at (0,-3)   (l3t0) {};
        \node[roundnode] at (0,-4.5)   (l4t0) {};
        \node[roundnode] at (0,-6.5)   (lwt0) {};
        \node[roundnode] at (1.5,0)   (l1t1) {};
        \node[roundnode] at (1.5,-1.5)   (l2t1) {};
        \node[roundnode] at (1.5,-3)   (l3t1) {};
        \node[roundnode] at (1.5,-4.5)   (l4t1) {};
        \node[roundnode] at (1.5,-6.5)   (lwt1) {};
        \node[roundnode] at (3,0)   (l1t2) {};
        \node[roundnode] at (3,-1.5)   (l2t2) {};
        \node[roundnode] at (3,-3)   (l3t2) {};
        \node[roundnode] at (3,-4.5)   (l4t2) {};
        \node[roundnode] at (3,-6.5)   (lwt2) {};
        \node[roundnode] at (4.5,0)   (l1t3) {};
        \node[roundnode] at (4.5,-1.5)   (l2t3) {};
        \node[roundnode] at (4.5,-3)   (l3t3) {};
        \node[roundnode] at (4.5,-4.5)   (l4t3) {};
        \node[roundnode] at (4.5,-6.5)   (lwt3) {};
        \node[roundnode] at (6,0)   (l1t4) {};
        \node[roundnode] at (6,-1.5)   (l2t4) {};
        \node[roundnode] at (6,-3)   (l3t4) {};
        \node[roundnode] at (6,-4.5)   (l4t4) {};
        \node[roundnode] at (6,-6.5)   (lwt4) {};
        \node[roundnode] at (7.5,0)   (l1t5) {};
        \node[roundnode] at (7.5,-1.5)   (l2t5) {};
        \node[roundnode] at (7.5,-3)   (l3t5) {};
        \node[roundnode] at (7.5,-4.5)   (l4t5) {};
        \node[roundnode] at (7.5,-6.5)   (lwt5) {};
        \node[roundnode] at (9,0)   (l1t6) {};
        \node[roundnode] at (9,-1.5)   (l2t6) {};
        \node[roundnode] at (9,-3)   (l3t6) {};
        \node[roundnode] at (9,-4.5)   (l4t6) {};
        \node[roundnode] at (9,-6.5)   (lwt6) {};
        
        %Labels
        \node[xshift=-2cm] at (l1t0) {Storage region 1};
        \node[xshift=-2cm] at (l2t0) {Storage region 2};
        \node[xshift=-2cm] at (l3t0) {Storage region 3};
        \node[xshift=-2cm] at (l4t0) {Storage region 4};
        \node[xshift=-1.7cm] at (lwt0) {Workstation};
        \node[yshift=0.7cm] at (l1t0) {t = 0};
        \node[yshift=0.7cm] at (l1t1) {t = 1};
        \node[yshift=0.7cm] at (l1t2) {t = 2};
        \node[yshift=0.7cm] at (l1t3) {t = 3};
        \node[yshift=0.7cm] at (l1t4) {t = 4};
        \node[yshift=0.7cm] at (l1t5) {t = 5};
        \node[yshift=0.7cm] at (l1t6) {t = 6};

        %Pod arcs 
        \draw[gray!70, ultra thick, ->] (l4t0.east) -- (lwt1.west);
        \draw[gray!70, ultra thick, ->] (lwt1.east) -- (l4t2.west);
        \draw[gray!70, ultra thick, ->] (l1t1.east) -- (lwt3.west);
        \draw[gray!70, ultra thick, ->] (lwt3.east) -- (l1t5.west);
        \draw[gray!70, ultra thick, ->] (l3t1.east) -- (lwt2.west);
        \draw[gray!70, ultra thick, ->] (lwt2.east) -- (l3t3.west);
        \draw[gray!70, ultra thick, ->] (l2t4.east) -- (lwt5.west);
        \draw[gray!70, ultra thick, ->] (lwt5.east) -- (l2t6.west);
        
        %Items picked
        \draw[draw=mymidblue, fill=mymidblue] (1.35,-7) rectangle ++(0.3,0.3);
        \draw[draw=mylightblue, fill=mylightblue] (2.85,-7) rectangle ++(0.3,0.3);
        \draw[draw=mydarkblue, fill=mydarkblue] (4.15,-7) rectangle ++(0.3,0.3);
        \draw[draw=mylightorange, fill=mylightorange] (4.55,-7) rectangle ++(0.3,0.3);
        \draw[draw=mymidorange, fill=mymidorange] (7.35,-7) rectangle ++(0.3,0.3);
        
        %Open orders
        \draw[black, thick, -] (1.3,-7.2) -- (1.3,-7.4);
        \draw[black, thick, -] (1.3,-7.4) -- (4.5,-7.4);
        \draw[black, thick, -] (4.5,-7.2) -- (4.5,-7.4);
        \node[yshift=-0.3cm] at (2.9,-7.4) {\small Order 1 open};
        
        \draw[black, thick, -] (7.7,-7.7) -- (7.7,-7.9);
        \draw[black, thick, -] (7.7,-7.9) -- (4.5,-7.9);
        \draw[black, thick, -] (4.5,-7.7) -- (4.5,-7.9);
        \node[yshift=-0.3cm] at (6.1,-7.9) {\small Order 2 open};
        
        %Inventory
        \draw[draw=mydarkblue, fill=mydarkblue] (-1.7,-0.7) rectangle ++(0.3,0.3);
        \draw[draw=mylightorange, fill=mylightorange] (-1.2,-0.7) rectangle ++(0.3,0.3);
        \draw[draw=black] (-4.4,-0.85) rectangle ++(3.8,0.6);
        \node[xshift=-3cm, yshift=-0.58cm] at (l1t0) {\text{\small Pod inventory: }};
        
        \draw[draw=mymidorange, fill=mymidorange] (-1.7,-2.2) rectangle ++(0.3,0.3);
        \draw[draw=mypink, fill=mypink] (-1.2,-2.2) rectangle ++(0.3,0.3);
        \draw[draw=black] (-4.4,-2.35) rectangle ++(3.8,0.6);
        \node[xshift=-3cm, yshift=-0.58cm] at (l2t0) {\text{\small Pod inventory: }};
        
        \draw[draw=mylightblue, fill=mylightblue] (-1.7,-3.7) rectangle ++(0.3,0.3);
        \draw[draw=mylightblue, fill=mylightblue] (-1.2,-3.7) rectangle ++(0.3,0.3);
        \draw[draw=black] (-4.4,-3.85) rectangle ++(3.8,0.6);
        \node[xshift=-3cm, yshift=-0.58cm] at (l3t0) {\text{\small Pod inventory: }};
        
        \draw[draw=mymidblue, fill=mymidblue] (-1.7,-5.2) rectangle ++(0.3,0.3);
        \draw[draw=myyellow, fill=myyellow] (-1.2,-5.2) rectangle ++(0.3,0.3);
        \draw[draw=black] (-4.4,-5.35) rectangle ++(3.8,0.6);
        \node[xshift=-3cm, yshift=-0.58cm] at (l4t0) {\text{\small Pod inventory: }};

        %Legend
        \node[roundnode2] at (10,-1)   (pod1) {};
        \node[roundnode2] at (11,-1)   (pod2) {};
        \draw[gray!70, ultra thick, ->] (pod1.east) -- (pod2.west);
        \node[xshift=0.2cm, anchor=west] at (pod2) {Pod movements};
        
        \node[roundnode2] at (11,-3) (order1) {};
        \draw[draw=mymidblue, fill=mymidblue] (10.5,-2.5) rectangle ++(0.3,0.3);
        \draw[draw=mylightblue, fill=mylightblue] (10.5,-3) rectangle ++(0.3,0.3);
        \draw[draw=mydarkblue, fill=mydarkblue] (10.5,-3.5) rectangle ++(0.3,0.3);
        \node[xshift=0.2cm, anchor=west] at (order1) {Order 1 \textit{(3 items)}};
        
        \node[roundnode2] at (11,-5) (order2) {};
        \draw[draw=mymidorange, fill=mymidorange] (10.5,-4.75) rectangle ++(0.3,0.3);
        \draw[draw=mylightorange, fill=mylightorange] (10.5,-5.25) rectangle ++(0.3,0.3);
        \node[xshift=0.2cm, anchor=west] at (order2) {Order 2 \textit{(2 items)}};
        
        \draw[draw=black] (9.75,-0.5) rectangle ++(5,-5.5);

    \end{tikzpicture}
    \caption{Time-space network representation of fulfillment operations.}
    \label{fig:TSnetwork}
    \vspace{-18pt}
\end{figure}

Figure~\ref{fig:TSnetwork} illustrates a time-space network with one workstation and four storage locations, each containing one pod and two items in inventory. The facility receives a three-item order (blue) and a two-item order (orange). The blue order is fulfilled using three pods brought to the workstations at times $t = $ 1, 2, and 3. The latter pod contains one item from the blue order as well as an item from the orange order, so it is brought to the workstation only once for both items to be picked. This solution enables to process five items with four pods, with only one order open at any time.

The time-space network coordinates pods and orders without relying on ``big-M'' constraints. In particular, this representation links pod movements to item processing at the workstations to formulate our throughput maximization objective and our workload constraints. Similarly, it encapsulates the spatial-temporal flows of pod movements to manage congestion in the facility.

Specifically, our model of congestion assumes that each intersection $j\in\calJ$ can be traversed by up to $Q_j$ pods at a time. Recall that each time-space arc $a\in\calA_{\text{tr}}$ captures a pod assignment between storage locations and workstations, but does not optimize movements in between to avoid creating a much larger time-space network. We assume that each pod travels through one of the several shortest paths in an $\ell_1$ grid, in line with the goal of minimizing complexity \citep{allgor2023algorithm}. We compute the pod's \textit{congestion contribution} $q_{ajt}$ as the fraction of all shortest paths corresponding to time-space arc $a$ that pass through intersection $j\in\calJ$ at time $t\in\calT$. Figure~\ref{fig:congestion} illustrates this model: for a time-space arc with 20 shortest paths, one passes through the blue intersection at $t=30$, nine pass through the pink intersection at $t=30$, and none passes through the orange intersection---with congestion contributions of $0.05$, $0.45,$ and $0$, respectively. We then constrain the total congestion contribution at each intersection-time. This model captures the interdependencies between task design decisions and congestion, while retaining an arc-based structure in the time-space network.

\begin{figure} %%
     \centering
        \begin{tikzpicture}[transform shape = 0.9,
        roundnode/.style={circle,fill,inner sep=1.5pt},
        roundnode2/.style={circle,color=white,fill,inner sep=1.5pt}
        ]

        %Start and end nodes
        \node[roundnode] at (5.5,4.375)   (orig) {};
        \node[roundnode] at (0.5,0.375)   (dest) {};
    
        %Intersections
        \foreach \x in {0,...,6}
            \foreach \y [count=\yi] in {0,...,4}  
              \draw[gray!80, thick, -] (\x*1.25-0.2,\y+0.8)--(\x*1.25-0.05,\y+0.95) ;
        \foreach \x in {0,...,6}
            \foreach \y [count=\yi] in {0,...,4}  
              \draw[gray!80, thick, -] (\x*1.25-0.2,\y+0.95)--(\x*1.25-0.05,\y+0.8) ;
        \foreach \x in {0,...,6}
            \foreach \y [count=\yi] in {0,...,4}  
              \draw[gray!80, thick, -] (\x*1.25-0.2,\y+0.8)--(\x*1.25-0.05,\y+0.95) ;
        \foreach \x in {0,...,6}
            \foreach \y [count=\yi] in {0,...,4}  
              \draw[gray!80, thick, -] (\x*1.25-0.2,\y+0.8)--(\x*1.25-0.05,\y+0.95) ;

        %Queues 
        \draw[gray!80, thick, -] (1.25-0.2,-0.5+0.8)--(1.25-0.05,-0.5+0.95) ;
        \draw[gray!80, thick, -] (1.25-0.2,-0.5+0.95)--(1.25-0.05,-0.5+0.8) ;
        \draw[gray!80, thick, -] (4*1.25-0.2,-0.5+0.8)--(4*1.25-0.05,-0.5+0.95) ;
        \draw[gray!80, thick, -] (4*1.25-0.2,-0.5+0.95)--(4*1.25-0.05,-0.5+0.8) ;
        
        %Outline
        \draw[draw=black] (-0.25,-0.25) rectangle ++(7.75,5.25);
        
        %Location squares
        \draw[draw=black] (0,0) rectangle ++(1,0.75);
        \draw[draw=black] (3.75,0) rectangle ++(1,0.75);
        \foreach \x in {0,...,5}
            \foreach \y [count=\yi] in {1,...,4}  
                \draw[draw=black] (\x*1.25,\y) rectangle ++(1,0.75);

        %Paths
        \draw[black, ultra thick, -] (4.875,4.375) -- (orig);
        \draw[black, ultra thick, -] (1.125,0.375) -- (1.125,3.875);
        \draw[black, ultra thick, -] (2.375,0.875) -- (2.375,3.875);
        \draw[black, ultra thick, -] (3.625,0.875) -- (3.625,3.875);
        \draw[black, ultra thick, -] (4.875,0.875) -- (4.875,4.375);   
        \draw[black, ultra thick, -] (1.125,0.875) -- (4.875,0.875);
        \draw[black, ultra thick, -] (1.125,1.875) -- (4.875,1.875);
        \draw[black, ultra thick, -] (1.125,2.875) -- (4.875,2.875);
        \draw[black, ultra thick, -] (1.125,3.875) -- (4.875,3.875); 
        \draw[black, ultra thick, <-] (dest) -- (1.125,0.375) ;
        
        %Highlighted intersections
        \draw[mymidblue, fill=mymidblue, ultra thick, -] (1.125,3.875) circle (0.11cm);
        \draw[white, very thick, -] (1.05,3.8) -- (1.2,3.95) ;
        \draw[white, very thick, -] (1.05,3.95) -- (1.2,3.8) ;
        
        \draw[mypink, fill=mypink, ultra thick, -] (3.625,1.875) circle (0.11cm);
        \draw[white, very thick, -] (3.55,1.8) -- (3.7,1.95) ;
        \draw[white, very thick, -] (3.55,1.95) -- (3.7,1.8) ;
        
        \draw[mymidorange, fill=mymidorange, ultra thick, -] (6.125,2.875) circle (0.11cm);
        \draw[white, very thick, -] (6.05,2.8) -- (6.2,2.95) ;
        \draw[white, very thick, -] (6.05,2.95) -- (6.2,2.8) ;
        
        %Congestion contributions
        \node[xshift=-0.3cm, anchor=west] at (8.5,4) {\underline{\textbf{Congestion contributions}}};
        \draw[mymidblue, fill=mymidblue, ultra thick, -] (9.3,3) circle (0.11cm);
        \draw[white, very thick, -] (9.375,3.075) -- (9.225,2.925) ;
        \draw[white, very thick, -] (9.375,2.925) -- (9.225,3.075) ;
        \node[xshift=0.2cm, anchor=west] at (9.3,3) {$q_{a \color{myblue}j \color{black}t} = \frac{1}{20} = 0.05$ };
        
        \draw[mypink, fill=mypink, ultra thick, -] (9.3,2) circle (0.11cm);
        \draw[white, very thick, -] (9.375,2.075) -- (9.225,1.925) ;
        \draw[white, very thick, -] (9.375,1.925) -- (9.225,2.075) ;
        \node[xshift=0.2cm, anchor=west] at (9.3,2) {$q_{a \color{mypink}j \color{black}t} = \frac{9}{20} = 0.45$ };
        
        \draw[mymidorange, fill=mymidorange, ultra thick, -] (9.3,1) circle (0.11cm);
        \draw[white, very thick, -] (9.375,1.075) -- (9.225,0.925) ;
        \draw[white, very thick, -] (9.375,0.925) -- (9.225,1.075) ;
        \node[xshift=0.2cm, anchor=west] at (9.3,1) {$q_{a \color{myorange}j \color{black}t} = \frac{0}{20} = 0.00$ };
        
    \end{tikzpicture}
    \caption{Congestion contribution calculation for a traveling arc $a\in\calA_{\text{tr}}$ that departs its origin at time $t=0$ and arrives at a workstation 60 seconds later. There are 20 shortest paths, shown in black. The congestion contribution for three intersections at time $t=30$ seconds is calculated by counting the fraction of those paths that pass through the intersection 30 seconds after leaving the origin.}
    \label{fig:congestion}
    \vspace{-12pt}
\end{figure}

\subsection{Formulation}

The TDS-CW optimizes order-workstation assignments, item-pod assignments, and pod movements in the warehouse. We define the following decision variables:
{\small\begin{align*}
   z_{mw} &=
  \begin{cases}
                                   1 & \text{if order $m\in\calM$ is assigned to workstation $w\in\calW$} \\
                                   0 & \text{otherwise} \\
  \end{cases}\\ 
  x_{impwt} &=
  \begin{cases}
                                   1 & \text{if item $i\in\calI$ for order $m\in\calM$ is picked from pod $p\in\calP$ at workstation $w\in\calW$ \textbf{by} time $t\in\calT$} \\
                                   0 & \text{otherwise} \\
  \end{cases}\\
  y_{pa} &=
  \begin{cases}
                                   1 & \text{if pod $p$ is assigned to time-space network arc $a\in\calA$} \\
                                   0 & \text{otherwise} \\
  \end{cases}\\
  v_{mwt} &=
  \begin{cases}
                                   1 & \text{if order $m$ is open at workstation $w\in\calW$ at time $t\in\calT$} \\
                                   0 & \text{otherwise} \\
  \end{cases}\\
  f_{mwt} &=
  \begin{cases}
                                   1 & \text{if at least one item from order $m\in\calM$ is delivered to workstation $w\in\calW$ \textbf{by} time $t\in\calT$} \\
                                   0 & \text{otherwise} \\
  \end{cases}\\
  g_{mwt} &=
  \begin{cases}
                                   1 & \text{if all items from order $m\in\calM$ have been delivered to workstation $w\in\calW$  \textbf{by} time $t\in\calT$} \\
                                   0 & \text{otherwise} \\
  \end{cases}
\end{align*}}

An important observation is the representation of the scheduling variables $\bx$, $\bff$, and $\bg$ characterizing an event happening \textbf{by} time $t\in\calT$, instead of \textbf{at} time $t\in\calT$. Therefore, the corresponding temporal vectors are of the form $(0,\cdots,0,1,\cdots,1)$, instead of $(0,\cdots,0,1,0,\cdots,0)$. For instance, item $i\in\calI$ from order $m\in\calM$ is picked up at workstation $w\in\calW$ at time $t\in\calT$ if $x_{impwt} - x_{impw,t-1}=1$. This representation, inspired by \cite{stockpatterson} in air traffic flow management, is crucial for modeling human workload restrictions with a tight polyhedral formulation.

\begin{table}
\centering\small\renewcommand{\arraystretch}{1.0}
\begin{tabular}{ll}\toprule
Notation & Description  \\ \midrule
$\calM$ & set of orders \\
$\calI$ & set of items; the subset of items making up order $m\in\calM$ is denoted by $\calI_m$ \\
$\calW$ & set of workstations \\
$\calL$ & set of storage locations \\
$\calJ$ & set of intersections \\
$\calP$ & set of pods; the subset of pods with supply of item $i\in\calI$ is denoted by $\calP_i$ \\
$\calT$ & set of time periods, with time discretization $\Delta$ \\
$\calN$ & set of time-space nodes, i.e. $(\calW \cup \calL) \times \{1,\cdots, T\}$ \\
$\calN_S$ & set of time-space nodes representing storage locations \\
$\calN_{\text{end}}$ & set of time-space nodes at the end of the time horizon\\
$N(\ell, t)$ & time-space node representing location $\ell \in \calW \cup \calL$ at time $t\in\calT$\\
$\calA$ & set of time-space arcs, divided into traveling arcs $\calA_{\text{tr}}$ and idle arcs $\calA_{\text{id}}$ \\
$\calA_n^{-}$, $\calA_n^{+}$ & set of incoming and outgoing arcs in time-space node $n\in\calN$ \\
$t_a^{0}$, $\ell_a^{0}$ & start time and start location of arc $a\in\calA$ \\
$s_p$ & storage location of pod $p\in\calP$ \\
$u_{ip}$ & initial inventory of item $i\in\calI$ at pod $p\in\calP$ \\
$N_p^{0}$ & starting node of pod $p\in\calP$ \\
$C_w$ & maximum number of orders that can be open at workstation $w\in\calW$ during any time period \\
$\delta^{\text{item}}$, $\delta^{\text{pod}}$ & time to pick an item and to process a pod at a workstation, respectively \\
$q_{ajt}$ & congestion contribution of time-space arc $a\in\calA$ to intersection $j\in\calJ$ at time $t\in\calT$ \\
\bottomrule
\end{tabular}
\caption{Input data for the integer optimization formulation.}
\label{tab:notation}
\vspace{-6pt}
\end{table}

The full integer optimization formulation is then given as follows:
{\small\begin{align} 
\max  & \displaystyle \sum_{m \in \calM} \sum_{i \in \calI_m} \sum_{p \in \calP_i} \sum_{w \in \calW} x_{impwT} \label{obj} \\
    	\text{s.t   } &\displaystyle \textbf{Workstation constraints} \nonumber  \\
    	&\displaystyle \sum_{w \in \calW} z_{mw} \leq 1 \quad \forall m \in \calM \label{wsassign} \\
    	&\displaystyle  \sum_{m \in \calM} v_{mwt} \leq C_w \quad \forall w \in \calW, t \in \calT \label{maxopen} \\
    	&\displaystyle \sum_{m \in \calM} \sum_{i \in \calI_m} \sum_{p \in \calP_i} \delta^{\text{item}} (x_{impwt} - x_{impw,t-1}) + \sum_{p \in \calP} \sum_{a \in \calA^+_{N(w,t)}} \delta^{\text{pod}} y_{pa} \leq \Delta \quad \forall w \in \calW, t \in \calT \label{throughput} \\
    	&\displaystyle \textbf{Pod operations and congestion} \nonumber \\
    	&\displaystyle  \sum_{a \in \calA^+_{N_p^{0}}} y_{pa} = 1 \label{podstart} \\
    	&\displaystyle  \sum_{a \in \calA^-_{n}} y_{pa} - \sum_{a \in \calA^+_{n}} y_{pa} = 0 \quad \forall p \in \calP, n \in \calN \setminus (\calN_{end} \cup \{ N_p^{0}\})  \label{podfb} \\
    	&\displaystyle \sum_{w \in \calW} \sum_{m \in \calM} x_{impwT} \leq u_{ip} \quad \forall i \in \calI, p \in \calP_i \label{inventory} \\
    	&\displaystyle \sum_{p \in \calP} \sum_{a \in \calA} q_{ajt} y_{pa} \leq Q_{j} \quad \forall j \in \mathcal{J}, t \in \calT \label{majtraffic} \\
    	&\displaystyle \textbf{Consistency constraints} \nonumber \\
    	&\displaystyle (x_{impwt} -  x_{impw,t-1}) \leq   \sum_{a \in \calA_{\text{tr}} \cap \calA^+_{N(w,t)}} y_{pa} \quad \forall m \in \calM, i \in \calI_m, p \in \calP_i, w \in \calW, t \in \calT \label{hylink} \\
    	&\displaystyle \sum_{p \in \calP_i} \left(x_{impwt} -  x_{impw,t-1}\right) \leq  z_{mw} \quad \forall m \in \calM, i \in \calI_m, w \in \calW, t \in \calT \label{hzlink} \\
    	&\displaystyle x_{impwt} \geq x_{impw,t-1} \quad \forall m \in \calM, i \in \calI_m, p \in \calP_i, w \in \calW, t \in \calT \label{nondec}\\
    	&\displaystyle  f_{mwt} \geq x_{impwt} \quad \forall m \in \calM, i \in \calI_m, p \in \calP_i, w \in \calW, t \in \calT \label{flink}  \\
    	&\displaystyle  g_{mwt} \leq \sum_{p \in \calP_i} x_{impwt} \quad \forall m \in \calM, i \in \calI_m, w \in \calW, t \in \calT \label{glink} \\
    	&\displaystyle  v_{mwt} = f_{mwt} - g_{mwt} \quad \forall m \in \calM, w \in \calW, t \in \calT \label{vlink} \\
    	&\displaystyle \textbf{Domain of definition: $\bx,\ \by,\bz,\ \bff,\ \bg,\ \bv$ binary} \nonumber \label{binvars}
\end{align}}

Equation~\eqref{obj} maximizes the number of items picked by the end of the horizon. Constraints~\eqref{wsassign} assign each order to at most one workstation. Constraints~\eqref{maxopen} enforce the maximum number of orders open at each workstation. Constraints~\eqref{throughput} apply the workload restrictions given the times to process a pod and to pick up an item. Constraints~\eqref{podstart} and~\eqref{podfb} enforce flow conservation in the time-space network. Constraints~\eqref{inventory} ensure that picking operations are consistent with available pod-item inventories. Constraints~\eqref{majtraffic} cap the congestion contribution to the capacity at each intersection. Constraints~\eqref{hylink} link assignments to flow variables: if item $i$ is picked from pod $p$ at workstation $w$ at time $t$, then pod $p$ departs from workstation $w$ at time $t$ to allow the next pod to be processed. Constraints~\eqref{hzlink} ensures that all items in the same order get delivered to the same workstation: if item $i$ for order $m$ is picked from pod $p$ at workstation $w$ at time $t$, then order $m$ is assigned to workstation $w$. Constraints~\eqref{nondec} ensure that $\bx$ is non-decreasing. Constraints~\eqref{flink} (resp. Constraints~\eqref{glink}) track whether \textit{at least one} item (resp. \textit{all} items) from order $m$ got picked at workstation $w$ by time $t$. Finally, Constraints~\eqref{vlink} define whether order $m$ is open at workstation $w$ at time $t$ based on the $\bff$ and $\bg$ indicators of its start and completion times.

Note that the description of the TDS-CW in this section allows a pod to visit multiple workstations in a single trip, but relies on the assumption that the pod must return to its original storage location after visiting the last workstation. The model can account for flexible pod storage policies by adding arcs to the time-space network and defining additional constraints in the model (e.g., capacity on the number of pods per storage location). We describe this extension in Appendices~\ref{subsec:flexible} and study the impact of allowing multiple workstation visits per pod trip in Appendix~\ref{subsec:multistop}. 

\subsection{Size and Complexity}

The TDS-CW consists of a large-scale assignment problem governing task design and scheduling decisions. The reliance on multi-dimensional assignment variables enables a strong integer optimization formulation, with a tight linear relaxation. However, the sheer size of the model makes it highly challenging. Table~\ref{tab:size} provide some statistics for several instances developed in collaboration with our industry partner. The problem can involve millions of binary variables and constraints in small-sized and medium-sized facilities, and up to tens of millions of binary variables and over a hundred million constraints in full-scale fulfillment centers encountered in practice. To get a sense of the problem's complexity, off-the-shelf solvers fail to return a feasible solution with non-zero throughput in the small-sized instance and run into memory issues in the medium-sized instance.

\begin{table} %%
\small
\centering
\renewcommand{\arraystretch}{1}
\begin{tabular}{lrrrrrrrr}\toprule
%&&&& \multicolumn{3}{c}{CPU (seconds)} &&& \\ \cmidrule(lr){5-7}
Instance & Workstations & Locations & Items & Pods & Orders & Time Periods & Variables & Constraints \\ 
\midrule
%Tiny & 2 & 2 & 100 & 16 & 100 & 30 & 68,760 & 278,380 \\ 
Small & 2 & 14 & 1,000 & 336 & 100 & 30 & 246,680 & 1,206,820 \\ 
Medium & 3 & 30 & 10,000 & 720 & 200 & 30 & 946,400 & 4,824,200 \\ 
Full & 20 & 140 & 200,000 & 3,360 & 1,000 & 30 & 22,853,600 & 120,543,400 \\ 
%Full-Large & 30 & 300 & 500,000 & 7,200 & 2,000 & 30 & 59,460,000 & 341,311,700 \\ 
%Full & 30 & 1,260 & 500,000 & 30,240 & 2,000 & 30 & 232,260,000 & 2,269,846,100 \\ 
\bottomrule
\end{tabular}
\caption{Model size for three instances of TDS-CW.}
\label{tab:size}
\vspace{-12pt}
\end{table}

Thus, we seek a heuristic that can reliably generate high-quality solutions to the TDS-CW. One possibility would be to decompose the problem into sequential decisions \citep[see, e.g.][]{valle2021order,wang2022robot,allgor2023algorithm}. In our case, this would involve separating order-workstation assignments, pod assignments at the pre-determined workstation, and pod movements to fulfill those tasks. This approach, however, may lead to poor-quality solutions by ignoring interdependencies between upstream assignments and downstream operations; moreover, it may still involve long computational times due to the complexity of the TDS-CW. In our results, decomposition heuristics achieve 30-75\% utilization in over 48 hours of computations (Section~\ref{sec:results}). In fact, these limitations point to the strength of integer optimization to simultaneously manage multiple constraints and generate high-quality solutions in large-scale environments. Next, we present our LSNS algorithm to generate TDS-CW solutions by solving a series of very small instances with up to twenty orders, thirty pods, one or two workstations, and a two- to three-minute horizon.

\section{Learn-then-optimize algorithm for large-scale neighborhood search} \label{sec:algorithm}

\subsection{Large-scale neighborhood search (LSNS)}

Let us present our algorithm in a generic optimization setting. We denote a set of ``elements'' $\calE$; let $\calS\subseteq2^\calE$ be the collection of subsets in $\calE$. Assignments are made for subsets $S\in\calS^{OPT}\subseteq\calS$, via binary assignment variables $v_{S}$. In a scheduling example, $\calE$ stores jobs and machines; $\calS$ can store jobs, machines, job-machine pairs, job-job pairs, machine-machine pairs, job-job-machine combinations, etc.; and variables $v_S$ for $S\in\calS^{OPT}$ encode two-dimensional job-machine assignments. In a vehicle routing example, $\calE$ stores customers and vehicles; $\calS$ can store locations, vehicles, location-vehicle pairs, location-location arcs, vehicle-vehicle pairs, location-location-vehicle combinations, etc.; and variables $v_S$ for $S\in\calS^{OPT}$ encode three-dimensional location-location-vehicle assignments. In our TDS-CW example, $\calE$ stores orders, items, pods, workstations and time periods, and $v_S$ encodes a five-dimensional assignment ($x_{impwt}$ variables from Section~\ref{sec:model}). A generic formulation with $K$ demand constraints is given as follows, where $c_S$ denotes the cost of the assignment corresponding to subset $S$, $\gamma^k_{S}$ represents the amount of demand $k$ that is served by the assignment corresponding to subset $S$, and $b_k$ is the amount of demand $k$ that needs to be served.
\begin{align} 
\min  & \displaystyle\quad \sum_{S \in \calS^{OPT}} c_{S} v_{S}  \\
    	\text{s.t.   } &\displaystyle\quad  \sum_{S \in \calS^{OPT}}  \gamma^k_{S} v_{S} \geq b_k \quad \forall k=1,\cdots,K \label{eq:LSNS1}\\
    	&\displaystyle\quad v_{S} \in \{0,1\} \quad \forall S \in \calS^{OPT},  \label{eq:LSNS2}
\end{align}

The LSNS algorithm solves large-scale TDS-CW instances by iteratively re-optimizing the solution over a large neighborhood, while maintaining global feasibility (Algorithm~\ref{alg:lsns}). Let $\overline{\bv}$ denote an incumbent solution at a given iteration. We define a neighborhood comprising a subset of elements $\calE^{SP}\subset\calE$ and the corresponding collection of subsets, $\calS^{SP} = \{S \in \calS^{OPT} \: : \: S\subseteq\calE^{SP}\}$. The subproblem $\textbf{SP}(\calS^{SP})$ re-optimizes assignments over $\calS^{SP}$, as follows:
\begin{align} 
&\textbf{SP}(\calS^{SP}) \quad  \min  \left\{\sum_{S \in \calS^{OPT}} c_{S} v_{S}: \text{Equations~\eqref{eq:LSNS1}--\eqref{eq:LSNS2}};\ v_{S} =  \overline{v}_{S} \quad \forall S \in\calS^{OPT}\setminus \calS^{SP}\right\}\label{eq:SP}
\end{align} 

\begin{algorithm}[h!]
\caption{Large-scale neighborhood search algorithm}
\label{alg:lsns}\small
    \begin{algorithmic}
    \State $\overline{\bv} \gets \text{ find initial solution} $
    \While {NOT termination}\Comment{Termination: time limit, number of iterations, or solution improvement}
      \State $\calS^{SP} \gets \text{ select subproblem}$
      \State $\overline{\bv} \gets \text{ update solution by solving $\textbf{SP}(\calS^{SP})$ (Equation~\eqref{eq:SP})}$
    \EndWhile
    \State \textbf{return } $\overline{\bv}$
    \end{algorithmic}
\end{algorithm}

The LSNS algorithm aims to identify high-synergy neighborhoods where the solution can be re-optimized locally via adjustments, swaps and new assignments. However, complex coupling constraints in the assignment problem make it difficult to infer where such improvements are beneficial. Our learn-then-optimize approach uses machine learning to predict throughput in each neighborhood and integer optimization to construct a subproblem with the highest predicted improvement.

Before proceeding, we define a learning-enhanced benchmark that trains an offline machine learning model to predict performance within each subproblem and then \textit{selects} the subproblem with the highest predicted improvement \citep{hottung2019neural,song2020general,li2021learning}. However, an assignment problem with $n$ elements induces $\binom{n}{k} = \calO(n^k)$ candidate subproblems of cardinality $k$. Even the small TDS-CW instance comprises ${\binom{|\calM|}{20}}{ 
 \binom{|\calP|}{30}}T\simeq10^{58}$ subproblems with $20$ orders, and $30$ pods. In the offline learning phase, this combinatorial explosion prevents us from building an exhaustive training set; and in the online LSNS phase, it prevents us from enumerating all candidate subproblems, defining their features and evaluating predicted improvements. In vehicle routing, \cite{li2021learning} successfully restrict the search to a subset of candidate neighborhoods by exploiting spatial locality. In our generic assignment context, however, it may be challenging to define ``good'' subproblems as a starting point. So the learning-enhanced benchmark creates a trade-off between considering many candidate subproblems (with high computational costs) versus few candidate subproblems (leading to reduced flexibility in LSNS). Instead, our learn-then-optimize approach \textit{generates} a new subproblem at each iteration. This methodology follows similar objectives as the one from \cite{liu2022machine}, but adopts a different approach by relying on a simpler machine learning structure (linear holistic regression vs. graph neural networks) but a dedicated optimization structure to construct a new subproblem (integer optimization vs. greedy heuristics).

\subsection{Learn-then-optimize approach to subproblem construction}

Our learn-then-optimize approach seeks a subproblem $\calS^{SP}$ with a high solution improvement potential. In the training phase, it learns the local objective from low-dimensional components of the neighborhood (e.g., orders, items, pods, workstations, time periods, and their pair-wise interactions in the TDS-CW). We then formulate an integer optimization model to construct a neighborhood that maximizes the predicted improvement, without enumerating all candidate subproblems.

\vspace{-6pt}
\subsubsection*{Training data.} We collect $U$ subproblem solutions corresponding to multiple problem instances, indexed by $u=1,\cdots,U$. Each subproblem $u$ contains a subset of the elements, $\calE^u \subseteq \calE$, which induces a collection $\calS^u = \{S \in \calS \: | \: S\subseteq \calE^u\}$. Note that multiple subproblems can correspond to the same neighborhood, if they are visited multiple times for one instance or across instances. For each subproblem $u$, we derive the optimal solution $\bv^u$ and record the local objective value $y^u$:
$$y^u=\sum\limits_{S \in \calS^{OPT} \cap \calS^u} c_{S} v^u_{S}$$ 

\vspace{-6pt}
\subsubsection*{Performance learning.} We impose the structure that the local value of the objective function $y^u$ be decomposable across low-dimensional components of the neighborhood. Specifically, each feature $f=1,\cdots,F$ is associated with a collection of subsets of small cardinality, denoted by $\calS_f\subseteq\calS$. In the TDS-CW example, the ``order size'' feature is defined for each order (so each subset in $\calS_f$ is a singleton) and the ``order-pod inventory overlap'' feature is defined for each order-pod pair (so each subset in $\calS_f$ is a two-tuple). We define $\alpha_{S,f}$ as the value of feature $f=1,\cdots,F$ for subset $S\in\calS_f$. The total value of feature $f=1,\cdots,F$ in subproblem $u=1,\cdots,U$ is then given by:
\begin{align}
 x_{uf}=\sum_{S\in\calS_f\cap\calS^u}\alpha_{S,f}\label{eq:project}
\end{align}

We seek a function $\widehat{f}$ that predicts the objective value, by minimizing the norm-based difference between the predicted and actual objective values across the subproblems in the training data:
\begin{align} 
 \min_{\widehat{f}}  \quad & \sum_{u=1}^{U} \left\|y^u-\widehat{f} \left( \left\{ \sum_{S\in\calS_f\cap\calS^u}\alpha_{S,f} \:|\: f=1,\cdots,F\right\} \right)\right\|  \label{train_obj}
\end{align} 

Next, we leverage the prediction function $\widehat{f}$ to construct a subproblem at the next iteration. In principle, this could be achieved by optimizing over various learning structures, such as tree-based models \citep{mivsic2020optimization} and neural networks \citep{anderson2020strong}. Since the subproblem generation model is applied at each LSNS iteration, however, it is critical to retain a simple and scalable integer optimization structure. Therefore, we restrict the learning procedure to linear functions:
\begin{align}
     \widehat{f} \left( \left\{ \sum_{S\in\calS_f\cap\calS^u}\alpha_{S,f} \:|\: f=1,\cdots,F\right\} \right) = \sum_{f=1}^{F}\hat{\beta}_f \sum_{S\in\calS_f\cap\calS^u}\alpha_{S,f}\label{eq:linear}
\end{align}

\vspace{-6pt}
\subsubsection*{Subproblem generation.}

We seek a subproblem that maximizes the predicted improvement---e.g., a neighborhood with high predicted throughput but where the incumbent solution achieves a low throughput. We formulate this problem as an optimization model via the following variables:
\begin{align*}
    z_i&=\begin{cases}
        1&\text{if element $i=1,\cdots,n$ is included in the subproblem, i.e., if $i \in \calE^{SP} \subseteq \calE$}\\
        0&\text{otherwise.}
    \end{cases}\\
    q_S&=\begin{cases}
        1&\text{if subset $S\in \bigcup\limits_{f=1}^F\calS_f$ is included in the subproblem, i.e., if $S \in \calS^{SP} \subseteq \calS$}\\
        0&\text{otherwise.}
    \end{cases}
\end{align*}

The subproblem generation model is then formulated as follows:
\begin{align} 
 \max_{\bz,\bw}  \quad &  \sum_{f=1}^{F}\hat{\beta}_f \sum_{S\in\calS_f\cap\calS^u}\alpha_{S,f}q_S -  \sum_{S \in \calS^{OPT}} c_S \overline{v}_S q_S-C\sum_{i\in\calS_{tabu}}z_i \label{ss_obj}\\
    	\st \quad & q_S\leq z_i,\ \forall i\in S,\ \forall S\in\bigcup_{f=1}^F\calS_f\label{ss_consistency}\\
        & \bz \in \calZ \label{ss_validsp}\\
       & z_i \in \{0,1\} \quad \forall i \in \calE \label{ss_bdsZ}\\
       & q_S \in \{0,1\} \quad \forall S \in \calS \label{ss_bdsW}
\end{align}

Equation~\eqref{ss_obj} maximizes the predicted solution improvement, with a penalty to incentivize exploration. The first term applies the linear function $\widehat{f}$ to predict the local objective. The second term captures the local objective in the incumbent solution $\overline{\bv}$. The last term promotes subproblem diversity by penalizing items in a ``tabu'' set (e.g., those visited in recent LSNS iterations), in order to prevent cycling by repeatedly selecting a subproblem with high predicted improvement but no actual improvement. Equation~\eqref{ss_consistency} ensures consistency between element-wise and subset variables, by forcing $z_i=1$ for all elements contained in the subset $S$ whenever $q_S=1$.  The generic constraint \eqref{ss_validsp} defines a ``valid'' subproblem; in the the TDS-CW example, this constraint limits the cardinality of the subproblem to retain tractability at each iteration, and selects adjacent elements to enable effective adjustments, swaps and re-assignments in the solution.

\subsection{Application to the TDS-CW}

\subsubsection*{Subproblem.}

We define $K_1$ problem instances from historical data or random sampling. We define $K_2$ neighborhoods and optimize each one for each instance. We obtain $U=K_1K_2$ subproblems. Let $\calM^{u}, \calI^{u}, \calP^{u}, \calW^{u}, \calT^{u}$ store the orders, items, pods, workstations, and times in subproblem $u=1,\cdots,U$, and $\calA^{u}$ store the related time-space arcs. The subproblem is then defined as follows, with objective value $y_{u}$.
\begin{align} 
\max  & \displaystyle \quad \sum_{m \in \calM} \sum_{i \in \calI_m} \sum_{p \in \calP_i} \sum_{w \in \calW} x_{impwT}   \nonumber \\
    	\text{s.t   } 
    	&\displaystyle\quad  \text{Equations~\eqref{wsassign}--\eqref{binvars}} \nonumber \\
    	&\displaystyle\quad  y_{pa} = \overline{y}_{pa}, \ \forall (p,a)\ \text{s.t.}\ p \notin \calP^{\text{u}} \text{ or } a \notin \calA(\calW^{u}, \calT^{u})\\
    	&\displaystyle\quad  x_{impwt} = \overline{x}_{impwt}, \ \forall (i,m,p,w,t)\ \text{s.t.}\ m \notin \calM^{u},\ i \notin \calI^{u},\ p \notin  \calP^{u},\ w \notin \calW^{u}  \text{ or } t \notin \calT^{u}
\end{align}

\subsubsection*{Performance learning.}
We define seven feature categories: (i) size of each order ($\calS_f^{M} = \calM$); (ii) percentage of requested items on each pod ($\calS_f^{P} = \calP$); (iii) number and centrality of workstations ($\calS_f^{W} = \calW$); (iv) number of time periods ($\calS_f^{T} = \calT$); (v) item overlap and average inventory, for each order-pod pair ($\calS_f^{MP} = \calM\times\calP$); (vi) x-distance and y-distance, for each workstation-pod pair ($\calS_f^{PW} = \calP\times\calW$); and (vii) congestion at the workstation (``queue congestion'') in its immediate vicinity (``local congestion'') and in its broader vicinity (``semi-local congestion''), for each workstation-time pair 
($\calS_f^{WT} = \calW\times\calT$). By design, this list is restricted to one-dimensional elements and two-dimensional interactions to retain an effective learning structure and tractability in LSNS. Indeed, higher-dimensional combinations (e.g., order-item-pod-workstation combinations) would induce higher-dimensional decision variables in the subproblem generation formulation. 

Let use denote the feature values by $\alpha^{M}_{m,f}$, $\alpha^{P}_{p,f}$, $\alpha^{W}_{w,f}$, $\alpha^{T}_{t,f}$, $\alpha^{MP}_{m,p,f}$, $\alpha^{WP}_{w,p,f}$ and $\alpha^{WT}_{w,t,f}$, and the fitted coefficients by $\widehat\beta^{M}_f$, $\widehat\beta^{P}_f$, $\widehat\beta^{W}_f$, $\widehat\beta^{T}_f$, $\widehat\beta^{MP}_f$, $\widehat\beta^{WP}_f$ and $\widehat\beta^{WT}_f$. We obtained the best fit with the holistic regression approach from \cite{bertsimas2016or} to avoid overfitting as well as to retain global sparsity and local sparsity among correlated features (see Appendix~\ref{app:holistic} for details).

\subsubsection*{Subproblem generation.}
    
We seek a subproblem, characterized by $\calM^{SP}, \calI^{SP}, \calP^{SP}, \calW^{SP}$ and $\calT^{SP}$. We pre-process a set $\calB$ comprising all blocks of $N_{W}$ workstations and to $N_{T}$ contiguous time periods, stored in subsets $\calW_b$ and $\calT_b$ for block $b\in\calB$. We characterize the new neighborhood via binary variables $q^{MP}_{m,p}$, $q^{WP}_{w,p}$ and $q^{WT}_{w,t}$ to select order-pod, workstation-pod and workstation-time pairs, as well as binary variables $z^{B}_b$, $z^M_m$, and $z^P_p$ to select workstation-time blocks, orders and pods. The subproblem generation problem is then formulated as follows:
{\small\begin{align} 
 \max_{\bz,\bw}  \quad &  \displaystyle  \sum_{m \in \calM} \widehat\beta^{M}_f \alpha^{M}_{m,f} z^M_{m} + \sum_{p \in \calP} \widehat\beta^{P}_f \alpha^{P}_{p,f} z^P_{p} + \sum_{b \in \calB} \left( \sum_{w \in \calW_b} \widehat\beta^{W}_f \alpha^{W}_{w,f} + \sum_{t \in \calT_b} \widehat\beta^{T}_f \alpha^{T}_{t,f} \right) z_b^{B}\nonumber\\
 & + \sum_{b \in \calB} \sum_{w \in \calW_b} \sum_{t \in \calT_b}  \widehat\beta^{WT}_f \alpha^{WT}_{w,t,f} z^{B}_b + \sum_{m \in \calM} \sum_{p \in \calP}  \widehat\beta^{MP}_f \alpha^{MP}_{m,p,f} q^{MP}_{m,p}+  \sum_{w \in \calW} \sum_{p \in \calP}  \widehat\beta^{WP}_f \alpha^{WP}_{w,p,f} q^{WP}_{w,p}  \nonumber\\
 &  -  \sum_{w \in \calW} \sum_{t \in \calT} \left( \sum_{m \in \calM} \sum_{i \in \calI_m} \sum_{p \in \calP_i} \bar{x}_{mipwt} - \bar{x}_{mipw,t-1}\right) q^{WT}_{w,t} - C \sum_{b \in \calB_{tabu}} z^B_b \label{ss_obj2}\\
       %-------------------------------------------------------%
       \text{s.t.} \quad 
    	&\displaystyle %\textbf{Size and continuity requirements} \nonumber \\
       %-------------------------------------------------------%
       %& \displaystyle 
       \sum_{b \in \calB} z^{B}_b = 1 \label{eq:B1} \\
       & \displaystyle \sum_{m \in \calM} z^M_m = N_M \label{eq:M1} \\ 
       & \displaystyle \sum_{p \in \calP} z_p^P = N_P \label{eq:P1} \\
        %-------------------------------------------------------%
    	%&\displaystyle \textbf{Coverage and flexibility} \nonumber \\
       %-------------------------------------------------------%
       & \displaystyle z^M_m \leq \sum_{p \in \calP_i} z_p^P \quad \forall m \in \calM, i \in \calI_m   \label{eq:coverage} \\
       & \displaystyle z^M_m \geq z_b^{B} \quad \forall b \in \calB, w \in \calW_b, t \in \calT_b, m \in \calM^{\text{open}}_{wt} \label{eq:openorders} \\
       & \displaystyle z_p^P \geq z_b^{B} \quad \forall b \in \calB, w \in \calW_b, t \in \calT_b, p \in \calP^{\text{open}}_{wt} \label{eq:openpods} \\
       %-------------------------------------------------------%
    	&\displaystyle \textbf{Consistency constraints (omitted)}  \\
       %-------------------------------------------------------%
       %-------------------------------------------------------%
    	%&\displaystyle \textbf{Domain of definition} \nonumber \\
       %-------------------------------------------------------%
       & \bz,\ \bq,\ \br\ \text{binary}
\end{align}}

Equation~\eqref{ss_obj2} maximizes the predicted improvement at the workstations $\calW^{SP}$ during the times in $\calT^{SP}$, and incentivizes exploration. The first two lines capture the predicted throughput, using the linear learning function; the next term captures the throughput from the incumbent solution (characterized by the incumbent $\bar{x}_{mipwt}$ variables); the last term disincentivizes revisiting previously-explored components. The constraints define a valid subproblem via four requirements:
\begin{itemize}
    \item[--] Size: for tractability, a valid subproblem must comprise $N_{W}$ workstations and $N_{T}$ time periods (Constraint~\eqref{eq:B1}) as well as $N_M$ orders (Constraint~\eqref{eq:M1}) and $N_P$ pods (Constraint~\eqref{eq:P1}).
    \item[--] Time continuity: the selected time periods also have to be contiguous in time, in order to enable effective swaps and re-assignments in the subproblem (Constraint~\eqref{eq:B1}).
    \item[--] Coverage and flexibility: Constraint~\eqref{eq:coverage} ensures that an order can only selected if we select pods that contain the corresponding items. In addition, Constraints~\eqref{eq:openorders} and~\eqref{eq:openpods} select all orders and all pods that are open at the selected workstations during the selected time periods in the incumbent solution, in order to facilitate swaps and re-assignments.
    \item[--] Consistency constraints: we ensure consistency between the element-wise variables $z^{B}_b$, $z^M_m$, and $z^P_p$, and the pair-wise variables $q^{MP}_{m,p}$, $q^{WP}_{w,p}$ and $q^{WT}_{w,t}$ (Equation~\eqref{ss_consistency}). We also avoid selecting any order that is assigned to other workstations in the incumbent solution, or any pod that is open at another workstation during the selected time period. Due to the large size of the TDS-CW, this choice maintains global feasibility without inputting the full model.
\end{itemize}

This formulation enables to search over an exponential number of subproblems, without resorting to exhaustive enumeration. The number of decision variables scales in $\calO(|\calW|^{N_W}|\calT|+|\calM||\calP|+|\calW||\calP|)$, as opposed to $\calO(|\calW||\calI||\calM||\calW||\calT|)$ in the original TDS-CW. By design, this formulation retains computational efficiency and enables its implementation at each LSNS iteration.

\subsubsection*{Implementation.}

In our learn-then-optimize framework, prediction quality is most important for the best subproblems. However, the training data does not initially contain many subproblems with very high throughput. Therefore, we first train the machine learning model on randomized training data, and we then integrate the subproblem generation model into the training data generation process. Specifically, we perform many iterations of LSNS using the learn-then-optimize approach to expand the training dataset, and re-train the model on the augmented dataset. We repeat the process several times, alternating model training and additional data generation.

\subsubsection*{Decomposition.}\label{subsec:decomposition}

The LSNS algorithm, armed with our learn-then-optimize routine, scales to medium-sized instances. Yet, due to the complexity of the TDS-CW, the algorithm fails to solve full-sized instances. We therefore solve full-sized instances by partitioning the fulfillment center into $R$ regions. Each region $r \in \calR$ contains a distinct set of workstations $\calW^r$, pods $\calP^r$, storage locations $\calL^r$ and intersections $\calJ^r$, so all constraints (Equations~\eqref{wsassign}--\eqref{vlink}) are fully decomposable after each order has been assigned to one region.

We extend our learn-then-optimize approach to assign each order to one region $r$, via binary order-region assignment variables $x_{mr}$. The formulation maximizes pod-order synergies, using learned parameters from the training phase (Equation~\eqref{part_obj}). Constraints ensure that each order is assigned to one region (Equation~\eqref{part_assign}), that the region contains all required items (Equation~\eqref{part_feasible}), and that the numbers of items are balanced across regions within a tolerance $\xi$ (Equation~\eqref{part_balance}).
{\small\begin{align}
    \max_{\bz,\bw}  \quad &  \displaystyle \sum_{m \in \calM} \sum_{r \in \calR} \sum_{p \in \calP^r}  \widehat\beta^{MP}_f \alpha^{MP}_{m,p,f} x_{mp} \label{part_obj}   \\
    \st \quad & \sum_{r \in calR} x_{mr} = 1 \quad \forall m \in \calM \label{part_assign}\\
    & x_{mr} \leq |\calP_i \cap \calP^r| \quad \forall m \in \calM, i \in \calI_m, r \in \calR \label{part_feasible} \\
    & \sum_{m \in \calM} |\calI_m| x_{mr} \geq \frac{1}{R} \sum_{m \in \calM}|\calI_m| - \xi \quad \forall r \in \calR \label{part_balance} \\
    & x_{mr} \in \{0,1\} \quad \forall m \in \calM, r \in \calR \label{part_var}
\end{align}}

The optimal assignments define a set of orders for each region $r$, $\calM^r = \{ m \in \calM \:|\: x^*_{mr} = 1\}$. The TDS-CW is then solved independently and in parallel for each region, via our learn-then-optimize LSNS algorithm. By design, this procedure yields a globally feasible solution to the full problem.

\subsection{Comparison with LSNS benchmarks} \label{sec:benchmarks}

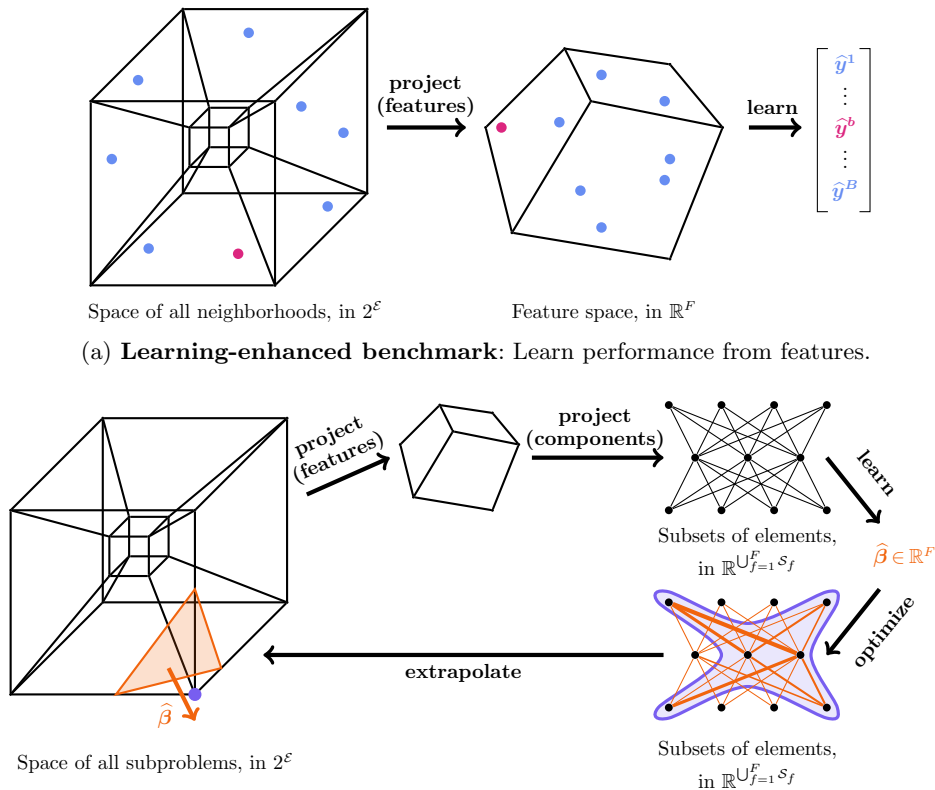
\begin{figure} %%
     \centering

        \subfloat[][\textbf{Learning-enhanced benchmark}: Learn  performance from features.]{\label{subfig:arclesstight1}
        \begin{tikzpicture}[scale=0.7,transform shape,text centered,
            roundnode/.style={circle,color=white,fill,inner sep=1.5pt},
            invisnode/.style={circle,color=black,draw=black,fill,inner sep=.00pt},
            pinknode/.style={circle,color=mypink,fill,inner sep=2pt},
            bluenode/.style={circle,color=mymidblue,fill,inner sep=2pt},
            blacknode/.style={circle,color=black,fill,inner sep=2pt}
            ]

            %Tesseract corners
            \node[invisnode] at (0,8.5)   (o1) {};
            \node[invisnode] at (1.75,10.25)   (o2) {};
            \node[invisnode] at (5.25,10.25)   (o3) {};
            \node[invisnode] at (3.5,8.5)   (o4) {};
            \node[invisnode] at (0,5)   (o5) {};
            \node[invisnode] at (1.75,6.75)   (o6) {};
            \node[invisnode] at (5.25,6.75)   (o7) {};
            \node[invisnode] at (3.5,5)   (o8) {};
    
            \node[invisnode] at (1.875,8)   (i1) {};
            \node[invisnode] at (2.25,8.375)   (i2) {};
            \node[invisnode] at (3,8.375)   (i3) {};
            \node[invisnode] at (2.625,8)   (i4) {};
            \node[invisnode] at (1.875,7.25)   (i5) {};
            \node[invisnode] at (2.25,7.625)   (i6) {};
            \node[invisnode] at (3,7.625)   (i7) {};
            \node[invisnode] at (2.625,7.25)   (i8) {};
    
            %Tesseract edges
            \draw[black, thick, -] (o1) -- (o2) ;
            \draw[black, thick, -] (o1) -- (o4) ;
            \draw[black, thick, -] (o1) -- (o5) ;
            \draw[black, thick, -] (o2) -- (o3) ;
            \draw[black, thick, -] (o2) -- (o6) ;
            \draw[black, thick, -] (o3) -- (o4) ;
            \draw[black, thick, -] (o3) -- (o7) ;
            \draw[black, thick, -] (o4) -- (o8) ;
            \draw[black, thick, -] (o5) -- (o6) ;
            \draw[black, thick, -] (o5) -- (o8) ;
            \draw[black, thick, -] (o6) -- (o7) ;
            \draw[black, thick, -] (o7) -- (o8) ;
    
            \draw[black, thick, -] (i1) -- (i2) ;
            \draw[black, thick, -] (i1) -- (i4) ;
            \draw[black, thick, -] (i1) -- (i5) ;
            \draw[black, thick, -] (i2) -- (i3) ;
            \draw[black, thick, -] (i2) -- (i6) ;
            \draw[black, thick, -] (i3) -- (i4) ;
            \draw[black, thick, -] (i3) -- (i7) ;
            \draw[black, thick, -] (i4) -- (i8) ;
            \draw[black, thick, -] (i5) -- (i6) ;
            \draw[black, thick, -] (i5) -- (i8) ;
            \draw[black, thick, -] (i6) -- (i7) ;
            \draw[black, thick, -] (i7) -- (i8) ;
    
            \draw[black, thick, -] (o1) -- (i1) ;
            \draw[black, thick, -] (o2) -- (i2) ;
            \draw[black, thick, -] (o3) -- (i3) ;
            \draw[black, thick, -] (o4) -- (i4) ;
            \draw[black, thick, -] (o5) -- (i5) ;
            \draw[black, thick, -] (o6) -- (i6) ;
            \draw[black, thick, -] (o7) -- (i7) ;
            \draw[black, thick, -] (o8) -- (i8) ;
    
            %Polyhedron corners
            \node[invisnode] at (9.5,8.5)   (p1) {};
            \node[invisnode] at (9,9.5)   (p2) {};
            \node[invisnode] at (11,9.2)   (p3) {};
            \node[invisnode] at (12,8)   (p4) {};
            \node[invisnode] at (8,6)   (p5) {};
            \node[invisnode] at (7.5,8)   (p6) {};
            \node[invisnode] at (11,5.5)   (p8) {};
    
            %Polyhedron edges
            \draw[black, thick, -] (p1) -- (p2) ;
            \draw[black, thick, -] (p1) -- (p4) ;
            \draw[black, thick, -] (p1) -- (p5) ;
            \draw[black, thick, -] (p2) -- (p3) ;
            \draw[black, thick, -] (p2) -- (p6) ;
            \draw[black, thick, -] (p3) -- (p4) ;
            \draw[black, thick, -] (p4) -- (p8) ;
            \draw[black, thick, -] (p5) -- (p6) ;
            \draw[black, thick, -] (p5) -- (p8) ;
    
            %Arrows 
            \draw[black, ultra thick, ->] (5.625,8) -- (7.125,8) ;
            \draw[black, ultra thick, ->] (12.5,8) -- (13.5,8) ;

            %Points (SP space)
            \node[bluenode] at (4.5,6.5)   (p1_tes) {};
            \node[bluenode] at (4,8.4)   (p2_tes) {};
            \node[bluenode] at (0.4,7.4)   (p3_tes) {};
            \node[bluenode] at (1.1,5.7)   (p4_tes) {};
            \node[pinknode] at (2.8,5.6)   (p5_tes) {};
            \node[bluenode] at (4.8,7.9)   (p6_tes) {};
            \node[bluenode] at (3.0,9.8)   (p7_tes) {};
            \node[bluenode] at (0.9,8.9)   (p8_tes) {};

            %Points (feature space)
            \node[bluenode] at (11,7.4)   (p1_feat) {};
            \node[bluenode] at (10.9,8.5)   (p2_feat) {};
            \node[bluenode] at (9.7,9)   (p3_feat) {};
            \node[bluenode] at (8.9,8.1)   (p4_feat) {};
            \node[bluenode] at (10.9,7)   (p5_feat) {};
            \node[pinknode] at (7.8,8.0)   (p6_feat) {};
            \node[bluenode] at (9.3,6.8)   (p7_feat) {};
            \node[bluenode] at (9.7,6.1)   (p8_feat) {};

            %Predictions
            \node[] at (14.3,9.2) {$\: \textcolor{mymidblue}{\boldsymbol{\widehat{y}^1}} $};
            \node[] at (14.3,8.7) {$\: \vdots $};
            \node[] at (14.3,8) {$\: \textcolor{mypink}{\boldsymbol{\widehat{y}^b}} $};
            \node[] at (14.3,7.5) {$\: \vdots $};
            \node[] at (14.3,6.8) {$\: \textcolor{mymidblue}{\boldsymbol{\widehat{y}^B}} $};
            \draw[black, thin, -] (14,9.5) -- (13.8,9.5) -- (13.8, 6.4) -- (14,6.4);
            \draw[black, thin, -] (14.6,9.5) -- (14.8,9.5) -- (14.8, 6.4) -- (14.6,6.4);
            
            %Labels
            \node[] at (6.36,8.6) {\textbf{\begin{tabular}{c}project\\(features)\end{tabular}}};
            \node[] at (12.95,8.4) {\textbf{learn}};
            \node[] at (2.75,4.5) {Space of all neighborhoods, in $2^{\calE}$};
            \node[] at (9.75,4.5) {Feature space, in $\mathbb{R}^{F}$};
        
        \end{tikzpicture}
         }

         \subfloat[][\textbf{Learn-then-optimize}: Project features into lower-dimensional space (e.g., graph of pair-wise elements); learn contribution of each element subset to each feature; construct neighborhood by extrapolating the prediction function into the full-dimensional space.]{\label{subfig:arclesstight2}
        \begin{tikzpicture}[scale=0.7,transform shape,
            roundnode/.style={circle,color=black,fill,inner sep=1.5pt},
            invisnode/.style={circle,color=black,draw=black,fill,inner sep=.00pt},
            objnode/.style={circle,color=mymidorange,fill,inner sep=1.5pt},
            purplenode/.style={circle,color=mypurple,fill,inner sep=2.5pt},
            ]
    
            %Tesseract corners
            \node[invisnode] at (0,8.5)   (o1) {};
            \node[invisnode] at (1.75,10.25)   (o2) {};
            \node[invisnode] at (5.25,10.25)   (o3) {};
            \node[invisnode] at (3.5,8.5)   (o4) {};
            \node[invisnode] at (0,5)   (o5) {};
            \node[invisnode] at (1.75,6.75)   (o6) {};
            \node[invisnode] at (5.25,6.75)   (o7) {};
            \node[invisnode] at (3.5,5)   (o8) {};
    
            \node[invisnode] at (1.875,8)   (i1) {};
            \node[invisnode] at (2.25,8.375)   (i2) {};
            \node[invisnode] at (3,8.375)   (i3) {};
            \node[invisnode] at (2.625,8)   (i4) {};
            \node[invisnode] at (1.875,7.25)   (i5) {};
            \node[invisnode] at (2.25,7.625)   (i6) {};
            \node[invisnode] at (3,7.625)   (i7) {};
            \node[invisnode] at (2.625,7.25)   (i8) {};
    
            %Tesseract edges
            \draw[black, thick, -] (o1) -- (o2) ;
            \draw[black, thick, -] (o1) -- (o4) ;
            \draw[black, thick, -] (o1) -- (o5) ;
            \draw[black, thick, -] (o2) -- (o3) ;
            \draw[black, thick, -] (o2) -- (o6) ;
            \draw[black, thick, -] (o3) -- (o4) ;
            \draw[black, thick, -] (o3) -- (o7) ;
            \draw[black, thick, -] (o4) -- (o8) ;
            \draw[black, thick, -] (o5) -- (o6) ;
            \draw[black, thick, -] (o5) -- (o8) ;
            \draw[black, thick, -] (o6) -- (o7) ;
            \draw[black, thick, -] (o7) -- (o8) ;
    
            \draw[black, thick, -] (i1) -- (i2) ;
            \draw[black, thick, -] (i1) -- (i4) ;
            \draw[black, thick, -] (i1) -- (i5) ;
            \draw[black, thick, -] (i2) -- (i3) ;
            \draw[black, thick, -] (i2) -- (i6) ;
            \draw[black, thick, -] (i3) -- (i4) ;
            \draw[black, thick, -] (i3) -- (i7) ;
            \draw[black, thick, -] (i4) -- (i8) ;
            \draw[black, thick, -] (i5) -- (i6) ;
            \draw[black, thick, -] (i5) -- (i8) ;
            \draw[black, thick, -] (i6) -- (i7) ;
            \draw[black, thick, -] (i7) -- (i8) ;
    
            \draw[black, thick, -] (o1) -- (i1) ;
            \draw[black, thick, -] (o2) -- (i2) ;
            \draw[black, thick, -] (o3) -- (i3) ;
            \draw[black, thick, -] (o4) -- (i4) ;
            \draw[black, thick, -] (o5) -- (i5) ;
            \draw[black, thick, -] (o6) -- (i6) ;
            \draw[black, thick, -] (o7) -- (i7) ;
            \draw[black, thick, -] (o8) -- (i8) ;

            %Funky purple shape
            \draw [draw=mypurple, very thick, fill=mypurple, fill opacity=0.15] plot [smooth cycle] coordinates {(12.5,6.95) (14,6.35) (15.5,6.95) (15.7,6.75) (15.2,5.75) (15.7,4.75) (15.5,4.55) (14,5.15) (12.5,4.55) (12.3,4.75) (13.5,5.75) (12.3,6.75)};
    
             %Polyhedron corners
            \node[invisnode] at (8.4,10)   (p1) {};
            \node[invisnode] at (8.15,10.5)   (p2) {};
            \node[invisnode] at (9.15,10.35)   (p3) {};
            \node[invisnode] at (9.65,9.75)   (p4) {};
            \node[invisnode] at (7.65,8.75)   (p5) {};
            \node[invisnode] at (7.4,9.75)   (p6) {};
            \node[invisnode] at (9.15,8.5)   (p8) {};
            
            %Polyhedron edges
            \draw[black, thick, -] (p1) -- (p2) ;
            \draw[black, thick, -] (p1) -- (p4) ;
            \draw[black, thick, -] (p1) -- (p5) ;
            \draw[black, thick, -] (p2) -- (p3) ;
            \draw[black, thick, -] (p2) -- (p6) ;
            \draw[black, thick, -] (p3) -- (p4) ;
            \draw[black, thick, -] (p4) -- (p8) ;
            \draw[black, thick, -] (p5) -- (p6) ;
            \draw[black, thick, -] (p5) -- (p8) ;
            
            %Graph nodes
            \foreach \x in {2,...,5} 
                \node[roundnode] at (1*\x+10.5,10.5)   (top\x) {};
            \foreach \x in {2,...,4} 
                \node[roundnode] at (1*\x+11,9.5)   (mid\x) {};
            \foreach \x in {2,...,5} 
                \node[roundnode] at (1*\x+10.5,8.5)   (bot\x) {};
            %\foreach \x in {2,...,5} 
            %    \node[roundnode] at (1*\x+6.5,10.75)   (top\x) {};
            %\foreach \x in {2,...,4} 
            %    \node[roundnode] at (1*\x+7,9.5)   (mid\x) {};
            %\foreach \x in {2,...,5} 
            %    \node[roundnode] at (1*\x+6.5,8.25)   (bot\x) {};   

            %Graph edges
            \foreach \x in {2,...,5}, \foreach \y in {2,...,4} 
                \draw[black, thin, -] (top\x) -- (mid\y) ;
            \foreach \x in {2,...,4}, \foreach \y in {2,...,5} 
                \draw[black, thin, -] (mid\x) -- (bot\y) ;

             %Graph nodes
            \foreach \x in {2,...,5} 
                \node[roundnode] at (1*\x+10.5,6.75)   (l_top\x) {};
            \foreach \x in {2,...,4} 
                \node[roundnode] at (1*\x+11,5.75)   (l_mid\x) {};
            \foreach \x in {2,...,5} 
                \node[roundnode] at (1*\x+10.5,4.75)   (l_bot\x) {};

            %Graph edges
            %\draw[mymidorange, ultra thin, -] (l_top1) -- (l_mid1) ;
            %\draw[mymidorange, very thin, -] (l_top1) -- (l_mid2) ;
            %\draw[mymidorange, thick, -] (l_top1) -- (l_mid3) ;
            %\draw[mymidorange, very thin, -] (l_top1) -- (l_mid4) ;
            %\draw[mymidorange, very thin, -] (l_top2) -- (l_mid1) ;
            \draw[mymidorange, thin, -] (l_top2) -- (l_mid2) ;
            \draw[mymidorange, very thick, -] (l_top2) -- (l_mid3) ;
            \draw[mymidorange, ultra thick, -] (l_top2) -- (l_mid4) ;
            %\draw[mymidorange, very thin, -] (l_top3) -- (l_mid1) ;
            \draw[mymidorange, ultra thin, -] (l_top3) -- (l_mid2) ;
            \draw[mymidorange, very thin, -] (l_top3) -- (l_mid3) ;
            \draw[mymidorange, very thin, -] (l_top3) -- (l_mid4) ;
            %\draw[mymidorange, ultra thin, -] (l_top4) -- (l_mid1) ;
            \draw[mymidorange, ultra thin, -] (l_top4) -- (l_mid2) ;
            \draw[mymidorange, very thin, -] (l_top4) -- (l_mid3) ;
            \draw[mymidorange, very thin, -] (l_top4) -- (l_mid4) ;
            %\draw[mymidorange, very thin, -] (l_top5) -- (l_mid1) ;
            \draw[mymidorange, very thin, -] (l_top5) -- (l_mid2) ;
            \draw[mymidorange, thick, -] (l_top5) -- (l_mid3) ;
            \draw[mymidorange, semithick, -] (l_top5) -- (l_mid4) ;

            %\draw[mymidorange, very thin, -] (l_bot1) -- (l_mid1) ;
            %\draw[mymidorange, thin, -] (l_bot1) -- (l_mid2) ;
            %\draw[mymidorange, thin, -] (l_bot1) -- (l_mid3) ;
            %\draw[mymidorange, thick, -] (l_bot1) -- (l_mid4) ;
            %\draw[mymidorange, very thick, -] (l_bot2) -- (l_mid1) ;
            \draw[mymidorange, very thin, -] (l_bot2) -- (l_mid2) ;
            \draw[mymidorange, ultra thin, -] (l_bot2) -- (l_mid3) ;
            \draw[mymidorange, very thick, -] (l_bot2) -- (l_mid4) ;
            %\draw[mymidorange, ultra thin, -] (l_bot3) -- (l_mid1) ;
            \draw[mymidorange, ultra thin, -] (l_bot3) -- (l_mid2) ;
            \draw[mymidorange, very thin, -] (l_bot3) -- (l_mid3) ;
            \draw[mymidorange, ultra thin, -] (l_bot3) -- (l_mid4) ;
            %\draw[mymidorange, ultra thin, -] (l_bot4) -- (l_mid1) ;
            \draw[mymidorange, very thin, -] (l_bot4) -- (l_mid2) ;
            \draw[mymidorange, ultra thin, -] (l_bot4) -- (l_mid3) ;
            \draw[mymidorange, thin, -] (l_bot4) -- (l_mid4) ;
            %\draw[mymidorange, very thin, -] (l_bot5) -- (l_mid1) ;
            \draw[mymidorange, semithick, -] (l_bot5) -- (l_mid2) ;
            \draw[mymidorange, thick, -] (l_bot5) -- (l_mid3) ;
            \draw[mymidorange, thick, -] (l_bot5) -- (l_mid4) ;
            
            %Arrows 
            \draw[black, ultra thick, ->] (5.625,8.8) -- (7.2,9.5) ;
            \draw[black, ultra thick, ->] (9.9,9.5) -- (12.4,9.5);
            \draw[black, ultra thick, ->] (15.5,9.5) -- (16.5,8.25) ;
            
            \draw[black, ultra thick, ->] (16.5,7) -- (15.5,5.75);
            \draw[black, ultra thick, ->] (12.4,5.75) -- (4.8,5.75);

            %Labels
            \node[anchor=west] at (16.25,7.65) {\textcolor{mymidorange}
            {$\boldsymbol{\widehat{\beta}} \in \mathbb{R}^{F}$}};
            \node[] at (11.1,10.05) {\textbf{\begin{tabular}{c}project\\(components)\end{tabular}}};
            \begin{scope} [rotate=24]
                \node[transform shape] at (9.55, 6.35) {\textbf{\begin{tabular}{c}project\\(features)\end{tabular}}};
            \end{scope}
            \node[] at (8.6,5.4) {\textbf{extrapolate}};
            \begin{scope} [rotate=-51]
                \node[transform shape] at (3.2,18.6) {\textbf{learn}};
            \end{scope}
            \begin{scope} [rotate=51]
                \node[transform shape] at (15.25,-9) {\textbf{optimize}};
            \end{scope}
            \node[] at (2.75,3.7) {Space of all subproblems, in $2^{\calE}$};
            \node[] at (14,8) {Subsets of elements,};
            \node[] at (14,7.5) {in $\R^{\bigcup_{f=1}^F\calS_f}$};
            \node[] at (14,3.95) {Subsets of elements,};
            \node[] at (14,3.45) {in $\R^{\bigcup_{f=1}^F\calS_f}$};
            
            %Objective           
            \filldraw[mymidorange, thick, fill=mymidorange, fill opacity=0.2, -] (2,5) node[anchor=north]{}
              -- (3.5,7) node[anchor=north]{}
              -- (4,5.5) node[anchor=south]{}
              -- cycle;
            \draw[mymidorange, ultra thick, ->] (3,5.5) -- (3.5,4.5);
            \node[anchor=east] at (3.2,4.6) {\textcolor{mymidorange}{$\boldsymbol{\widehat{\beta}}$}};

            %Optimal solution
            \node[purplenode] at (3.5,5)   (opt) {};
               
        \end{tikzpicture}
         }   
    \caption{Learning-enhanced subproblem selection vs. learn-then-optimize subproblem generation.}
    \label{fig:staticvsdynamic}
    \vspace{-12pt}
\end{figure}

Figure~\ref{fig:staticvsdynamic} provides a schematic comparison of the learning-enhanced benchmark and our learn-then-optimize approach. Both methods seek a subproblem from a high-dimensional space, which grows exponentially with the number of elements in the problem. Both rely on a projection operator in the learning phase, by leveraging a low-dimensional feature-based representation of the local objective. They differ, however, in the process used to then select or generate a new subproblem:
\begin{itemize}
    \item[--] The learning-enhanced benchmark relies on enumeration to \textit{select} a neighborhood, by comparing all candidate neighborhoods directly in the lower-dimensional space. Computationally, its main bottleneck is the need to compute each feature for each candidate neighborhood; given that we consider pair-wise features, this involves $\calO(k^2)$ computations for each of ${\binom{n}{k}} =\calO(n^k)$ candidate neighborhoods of cardinality $k$. To retain tractability, this benchmark restricts the search to a budget of $B$ candidate neighborhoods out of $\calO(n^k)$ candidates.
    \item[--] Our learn-then-optimize method uses another projection operator and an optimization model to \textit{generate} a subproblem. The projection step further decomposes the local objective as a function of features of low-dimensional subsets of elements---by expressing the neighborhood features $x_{uf}$ as a linear combination of the $\alpha_{S,f}$ features in Equation~\eqref{eq:project}. The prediction function can then be lifted into the full-dimensional space of all neighborhoods (Equation~\eqref{eq:linear}). Finally, the subproblem generation problem avoids exhaustive enumeration thanks to a binary linear optimization formulation, enabled by the representation of each neighborhood as a collection of elements and by our restriction to linear predictive models.
\end{itemize}

These approaches involve a trade-off between approximation vs. optimization. The learning-enhanced benchmark can apply any machine learning method with any neighborhood features, leading to a smaller approximation error (i.e., stronger statistical power); but it is restricted to $B$ candidate neighborhoods, leading to a higher optimization error. Vice versa, our learn-then-optimize approach considers a linear prediction function with features of element subsets, leading to a higher approximation error; but it leverages binary optimization to select a neighborhood over the full set of candidates, leading to a smaller optimization error (i.e., stronger generative power).

\paragraph{Optimization error.} Consider $N$ neighborhoods, where $N$ grows exponentially with the problem size. The learn-then-optimize approach leads to an expected improvement of $\E [\max_{i=1,\ldots, N} X_i]$, whereas the learning-enhanced benchmark takes the maximum over a random subset of $B$ neighborhoods. Consider the case where the true improvements in neighborhoods $i=1,\cdots,N$ follow independent Poisson random variables $X_i$ with mean $\lambda$. Whereas this assumption is obviously a rough simplification of reality, it is nonetheless guided by our computational experiments, in which the improvements exhibit a long tail characteristic of the Poisson distribution. The expected value of the maximum of $k$ independent Poisson random variables is $\log{k} / W(\log{k} / e \lambda)$, where $W(\cdot)$ is the Lambert W function and grows logarithmically in $k$ \citep{briggspoisson}. Thus, the learning-enhanced benchmark induces an optimization loss of $\calO(\log_{N}{B})$:
\begin{align*}
\frac{\mathbb{E} [\max_{i = 1, \ldots, B} X_i] }{\mathbb{E} [\max_{i = 1, \ldots, N} X_i]} = \frac{\frac{\log{B}}{W(\log{B} / e \lambda)}}{\frac{\log{N}}{W(\log{N} / e \lambda)}} = \calO \left( \frac{\log{B}}{\log{N}} \cdot \frac{\log(\log{N} / e \lambda)}{\log(\log{B} / e \lambda)} \right) = \calO \left( \log_{N}{B} \right).
\end{align*}

\paragraph{Approximation error.}

In fact, the ``true'' improvement of each neighborhood is unknown, and estimated via a machine learning model. Our learn-then-optimize approach, by design, considers a prediction function within a restricted class $\calF$ (in our case, linear functions over features that are decomposable across element subsets). In contrast, the learning-enhanced benchmark may consider a larger function class $\calG$ (e.g., non-linear functions with other neighborhood features). Since $\calF \subseteq \calG$, the approximation error of the learn-then-optimize approach, denoted by $\varepsilon_{\text{app}}(\calF)$, is smaller than the one of the learning-enhanced benchmark, denoted by $\varepsilon_{\text{app}}(\calG)$: $\varepsilon_{\text{app}}(\calF)\leq\varepsilon_{\text{app}}(\calG).$

\begin{figure} %%
\centering
\begin{tikzpicture}[scale=0.75,transform shape]
    %Image file
    \node[inner sep=0pt] (matr) at (0,0)
        {\includegraphics[width=15cm]{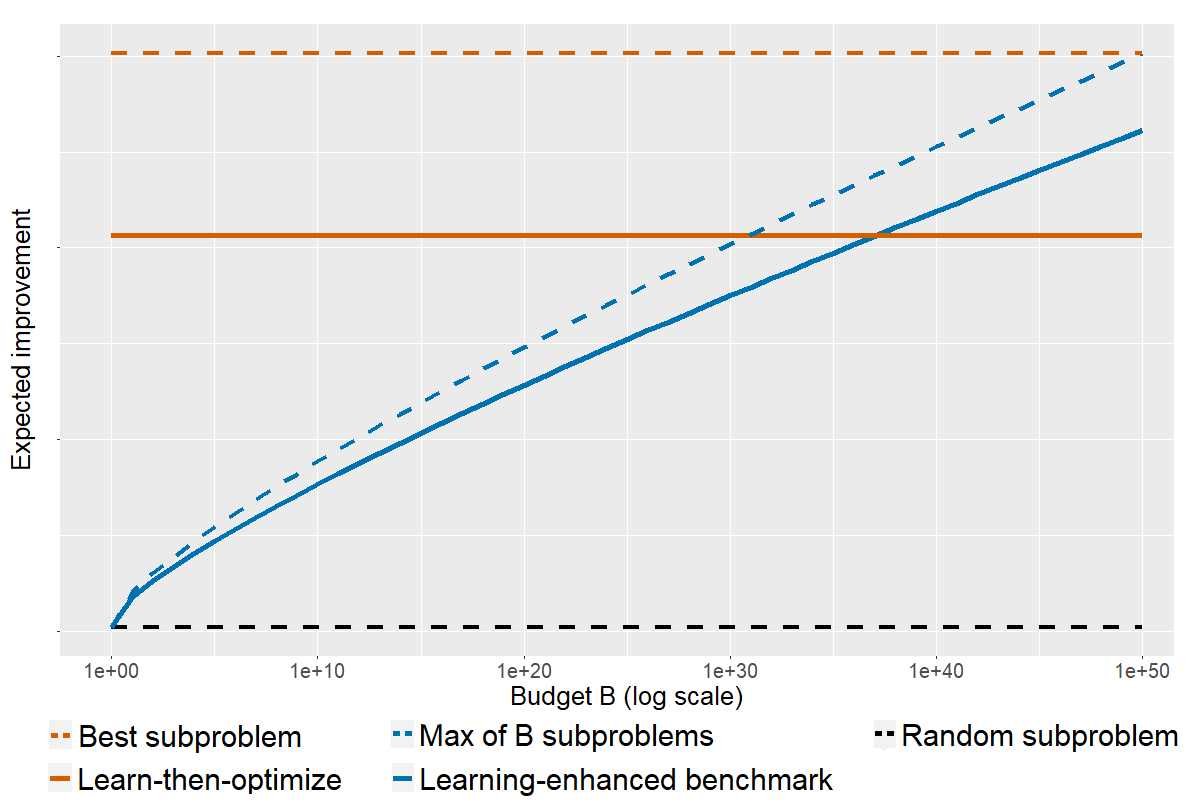}};
    
    %Text
    \node[text=mainblue] at (8.3,3.83) {\large $\varepsilon_{\text{app}}(\calG)$};
    \node[text=mainorange] at (10.7,3.1) {\large $\varepsilon_{\text{app}}(\calF)$};
    \node[text=black] at (7.6,-2.8) {\large $\lambda $};

    %Braces
    \draw [very thick, color=mainblue,decorate,decoration={brace,amplitude=5pt,mirror},xshift=0pt,yshift=0pt] (7,3.3) -- (7,4.35); 
    \draw [very thick, color=mainorange,decorate,decoration={brace,amplitude=6pt,mirror},xshift=0pt,yshift=0pt] (9.3,1.9) -- (9.3,4.4);

    %White text to center the figure
    \node[text=white] at (-9.2,1) {widen};
   
\end{tikzpicture}
\caption{Trade-off between optimization error and approximation error, with $10^{50}$ candidate subproblems.}
\label{fig:poisson}
\end{figure}
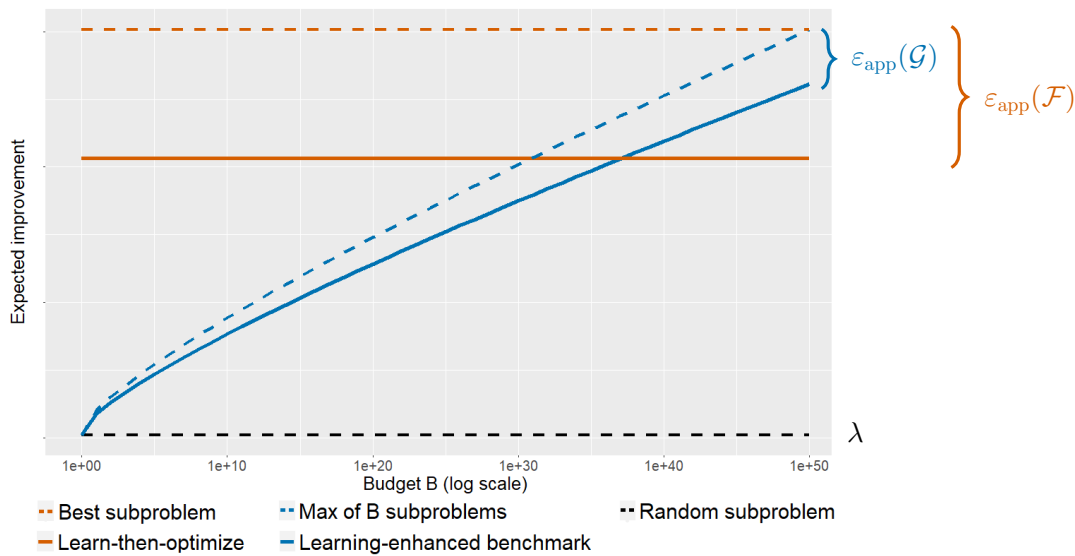

\paragraph{Visualization.} This trade-off is illustrated in Figure~\ref{fig:poisson} with $N=10^{50}$ possible subproblems and a mean improvement of $\lambda=0.24$ (inspired by our experiments). At one extreme, randomized LSNS yields an expected improvement of 0.24 (dashed black line). At the other extreme, the best neighborhood yields an expected improvement of 30 in the absence of approximation error (dashed orange line). The learning-enhanced benchmark yields an expected improvement of 5 with $B=10^{4}$, and of 20 with $B=10^{30}$ (dashed blue line). So, even with a much larger budget $B$ than can be tractably handled, the benchmark may still exhibits a significant optimization loss. On the other hand, our learn-then-optimize approach features a higher approximation error than the learning-enhanced benchmark (difference between the solid and dashed lines). Nonetheless, our computational results that our learn-then-optimize approach can still generate strong solution improvements, with significant performance improvements against the learning-enhanced benchmark.

\section{Experimental results}
\label{sec:results}

We developed problem instances in collaboration with our industry partner to replicate large-scale warehousing operations. We define instances with a time discretization of 30 seconds and a planning horizon of 15 minutes (we support the choice of a 30-second discretization in Appendix~\ref{subsec:discretization}). To train our machine learning model, we created a labeled dataset comprising 10,000 to 100,000 subproblems comprising a small subset of workstations, time periods, pods and orders (we report results to guide this calibration in Appendix~\ref{subsec:subproblem}). We tuned all hyper parameters of the holistic linear regression model using $k$-fold cross validation. Armed with our learn-then-optimize approach, we tested the resulting LSNS algorithm on 20 out-of-sample optimization instances. All models are solved with Gurobi v10.0, using the optimization package JuMP in Julia \citep{dunning2017jump}. All runs were performed on a laptop equipped with a 6-core Intel i7-10750H CPU (2.6 GHz) and 32 GB RAM, with an eight-hour limit. Source code and data are made available online.\footnote{https://github.com/alexschmid3/warehouse-task-assignment}

\subsection{Benefits of integrated task design and scheduling} \label{sec:decomp}

A critical feature of the TDS-CW is the integration of task design (order-workstation and item-pod assignments) and task scheduling (pod flows and item processing at the workstations). Due to the scale of these problems, a common approach consists of breaking the problem down into sequential decisions. Accordingly, we evaluate our integrated approach against two decomposition benchmarks, detailed in Appendix~\ref{app:decomp}: (i) a two-stage decomposition that first solves order-item-pod-workstation assignments, and then pod flows and workstation scheduling decisions---mirroring the approach from \cite{allgor2023algorithm}; and (ii) a three-stage decomposition that further breaks down the first phase, by solving order-workstation assignments, then order-item-pod assignments, and finally pod flows and workstation scheduling decisions---mirroring the approach from \cite{boysen2017parts}. Table~\ref{tab:decomp} compares these benchmarks to our greedy heuristic, described in Appendix~\ref{app:greedy}.

\begin{table} %%
\caption{Greedy heuristic vs. two-stage, three-stage decomposition (``2-decomposition'', ``3-decomposition'').}
\label{tab:decomp}
\centering\footnotesize\renewcommand{\arraystretch}{1}
\begin{tabular}{llcccccccccccc} \toprule
 &  & \multicolumn{4}{c}{Tight Congestion} & \multicolumn{4}{c}{Moderate Congestion}  & \multicolumn{4}{c}{No Congestion}  \\
 \cmidrule(lr){3-6}\cmidrule(lr){7-10}\cmidrule(lr){11-14} 
Instance & Method & \multicolumn{2}{c}{Objective} & Util & CPU & \multicolumn{2}{c}{Objective} & Util & CPU & \multicolumn{2}{c}{Objective} & Util & CPU \\
\midrule
Small  
 & 2-decomposition & 76 & (-40\%) & 32\% & n/a & 84 & (-49\%) & 35\% & n/a & 166 & (-25\%) & 72\% & n/a \\
 & 3-decomposition & 80 & (-37\%) & 33\% & n/a & 75 & (-54\%) & 32\% & n/a & 63 & (-79\%) & 27\% & n/a \\
 & \textbf{Greedy} & 127 & (base) & 57\% & 1 min. & 165 & (base) & 75\% & 1 min. & 222 & (base) & 96\% & 1 min.\\
 \midrule
Medium  
& 2-decomposition & 144 & (-40\%) & 45\% & n/a & 155 & (-42\%) & 50\% & n/a & 183 & (-36\%) & 63\% & n/a \\
 & 3-decomposition & 113 & (-52\%) & 35\% & n/a & 165 & (-38\%) & 53\% & n/a & 208 & (-27\%) & 71\% & n/a \\
 & \textbf{Greedy} & 239 & (base) & 81\% & 3 min. & 268 & (base) & 91\% & 3 min. & 284 & (base) & 97\% & 1 min. \\
\bottomrule
\end{tabular}
\begin{tablenotes}\linespread{1}
%\vspace{-6pt}
\item ``n/a'' indicates that the algorithm does not terminate within a two-day time limit.
\end{tablenotes}
\vspace{-12pt}
\end{table}

Note that the decomposition approaches fail to find high quality solutions because they ignore crucial constraints in early stages. Specifically, the TDS-CW includes tight constraints on the number of orders open at each workstation and on warehouse congestion, neither of which is captured in earlier studies where decomposition was successful \citep{allgor2023algorithm,boysen2017parts}. Under these additional constraints, our greedy heuristic outperforms decomposition benchmarks in terms of both solution quality and computational time. When congestion is not a bottleneck, the greedy heuristic achieves 96-97\% workstation utilization within just a minute of solve time, versus up to 72\% utilization within two days of computation for the benchmarks. Moreover, the problem becomes increasingly harder in the presence of congestion, due to resulting interdependencies across task design and scheduling decisions. Both benchmarks lead to a significant drop in utilization by ignoring the impact of upstream task design decisions on downstream pod flows and sequencing dynamics. In comparison, the greedy heuristic results in higher throughput, oftentimes by a wide margin, in much shorter computational times (minutes versus days). Nonetheless, the greedy heuristic still results in limited utilization with congestion, motivating our LSNS algorithm.

\subsection{Performance of learn-then-optimize algorithm}

We now evaluate our learn-then-optimize approach to LSNS against three benchmarks:
\begin{itemize}[itemsep=0pt,topsep=0pt]
    \item[--] Random sampling: pick a subproblem uniformly at random out of all candidates that satisfy the computational and practical requirements (Equation~\eqref{ss_validsp}). In the TDS-CW, we randomly select a contiguous workstation-time block from the set $\calB$, select orders and pods that are already assigned to that block, and then randomly select additional orders and pods.
    \item[--] Domain-based heuristic: pick a subproblem with ``synergistic'' elements. This benchmark ignores the learning component, but relies on proxies and heuristics to devise a feasible neighborhood in the subproblem generation phase (see Appendix~\ref{app:synergy} for details).
    %To prevent ``leakage'' from our results, we developed this heuristic before our learn-then-optimize approach.
    \item[--] Learning-enhanced LSNS: apply the machine learning model to select the subproblem with the strongest predicted improvement out of $B$ candidates, described by sets $\calS^1,\cdots,\calS^B$:
    $$b^*\in\argmax_{b=1,\cdots,B}\left(\widehat{f} \left( \left\{ \sum_{S\in\calS_f\cap\calS^b}\alpha_{S,f} \:|\: f=1,\cdots,F\right\} \right) -  \sum_{S \in \calS^{OPT}\cap\calS^b} c_S \overline{v}_S\right)$$
    When $B=1$, this benchmark reduces to random sampling. When $B=N$, it is equivalent to our approach, albeit with more learning flexibility because it does not need to rely on linear models. In-between, it leverages the learning component but not the optimization component.
\end{itemize}

We run all methods both with an empty solution (``cold start'') and our greedy solution as a starting point (``warm start''), and return the best solution. Table~\ref{tab:bigtable} reports, for each method, the throughput, computational times, average workstation utilization, and average numbers of items picked per pod. The table also reports the greedy solution from Table~\ref{tab:decomp} as a baseline.

\begin{table} %%
\caption{Performance comparison of the learn-then-optimize LSNS algorithm to LSNS benchmarks.}
\label{tab:bigtable}
\centering
\scriptsize
\renewcommand{\arraystretch}{1}
\begin{tabular}{lllcccccccc} \toprule
 &  &  &  &  &  & \multicolumn{3}{c}{Time per iteration \scriptsize{(s)}} &  \multicolumn{2}{c}{Solution}  \\ \cmidrule(lr){7-9}\cmidrule(lr){10-11}
  &  &  &  \multicolumn{2}{c}{Objective} & Full solve & Predict/ & SP & Feat/sol & Util & Picks \\ 
Instance & Congestion & Method & \multicolumn{2}{c}{(throughput)} & time \scriptsize{(s)} & select & re-opt. & updates & \% & per pod \\
 \midrule \midrule
 Small & Moderate & Greedy & 165 & (-33\%) & 84 & --- & --- & --- & 75\% & 1.61 \\
 &  & Random & 203 & (-11\%) & 172 & $<$1 & 1 & 2 & 85\% & 1.78 \\
 &  & Domain-based & 228 & (-8\%) & 244 & 1 & 2 & 3 & 93\% & 1.96 \\
 &  & LE ($B=100$) & 196 & (-38\%) & 2,625 & 81 & 2 & 2 & 84\% & 1.67 \\
 &  & LE ($B=1000$) & 174 & (-46\%) & 7,200+ & 257 & 0 & 1 & 78\% & 1.51 \\
 &  & \textbf{LTO} & \textbf{247} & (base) & \textbf{338} & \textbf{6} & \textbf{4} & \textbf{2} & \textbf{95\%} & \textbf{2.30} \\
  \cmidrule(lr){2-11} 
 & Tight & Greedy & 127 & (-45\%) & 85 & --- & --- & --- & 57\% & 1.56 \\
 &  &  Random & 180 & (-22\%) & 87 & $<$1 & 1 & 2 & 73\% & 1.94 \\
 &  & Domain-based & 199 & (-14\%) & 237 & 1 & 2 & 2 & 79\% & 2.14 \\
 &  & LE ($B=100$) & 144 & (-21\%) & 2,546 & 80 & 1 & 2 & 62\% & 1.65 \\
 &  & LE ($B=1000$) & 136 & (-30\%) & 7,200+ & 246 & 1 & 1 & 59\% & 1.60 \\
 &  & \textbf{LTO} & \textbf{231} & (base) & \textbf{333} & \textbf{6} & \textbf{3} & \textbf{2} & \textbf{86\%} & \textbf{2.62} \\
\midrule
Medium & Moderate & Greedy & 268 & (-17\%) & 165 & --- & --- & --- & 91\% & 1.20 \\
 &  & Random & 279 & (-13\%) & 309 & $<$1 & 1 & 7 & 91\% & 1.22 \\
 &  & Domain-based & 309 & (-4\%) & 467 & 1 & 1 & 8 & 94\% & 1.43 \\
 &  & LE ($B=100$) & 272 & (-16\%) & 7,200+ & 240 & 1 & 4 & 91\% & 1.16 \\
 &  & LE ($B=1000$) & 269 & (-17\%) & 7,200+ & 334 & 1 & 1 & 91\% & 1.13 \\
 &  &  \textbf{LTO} & \textbf{323} & (base) & \textbf{609} & \textbf{14} & \textbf{2} & \textbf{7} & \textbf{91\%} & \textbf{1.80} \\
  \cmidrule(lr){2-11}
 & Tight & Greedy & 239 & (-24\%) & 189 & --- & --- & --- & 81\% & 1.20 \\
 &  & Random & 258 & (-17\%) & 287 & $<$1 & 1 & 7 & 83\% & 1.25 \\
 &  & Domain-based & 292 & (-6\%) & 467 & 1 & 1 & 14 & 86\% & 1.56 \\
 &  & LE ($B=100$) & 247 & (-21\%) & 7,200+ & 240 & 1 & 4 & 82\% & 1.16 \\
 &  & LE ($B=1000$) & 241 & (-23\%) & 7,200+ & 362 & 1 & 1 & 81\% & 1.14 \\
 &  & \textbf{LTO} & \textbf{314} & (base) & \textbf{500} & \textbf{13} & \textbf{2} & \textbf{5} & \textbf{89\%} & \textbf{1.84} \\
\bottomrule
\end{tabular}
\begin{tablenotes}\linespread{1}
\item LE: Learning-enhanced LSNS given subproblem budget $B$. 
\item Runs capped at 2 hours, 50 LSNS iterations, or 5 iterations without improvement.
\end{tablenotes}
\end{table}

Note, first, that the domain-based heuristic enables significant improvements to the LSNS algorithm, as compared to the greedy heuristic and randomized LSNS. In the medium instance with tight congestion, for example, the LSNS algorithm with randomized subproblem selection achieves a throughput of 258 items---an 8\% improvement from the greedy heuristic; with the domain-based heuristic for subproblem selection, LSNS further increases throughput to 292 items---another 13\% improvement. These results underscore the impact of the LSNS algorithm itself, as well as the importance of guiding the search toward ``synergistic'' neighborhoods. Both learning-based methods (i.e., the learning-enhanced LSNS benchmark and our learn-then-optimize LSNS algorithm) aim to achieve that goal by using machine learning results to identify ``synergistic'' neighborhoods.

However, the learning-enhanced LSNS benchmark deteriorates the solution from the domain-based heuristic and the randomized benchmark. This can be surprising, since this approach involves machine learning rather than randomization to guide subproblem selection. However, computational requirements in the learning-enhanced benchmark become prohibitively expensive, so the algorithm times out after only a few LSNS iterations. This limitations become even stronger with a larger budget of subproblems. For example, increasing $B$ from 100 to 1,000 increases computational times per iteration by a factor of 3 for marginal gains in improvement at each iteration, resulting in timeouts. In the medium instances, the learning-enhanced method with $B=1,000$ does not even improve on the initial solution within the eight-hour limit. These results suggest that learning-enhanced LSNS may not necessarily provide low-hanging fruits toward accelerating LSNS algorithms when the overall number of candidate subproblems is very large.

\begin{figure} %%
    \centering
  \includegraphics[width=0.5\textwidth]{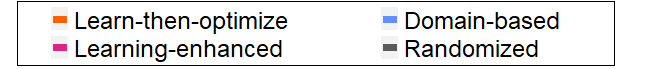}\hspace{2cm}
  \subfloat[Iteration view]{\includegraphics[width=0.4\textwidth]{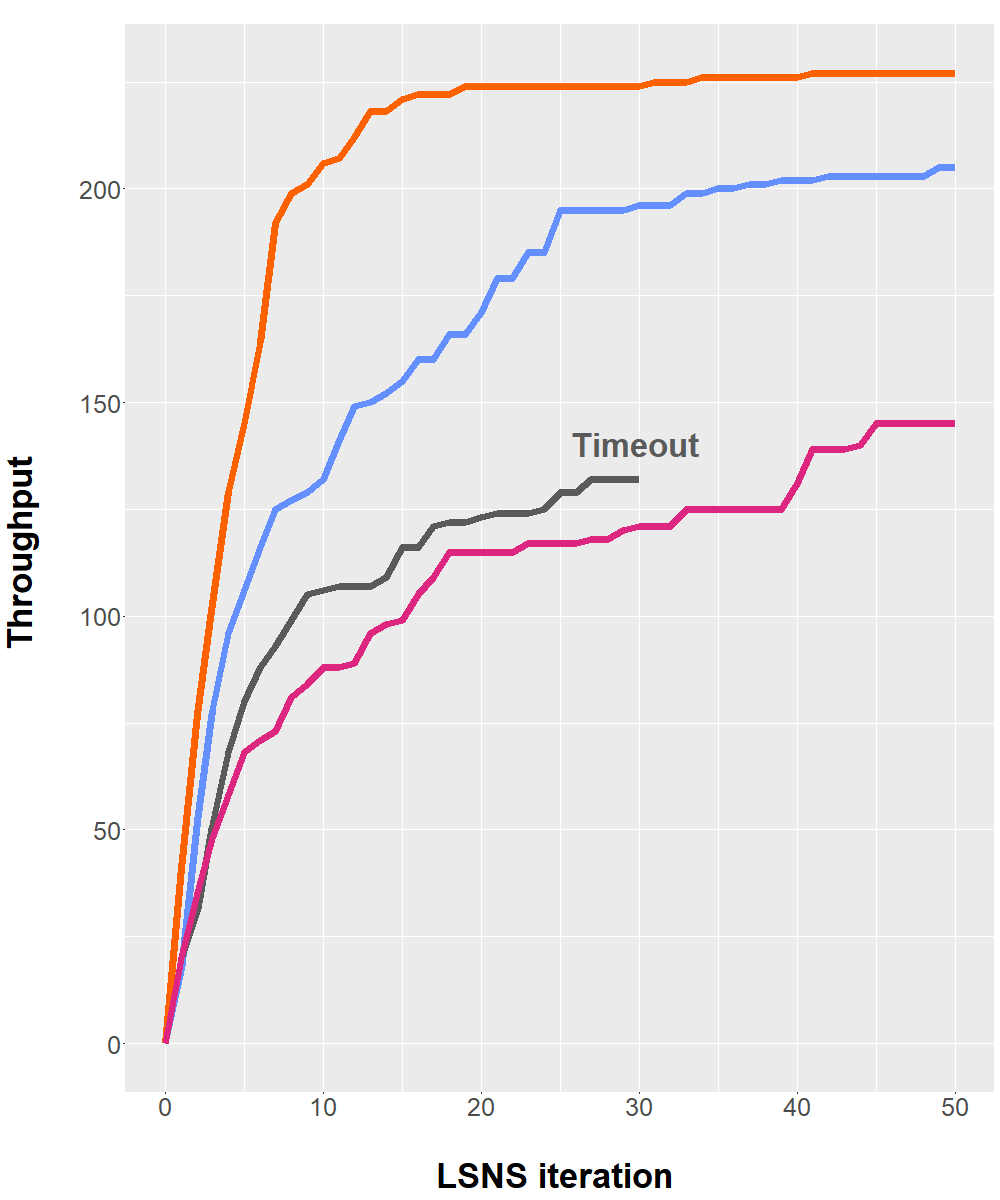}\label{fig:lsnsiter_sub}}
  \hfill
  \subfloat[Time-based view]{\includegraphics[width=0.4\textwidth]{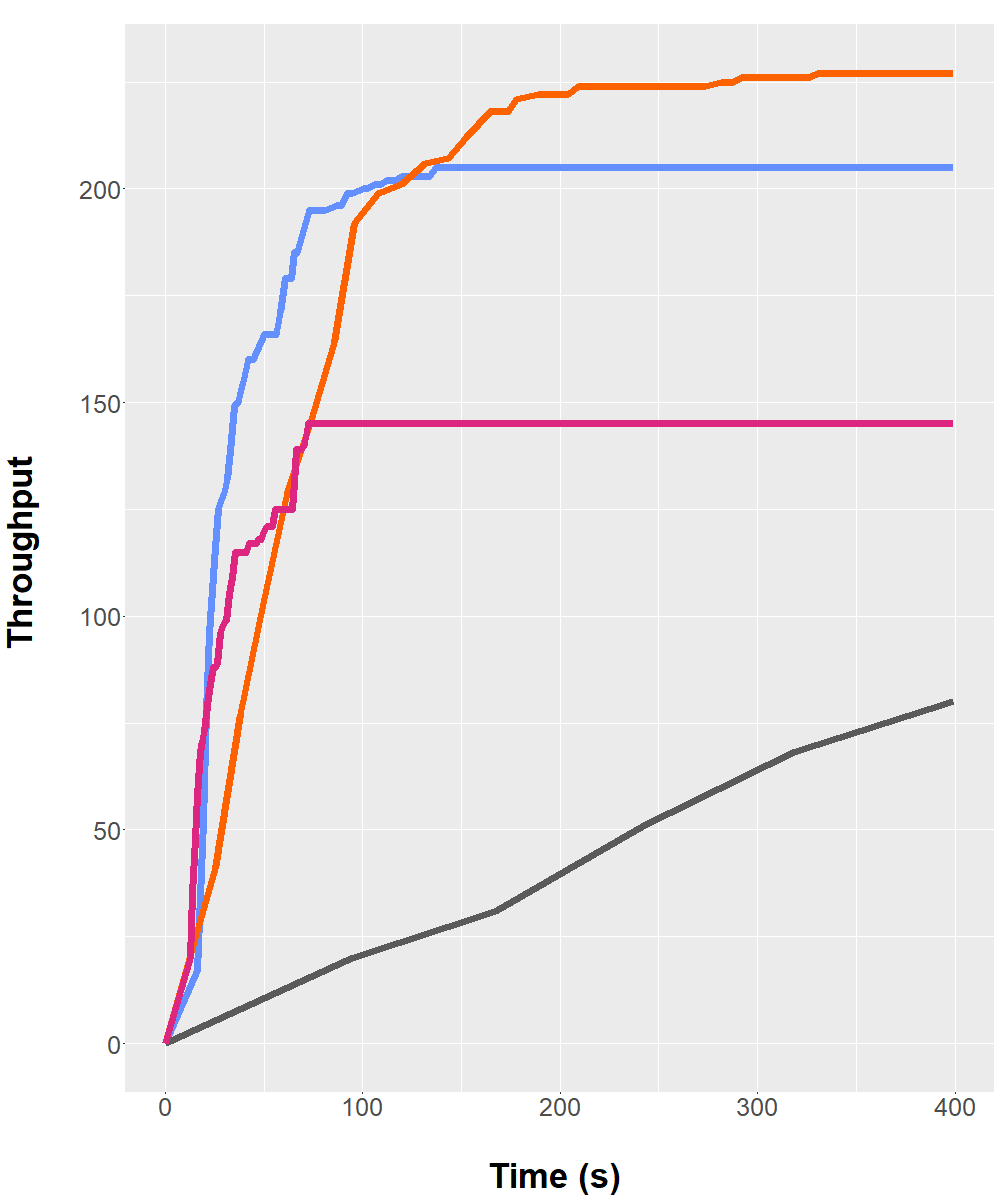}\label{fig:lsnstime_sub}}
  \caption{Iteration-by-iteration comparison across LSNS methods (small instance, moderate congestion).}
  \label{fig:lsnsiter}
  \vspace{6pt}
\end{figure}

In contrast, our learn-then-optimize approach consistently returns the best solution among all methods, with significant solution improvements. Figure~\ref{fig:lsnsiter} summarizes these results by plotting solution quality at each LSNS iteration and over time. This visualization underscores the progress of LSNS algorithms, the limitations of the learning-enhanced benchmark, and the strong performance of our learn-then-optimize approach. While the benchmarks exhibit slow progress over 50 LSNS iterations, the learn-then optimize algorithm improves rapidly in the first 10 iterations. Obviously, the learn-then-optimize algorithm takes more time per iteration, due to the extra time spent constructing a high-synergy subproblem at each iteration. Still, learn-then optimize outperforms all benchmarks after just two minutes of computations, ultimately resulting in an 11\% throughput increase as compared to the second-best solution for this instance. Meanwhile, the learning-enhanced benchmark is severely limited by the subproblem budget $B=100$ in two ways: the relatively small set of subproblems to select from limits the throughput improvement (as shown in the iteration-based view in Figure~\ref{fig:lsnsiter_sub}) and the necessity to evaluate all subproblems at each iteration increases computational speed significantly (as shown in the time-based view in Figure~\ref{fig:lsnstime_sub}). These results emphasize the strength of the learn-then-optimize methodology to effectively guide the LSNS toward high-synergy neighborhoods at moderate computational costs.

The learn-then-optimize algorithm also yields regression coefficients from the training phase, which provide interpretable insights regarding the drivers of solution improvements across neighborhoods. Table~\ref{tab:mlcoeffs} reports the coefficients of the holistic linear regression at the core of our learn-then-optimize approach. The first ten features characterize the input data. The last three characterize the congestion in the incumbent solution, which need to be re-computed at each iteration. The model does not include an intercept term for scaling and interpretability---namely, this choice ensures that the model predicts an objective of zero for an empty subproblem.

\begin{comment}
\begin{table} %%
\caption{Coefficients of the machine learning model (medium instance, moderate congestion).}
\centering
\footnotesize
\renewcommand{\arraystretch}{1}
\begin{tabular}{llcccc} \toprule
 & &  & \multicolumn{3}{c}{Example features for three subproblems} \\ 
  \cmidrule(lr){4-6}
Feature Group & Feature & Coefficients & High Synergy & Avg Synergy & Low Synergy \\ 
\toprule
Order & order size & -0.09 & 83 & 68 & 44 \\ 
Pod & percent requested items & -0.19 & 23.58 & 23.9 & 23.92 \\ 
Workstation & count & 0.74 & 2 & 1 & 2 \\ 
 & centrality  & 0.14 & 62.67 & 30 & 70.67 \\ 
Time & count  & 0.63 & 5 & 6 & 5 \\ 
Order-Pod & item overlap  & 1.14 & 49 & 34 & 21 \\ 
 & avg inventory  & -0.07 & 259 & 150.5 & 130.5 \\ 
Pod-Workstation & x-distance & -0.01 & 592 & 304 & 848 \\ 
 & y-distance & 0 & 828 & 420 & 984 \\
Workstation-Time & queue congestion  & -0.21 & 3 & 1.97 & 5.9 \\ 
 & local congestion  & 2.12 & 3.49 & 2.52 & 3.75 \\ 
 & semi-local congestion & -0.88 & 5.86 & 3.97 & 7.75 \\ 
\midrule
 &   \multicolumn{2}{r}{Predicted objective:} & 33.66 & 24.29 & 11.89 \\ 
 &   \multicolumn{2}{r}{Actual objective:} & 34 & 23 & 12 \\ 
\bottomrule
\end{tabular}
\label{tab:mlcoeffs_old}
\end{table}
\end{comment}

\begin{table} %%
\caption{Coefficients of the machine learning model (small instance, moderate congestion).}
\centering
\footnotesize
\renewcommand{\arraystretch}{1}
\begin{tabular}{llcccc} \toprule
 & &  & \multicolumn{3}{c}{Example features for three subproblems} \\ 
  \cmidrule(lr){4-6}
Feature Group & Feature & Coefficients & High Synergy & Avg Synergy & Low Synergy \\ 
\midrule
Order  & order size  & -0.414 & 56 & 40 & 54 \\
Pod  & percent requested items  & 0.117 & 23.4 & 23.2 & 23.2 \\
 & avg dist to stations & 0.002 & 634 & 640 & 618 \\
Workstation  & count  & -4.959 & 2 & 1 & 2 \\
  & centrality   & 0.355 & 44 & 20 & 24 \\
Time  & count   & 0.353 & 6 & 10 & 6 \\
Order-Pod  & item overlap   & 1.102 & 96 & 59 & 64 \\
  & avg inventory   & -0.051 & 992.33 & 578 & 710 \\
Pod-Workstation  & x-distance  & -0.012 & 608 & 208 & 304 \\
  & y-distance  & 0.004 & 660 & 336 & 354 \\
Workstation-Time  & queue congestion   & 2.219 & 1.72 & 1.17 & 0 \\
  & local congestion   & 0.126 & 10.17 & 9.42 & 0 \\
  & semi-local congestion  & -0.187 & 6.19 & 5.77 & 0 \\
\midrule
 & \multicolumn{2}{r}{Predicted objective:} & 43.18 & 30.34 & 14.44 \\
 & \multicolumn{2}{r}{Actual objective:} & 46 & 29 & 13 \\
 \bottomrule
\end{tabular}
\label{tab:mlcoeffs}
\end{table}

The feature with the most predictive power is item overlap across order-pod pairs, meaning that a subproblem is more likely to improve the solution if the pods contain useful items to fulfill the orders. Then, the distance between a pod and a workstation has a negative coefficient (negative coefficient for x-distance, smaller positive coefficient for y-distance), because assigning a pod to a workstation located far away has a negative impact on congestion. Order size also has a negative coefficient, reflecting that larger orders are harder to handle. %In contrast, workstations and time have positive intercepts because larger subproblems lead to higher throughput. 
Finally, local congestion has a negative coefficient, reflecting that existing congestion makes it more difficult to alter the incumbent solution toward a higher throughput. Albeit, local and queue congestion have positive coefficients due to multi-collinearity. In other words, congestion makes it more difficult to alter the incumbent solution toward a higher throughput. We explore these dynamics in more detail in Section~\ref{subsec:practical}.

\subsubsection*{Solving full-sized instances.}

\begin{figure} %%
    \centering
  \includegraphics[width=0.5\textwidth]{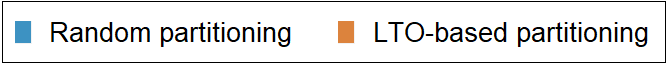}\hspace{2cm}
  \subfloat[Throughput]{\includegraphics[width=0.45\textwidth]{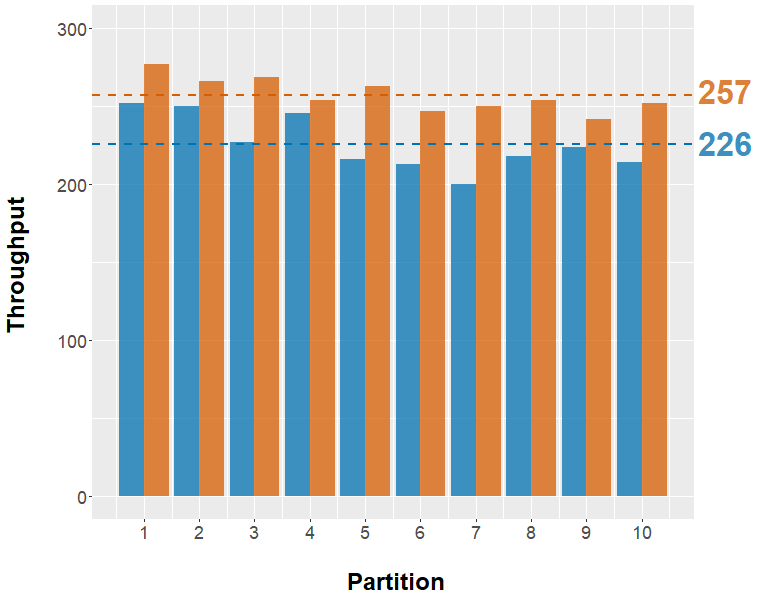}}\label{fig:partitionobj}
  \hfill
  \subfloat[Items picked per pod]{\includegraphics[width=0.45\textwidth]{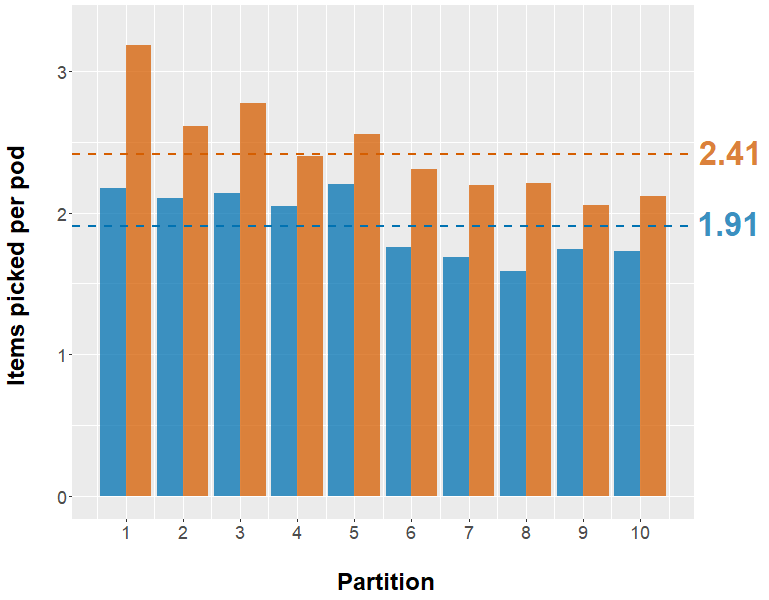}}\label{fig:partitionpileon}
  \caption{Comparison of order-region decomposition methods for a full-sized instance with 10 regions, where each region was solved via learn-then-optimize LSNS in under 10 minutes.}
  \label{fig:partitions}
  \vspace{-12pt}
\end{figure}

Recall that the learn-then-optimize LSNS approach does not scale beyond the medium-scale instances shown in Table~\ref{tab:bigtable}. We therefore apply the partitioning scheme presented in Section~\ref{subsec:decomposition} to decompose large-scale instances into spatial regions, using the learn-then-optimize outputs. We use $R=10$ regions, so each one corresponds to a small instance of the TDS-CW---recall that the full instance is roughly 10 times larger than the small instance (Table~\ref{tab:size}). Thanks to this decomposition, we obtain a solution in less than 10 minutes for the full-sized instance. The partitioning scheme is tested against a randomized order-region assignment benchmark, which defines a random feasible solution to the problem given in Equations~\eqref{part_assign}--\eqref{part_var}. Figure~\ref{fig:partitions} visualizes the throughput and number of items picked per pod under each solution.

Note that the learn-then-optimize approach results in a stronger partition, leading to a 14\% improvement in throughput and a 26\% improvement in item-pod consolidation as compared to the randomized benchmark. In other words, the learn-then-optimize results provide strong proxies for the synergies between entities, which provide additional benefits here to decompose the problem into separate subproblems. Moreover, the two performance metrics are consistently strong across regions, consistent with the results reported in Table~\ref{tab:bigtable}. These results suggest the robustness of the LTO-based assignments across spatial regions of the warehouse. Overall, our optimization methodology returns high-quality solutions to the full-sized instances---corresponding to the largest warehouses in practice---in reasonable computational times consistent with practical requirements.

\subsection{Practical impact}
\label{subsec:practical}

\begin{figure} %%
    \centering
    \includegraphics[width=\textwidth]{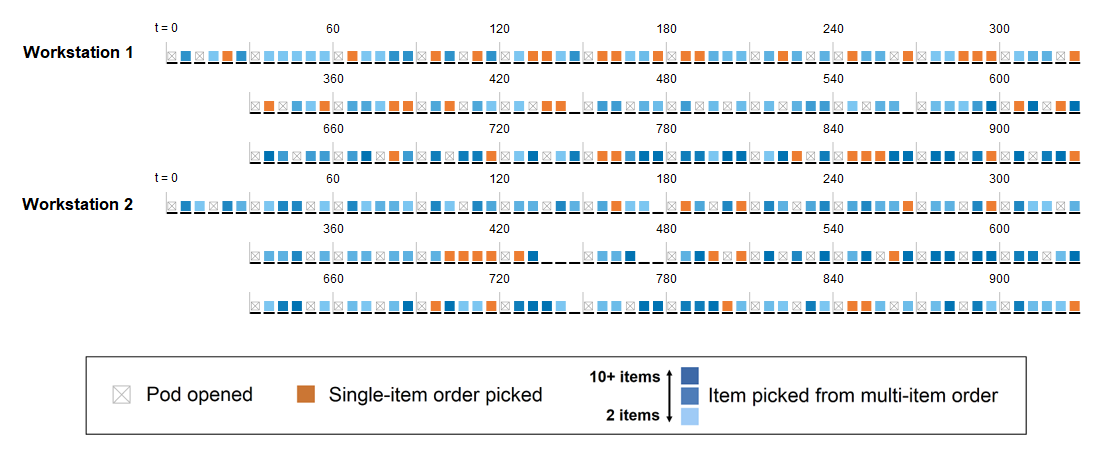}
    \caption{Sample task schedule at two workstations over the 15-minute horizon.}
    \label{fig:fullsol}
    \vspace{-12pt}
\end{figure}

We conclude with interpretable visualizations of the optimized solution to identify the drivers of robotic warehousing operations. Figure~\ref{fig:fullsol} displays a schedule at two workstations. At each time period, the picker can open new pods (gray boxes) and pick items (orange and blue boxes). The solution enables high utilization, with just 45 seconds of idle time over 15 minutes. One of the key drivers of high utilization is that multiple items are often picked from the same pod to minimize changeover times. In particular, single-item orders (shown in orange) are highly flexible, and can thus be served most often ``for free'' from a pod that is also serving another order. This situation occurs 52 times out of 56, in this example. In such cases, serving single-item orders does not impose pod changeover times and does not contribute to workstation capacity or to warehouse congestion.

\begin{figure}
    \centering
    \subfloat[Item and order picks per pod]{\includegraphics[width=0.4\textwidth]{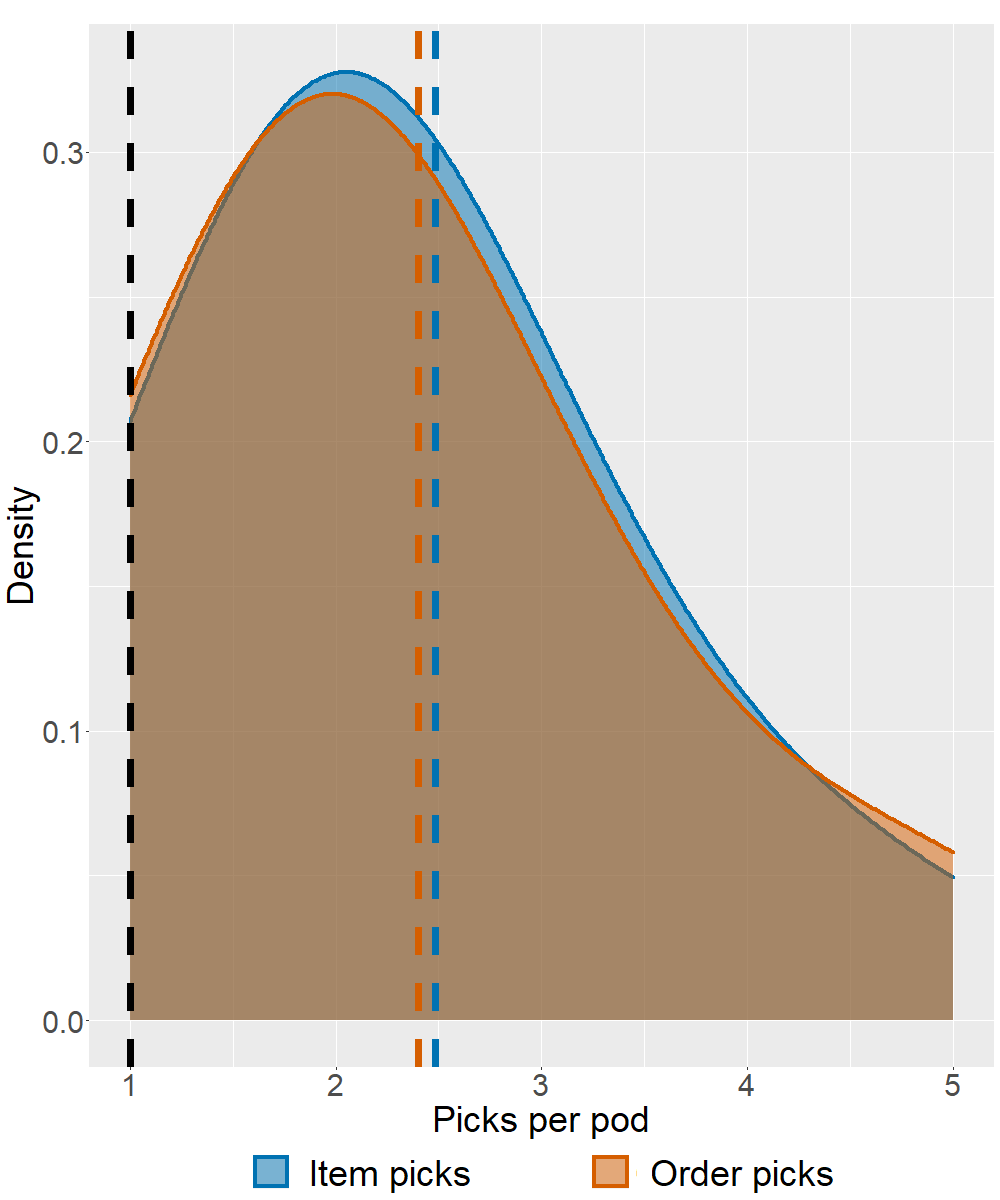}\label{fig:podpicks}}
     \:
    \subfloat[Distance traveled per pod]{\includegraphics[width=0.4\textwidth]{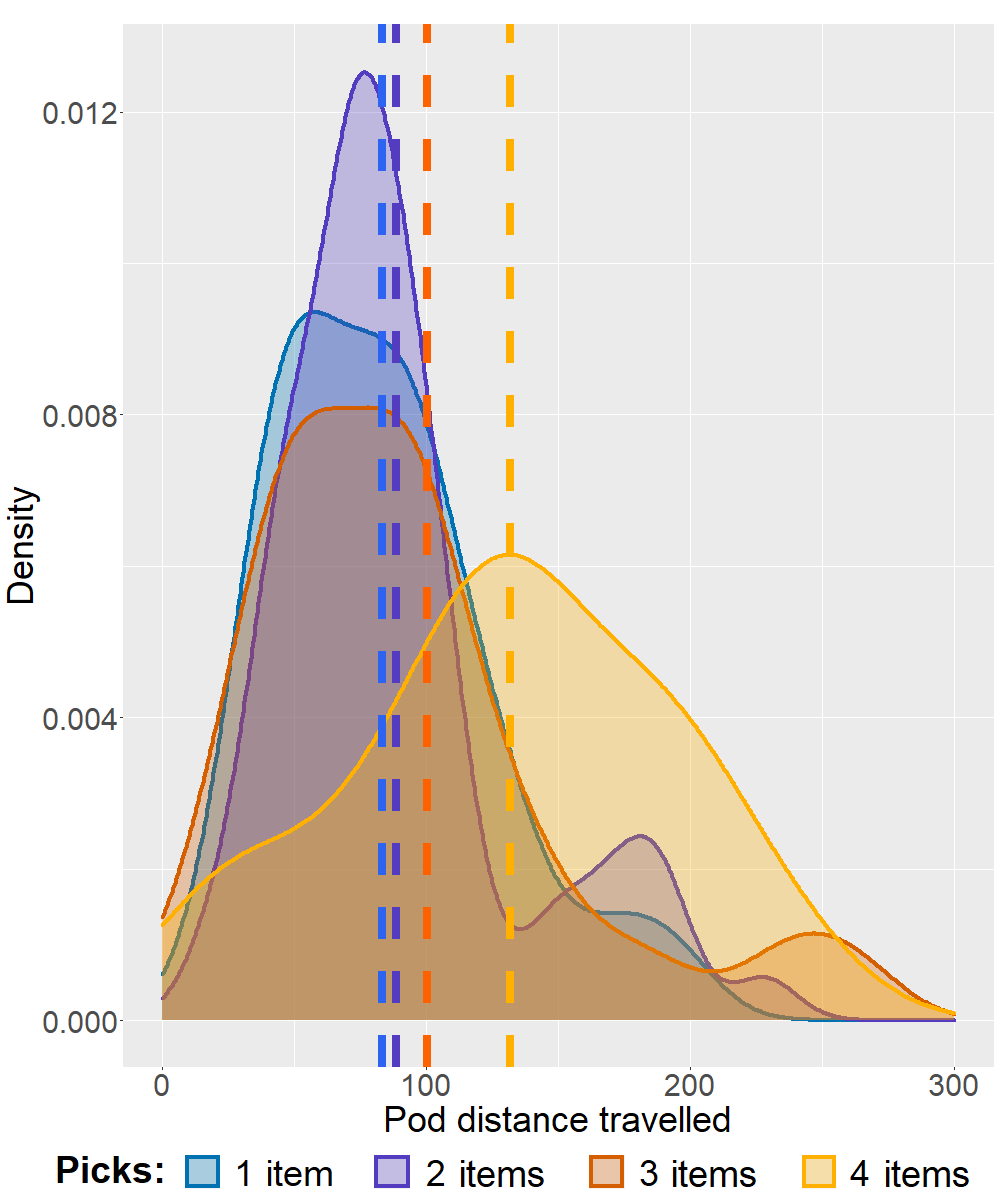}\label{fig:poddistance}}
    \caption{Density plots showing the dynamics of order coordination in the optimized solution.}
    \label{fig:distributions}
    \vspace{-12pt}
\end{figure}

At the aggregate level, Figure~\ref{fig:podpicks} shows that high utilization is indeed driven by pod consolidation, with 2.48 items picked per pod on average. One type of consolidation involves within-order consolidation, by bringing a pod that contain multiple items for one order---a ``low-hanging fruit'' since these pods are easily identifiable from input data. Another type of consolidation involves cross-order coordination, by bringing one pod to a workstation to pick items for different orders---such coordination is much more challenging to achieve because it requires those orders to be processed at the same workstation and at the same time. As the figure shows, each pod contributes 2.48 items on average toward fulfilling 2.40 orders, so our solution relies heavily on cross-order consolidation (with only within-order consolidation, each pod would contribute to one order; with only cross-order consolidation, each pod would contribute to 2.48 orders; our solution is much closer to the latter outcome than the former). In other words, the high throughput achieved in our solution does not merely stem from order and storage patterns in the input data, but primarily by coordinating order processing and pod movements to achieve high pick efficiency via cross-order consolidation.

\begin{figure} %%
    \centering
    \includegraphics[width=14cm]{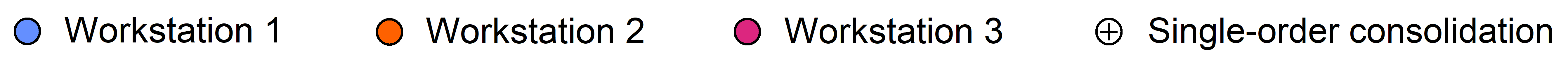}
    \subfloat[Greedy heuristic solution ]{\label{subfig:ordergraph_greedy}
            \includegraphics[width=5.5cm]{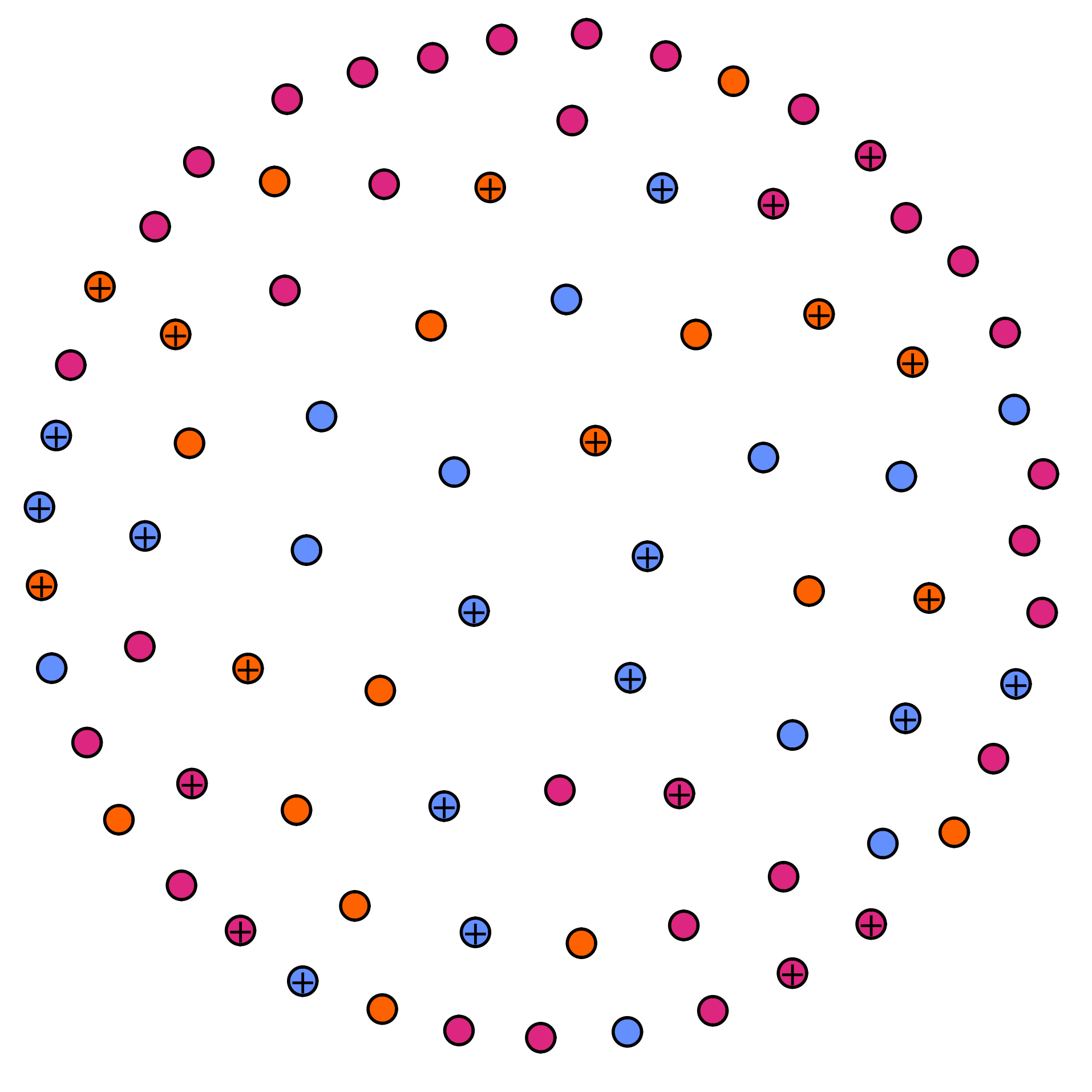}
        }\hspace{0.1 cm}
    \subfloat[Randomized LSNS solution]{\label{subfig:ordergraph_random}
            \includegraphics[width=5.5cm]{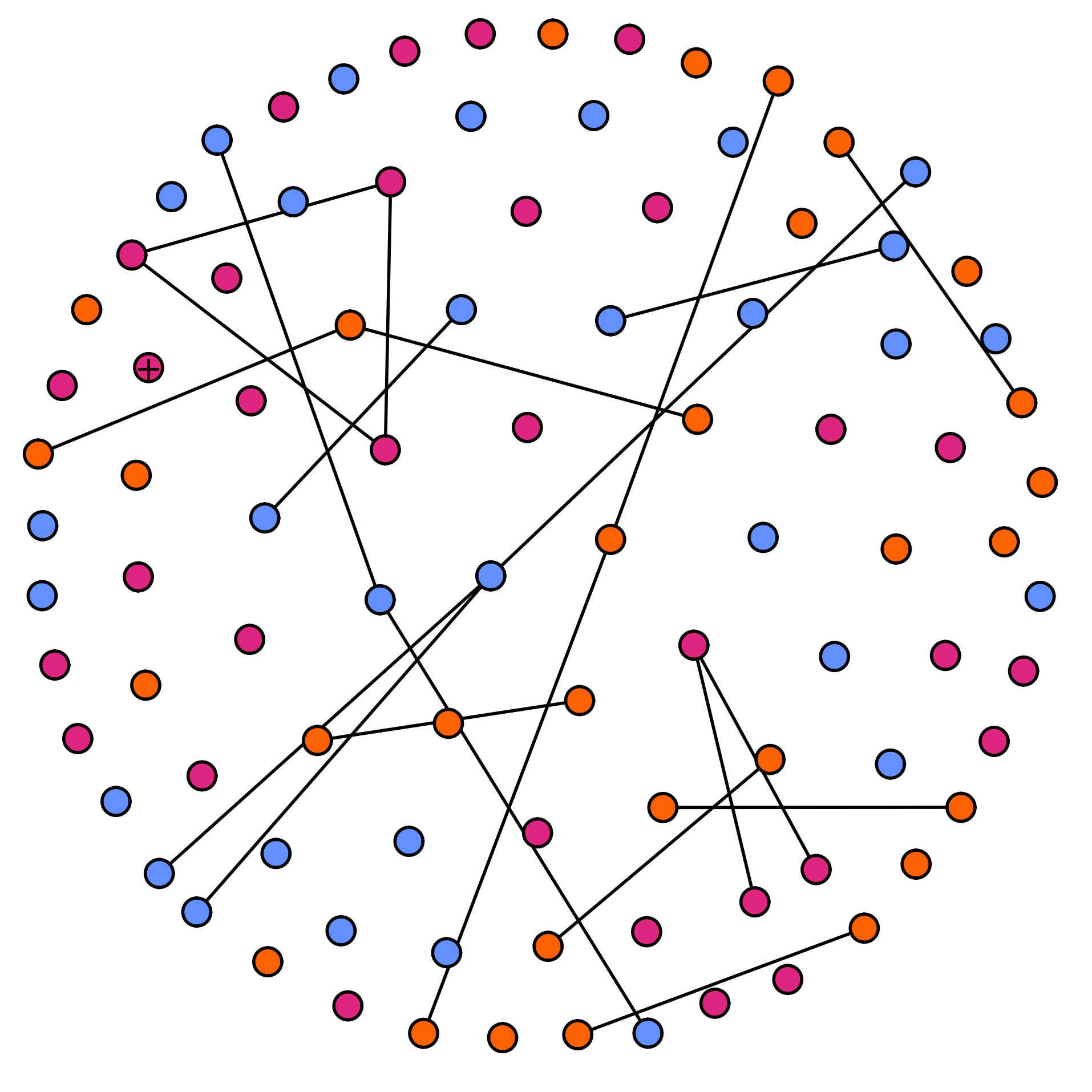}
        }\vspace{0.1 cm}
    \subfloat[Domain-based LSNS solution]{\label{subfig:ordergraph_synergy}
            \includegraphics[width=5.5cm]{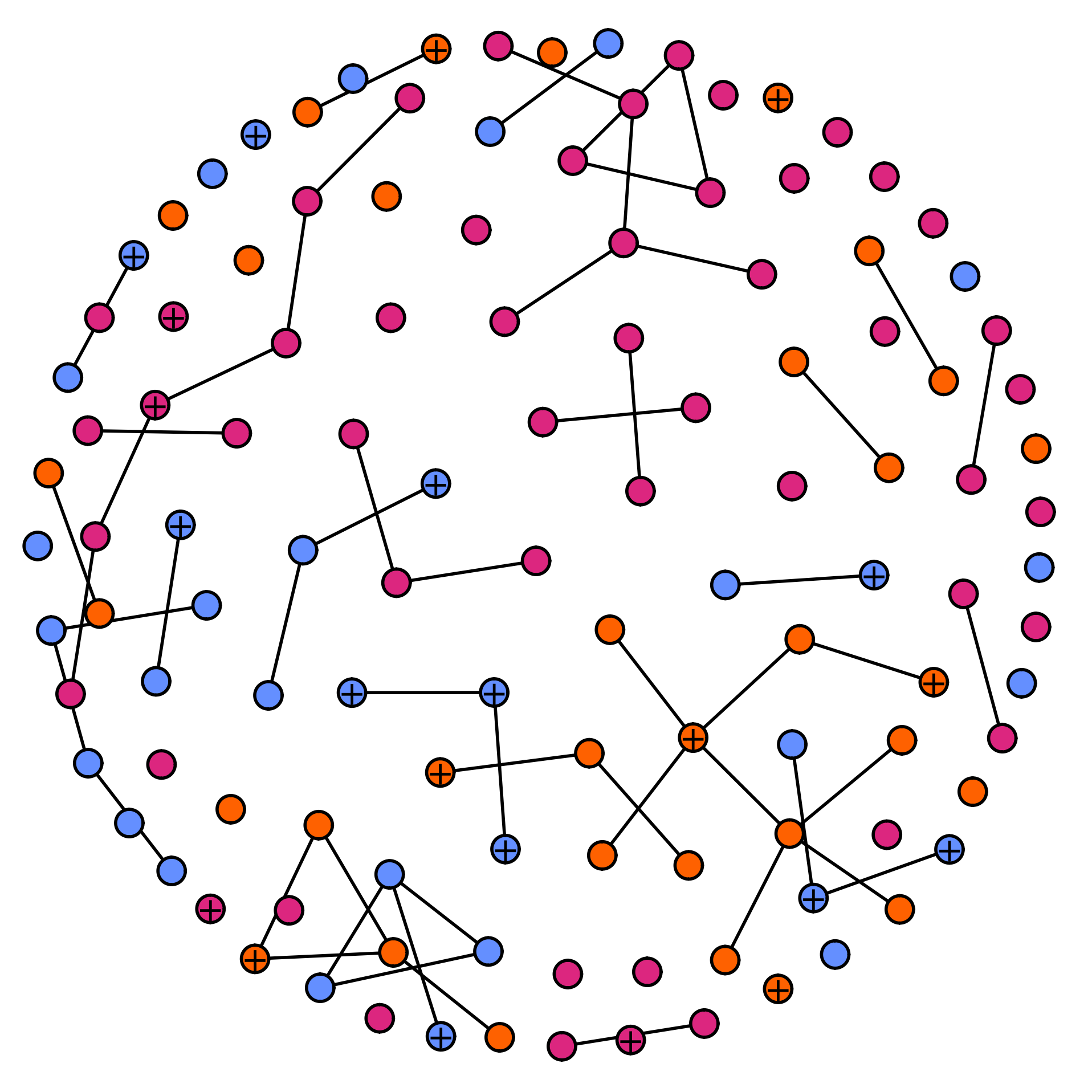}
        }\hspace{0.1 cm}
    \subfloat[Learn-then-optimize LSNS solution]{\label{subfig:ordergraph_lto}
            \includegraphics[width=5.5cm]{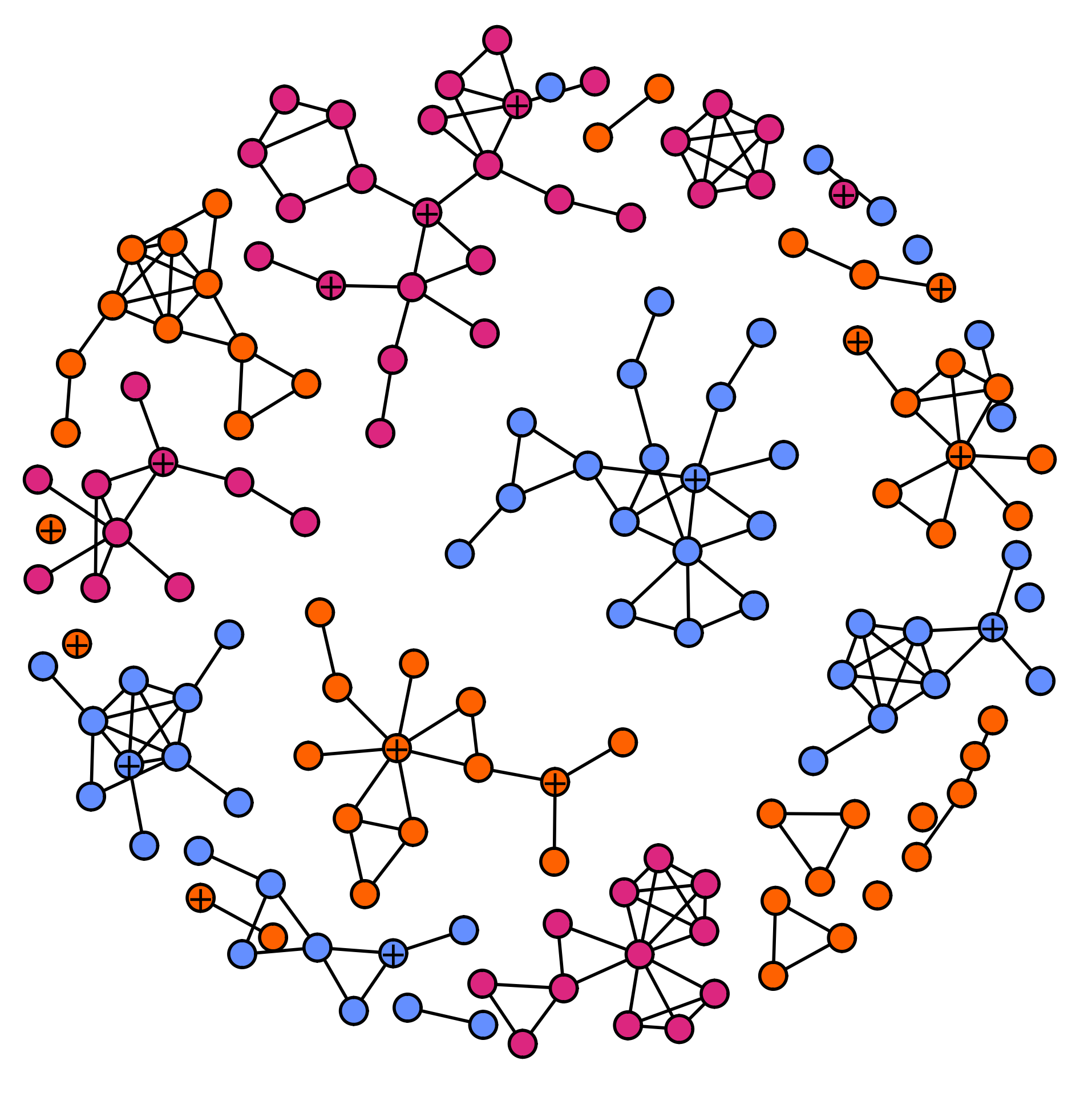}
        }
    \caption{Order synergy graphs. Each node represents an order, with the color showing its workstation assignment. Two orders are connected if at least one item was picked from a single shared pod. Orders labeled with a ``+'' are instances of within-order consolidation: picking two requested items from a single pod. 
    }
    \label{fig:ordergraphs}
\end{figure}

To stress this point, Figure~\ref{fig:ordergraphs} compare the within-order cross-order consolidation patterns in our solution compared to the benchmarks. The graphs comprise a node for each order and an edge for each pair of orders that are fulfilled, at least in part, by the same pod at the same time. In this representation, the extent of cross-order consolidation can be visualized by the connectedness of the graph. By assigning orders one-by-one, the greedy heuristic produces \emph{zero} cross-order picks. The domain-based heuristic is specifically designed to maximize overlapping inventory between selected orders and selected pods, and therefore results in more cross-order consolidation than the randomized LSNS solution (55 vs. 21 edges overall). However, it fails to execute as well as our learn-then-optimize solution. In comparison, the learn-then-optimize solution leads to the most connected graph by far (192 edges). Note, also, a decrease in within-order consolidation (17 instances for LTO vs. 25 for domain-based LSNS vs. 32 for greedy heuristic). Furthermore, the randomized LSNS solution has just one instance of within-order consolidation, indicating that these opportunities are unlikely to arise ``naturally'' without targeting synergistic order and pods. This suggests that the greedy heuristic and domain-based solutions are focused on maximizing within-order consolidation, at the expense of the harder-to-find but ultimately more fruitful cross-order consolidation. In turn, these results underscore the benefits of the predictive model to learn cross-order synergies and of the optimization-based subproblem selection model to capitalize on these synergies.

Finally, Figure~\ref{fig:poddistance} shows that pods often travel farther when the workstation will picks more items from it, while pods selected for a single-item pick are more often brought from nearby. This result highlights the cost of coordination in terms of distance traveled and congestion contributions: whereas it is generally preferable to bring pods to nearby workstations, it may be worth assigning a pod to a further workstation if it creates cross-order coordination opportunities. By optimizing over these trade-offs, our solution enables high throughput in high-density fulfillment operations.

\subsection{Benefits of congestion-aware task design and scheduling}

One of the main contributions of our model is to consider congestion in high-density fulfillment operations. To retain tractability, however, the TDS-CW formulation approximated congestion patterns via a proxy based on the shortest-path routes for each possible pod-workstation assignment. To evaluate the performance of this approach against a more realistic model of fulfillment center operations, we lock all workstation-order, order-item-pod, and order sequencing decisions and then run a downstream routing model that optimizes the movement of each pod to minimize task delays subject to congestion constraints at the intersections (see Appendix~\ref{app:routing} for details). In other words, our methodology optimizes task assignment and scheduling decisions using the congestion proxy, and then evaluates these decisions using a more realistic model of congestion.

Table~\ref{tab:congaware} compares the performance of our optimized solution against a simpler congestion-blind benchmark. By design, our congestion-aware solution keeps traffic within the allowed capacity at each intersection by spreading assignments across the warehouse, resulting in slightly higher traffic in most intersections but less in the highest-traffic areas. In comparison, the congestion-blind solution exceeds allowable capacity in several intersections, particularly in the corridors leading to the workstations. Obviously, these capacity limits are not hard constraints in practice, so Table~\ref{tab:congaware} estimates the downstream impact of both assignments on congestion and throughput.

\begin{table} %%
\centering
\small
\caption{Comparison of the congestion-blind and TDS-CW solutions.}
\renewcommand{\arraystretch}{1}
\begin{tabular}{lllccccc}\toprule\toprule
 &  & & \multicolumn{2}{c}{Planned} & \multicolumn{3}{c}{Realized} \\
 \cmidrule(lr){4-5}\cmidrule(lr){6-8}
Instance & Congestion & Model & Throughput & Util & Throughput & Util & Delay (sec.) \\
\midrule\midrule
Small & Tight & Congestion-blind & 189 & 89\% & 137 & 64\% & 95 \\ 
 && \textbf{TDS-CW} & 183 & 73\% & \textbf{182} & \textbf{72\%} & \textbf{42} \\
 \midrule
Small & Moderate & Congestion-blind & 239 & 95\% & 194 & 88\% & 68 \\ 
 && \textbf{TDS-CW} & 222 & 91\% & \textbf{218} & \textbf{89\%} & \textbf{57} \\
 \midrule
Large & Tight & Congestion-blind & 287 & 94\% & 230 & 77\% & 102 \\
 && \textbf{TDS-CW} & 258 & 77\% & \textbf{256} & \textbf{76\%} & \textbf{37} \\
 \midrule
Large & Moderate & Congestion-blind & 294 & 97\% & 243 & 81\% & 43 \\
 && \textbf{TDS-CW} & 278 & 90\% & \textbf{278} & \textbf{90\%} & \textbf{15} \\
\bottomrule\bottomrule
\end{tabular}
\label{tab:congaware}
\end{table}

These results indicate that, when evaluated with congestion constraints, the congestion-blind solution incurs a 17-28\% throughput loss against our congestion-aware solution, along with higher delays (43-102 seconds per pod on average) and more idle time at the workstations (2.5-8.0 extra minutes on average). Although approximate, our TDS-CW model buffers against these delays in the task design and assignment phase, leading to merely 15-57 seconds of delay and only 15-45 seconds of extra idle time. Ultimately, our congestion proxy, despite inducing significant computational complexities, enables stronger task design and scheduling decisions with significant practical benefits: higher throughput, higher utilization and lower delays.

\section{Conclusion}

This paper formulated a \textit{task design and scheduling model with congestion and workload restrictions (TDS-CW)} to optimize fleet management operations in an automated fulfillment center. The integer optimization model maximizes throughput by assigning each customer order to a workstation and by bringing pods to the workstations for a human picker to retrieve the corresponding items, while minimizing human workload and warehouse congestion. We proposed a large-scale neighborhood search to iteratively improve an incumbent solution, and a learn-then-optimize approach to generate a subproblem at each iteration. Specifically, the learn-then-optimize approach first learns an offline machine learning model to predict the objective function of a function of features of low-dimensional elements of a subproblem, and then formulates a binary optimization problem to find the neighborhood with maximal predicted improvement. This approach augments existing learning-enhanced LSNS algorithms by optimizing over the full-dimensional space of all possible subproblems, as opposed to selecting one from a pre-defined set of candidates. 

Based on an ongoing collaboration with Amazon, extensive computational experiments demonstrate that the methodology scales to otherwise-intractable instances, and improves throughput by 4-14\% against state-of-the-art benchmarks. When augmented with a partitioning-based decomposition scheme, our learn-then-optimize approach can return high-quality solutions to real-world instances of the TDS-CW in minutes. The main drivers of high utilization involve picking multiple items from fewer pods and avoiding congestion in the warehouse---and carefully managing these conflicting objectives when pods containing multiple items are far away from the workstations. At a time when robotic technologies are becoming ubiquitous, this paper contributes new models and algorithms to manage and coordinate large fleets of autonomous agents in fulfillment center operations while regulating human workload and overcrowding in physically-constrained environments.

\ACKNOWLEDGMENT{The help of Riley Lenaway with data analyses and exploratory experiments is gratefully acknowledged. The authors are thankful for the guidance and support from Michael Caldara and Armon Shariati from Amazon.}

\bibliographystyle{informs2014}
\bibliography{refs}

\newpage

\begin{APPENDICES}

\section{Details of learn-then-optimize for the TDS-CW}\label{app:leanoptimizeTDSCW}

\subsection{Greedy heuristic} \label{app:greedy}

LSNS must be initialized with a feasible solution. In the case of the TDS-CW, an ``empty'' solution with no orders or pods assigned is trivially feasible and can serve as the LSNS initial solution. We also propose a greedy heuristic to improve upon the empty solution and accelerate the LSNS algorithm. Note that we use the greedy heuristic both as a benchmark and a warm start in our LSNS algorithms.

The key difficulty in solving TDS-CW heuristically is that the numerous complicating capacity constraints make it hard to find reasonable feasible solutions without some an optimization model. However, the operational flexibility of the fulfillment center makes the solution space prohibitively large to search over. We thus propose a heuristic that solves many single-order optimization problems in succession, while restricting the solution space for each to retain tractability.

The heuristic treats orders one-by-one, beginning from the largest orders, and solving two optimization problems to find a feasible assignment for each. The first model finds a workstation-time-block that has physical space and sufficient idle time to complete the order. The second model then solves the full problem for that order-workstation-time-block and makes a final assignment that satisfies all problem constraints. Once an order is assigned, it is never moved or re-assessed. The detailed algorithm is described in Algorithm~\ref{alg:greedy}. 
\begin{algorithm}[h!]
\vspace{0.5em}
\caption{Greedy heuristic for an initial solution to TDS-CW}
\label{alg:greedy}\small
    \begin{algorithmic} 
    \State $\overline{\bx}, \overline{\by} \gets$ empty solution 
    \State $\calM^\prime \gets$ sort $\calM$ descending by $|\calI_m|$ 
    \For {$m \in \calM^\prime$}
        \State $\calW^\prime \gets$ sort $\calW$ ascending by best-case pod distance to workstation to fulfill order $m$
        \For {$w \in \calW^\prime$}
            \State $\calT^\prime \gets$ Solve (insert eq refs here) to find a workstation-time block
            \State $\bx^\prime, \by^\prime \gets $  Solve (insert eq refs here) over $\calT^\prime$ to find a feasible assignment
            \If {found feasible $\bx^\prime, \by^\prime$}
                \State Update $\overline{\bx}, \overline{\by} $ with new assignments $\bx^\prime, \by^\prime $ for order $m$
                \State Move on to next order
            \Else
                \State Move on to next workstation
            \EndIf
        \EndFor
      \State If no assignment found, do not assign $m$ and move on to next order
    \EndFor
    \State \textbf{return } $\overline{\bx}, \overline{\by}$
    \end{algorithmic}
\end{algorithm}

\subsection{Holistic linear regression model}\label{app:holistic}

Recall that we define seven categories of feature :
\begin{enumerate}[label=(\roman*)]
    \item Feature of each order, with $\calS_f^{M} = \calM$: size of each order.
    \item Feature of each pod, with $\calS_f^{P} = \calP$: percentage of requested items on each pod.
    \item Feature of each workstation, with $\calS_f^{W} = \calW$: intercept, centrality of workstations
    \item Feature of each time period, with $\calS_f^{T} = \calT$: intercept
    \item Features of order-pod pairs, with $\calS_f^{MP} = \calM\times\calP$: item overlap and average inventory.
    \item Features of pod-workstation pairs, with $\calS_f^{PW} = \calP\times\calW$: x-distance and y-distance.
    \item Features of workstation-time pairs, with $\calS_f^{WT} = \calW\times\calT$: congestion at the workstation (``queue congestion'') in its immediate vicinity (``local congestion'') and in its broader vicinity (``semi-local congestion'').
\end{enumerate}
In addition, we define two non-linear transformations on each feature, $log(\cdot)$ and $(\cdot)^2$. Let $\calF$ be the set of all features and transformed features and let $\calT_f$ be the set of features that are transformations of feature $f$. 

We use a holistic regression framework to find an optimal linear regression model that satisfies standard modeling constraints. To capture this, we introduce a binary variable $z_f$ that takes value one if feature $f$ is used by the model, i.e. the corresponding $\beta_f$ is non-zero. The learning model is given as follows:
\begin{align} 
 \min_{\boldsymbol{\beta}}  \quad & \sum_{u=1}^U \bigg\|y_{u}- \sum_{m \in \calM^u} \beta^{M}_f \alpha^{M}_{m,f} - \sum_{p \in \calP^u} \beta^{P}_f \alpha^{P}_{p,f} - \sum_{w \in \calW^u} \beta^{W}_f \alpha^{W}_{w,f} + \sum_{t \in \calT^u} \beta^{T}_f \alpha^{T}_{t,f}  \label{hol_obj}\\ 
 & \quad\quad\quad - \sum_{w \in \calW^u} \sum_{t \in \calT^u}  \beta^{WT}_f \alpha^{WT}_{w,t,f} - \sum_{m \in \calM^u} \sum_{p \in \calP^u}  \beta^{MP}_f \alpha^{MP}_{m,p,f} \nonumber -  \sum_{w \in \calW^u} \sum_{p \in \calP^u}  \beta^{WP}_f \alpha^{WP}_{w,p,f} \bigg\|  + \Gamma || \boldsymbol{\beta} ||\\ 
 \text{s.t   } \quad &\displaystyle -M z_f \leq \beta_f \leq M z_f \quad \forall f\in\calF \label{hol_bigm} \\
 &\displaystyle z_f + z_j \leq 1 \quad \forall (f,j) \in \calH\calC \label{hol_hc}\\
  &\displaystyle \sum_{j \in \calT_f} z_j \leq 1 \quad \forall f \in \calF \label{hol_trans}\\
 &\displaystyle  z_f \in \{0,1\} \quad \forall f\in\calF
\end{align}

The objective~\eqref{hol_obj} minimizes the absolute difference between predicted and actual objective improvements, in addition to a regularization term controlled by parameter $\Gamma$. Constraints \eqref{hol_bigm} ensure consistency between the $\beta$ and $z$ variables using the big-M method. Constraints \eqref{hol_hc} then ensure that pairs of features that are highly correlated are not selected together, where $\calH\calC$ is the set of the highly-correlated feature pairs. In our model, this includes pod-workstation total distance versus x- or y-distance, order size versus the single-item order flag, and total congestion versus max congestion. Finally, constraints \eqref{hol_trans} enforce that we can select at most one transformation of each raw feature. 

\begin{table} %%
\caption{Summary of training data and holistic regression.}
\label{tab:training}
\centering\footnotesize\renewcommand{\arraystretch}{1}
\begin{tabular}{llcccc} \toprule
\toprule
 &  & Small inst. & Small inst. & Medium inst. & Medium inst. \\
 &  & Tight cong. & Moderate cong. & Tight cong. & Moderate cong. \\
\midrule
Training data & Data points & 20,000 & 20,000 & 20,000 & 20,000 \\
 & Generation time & 9.31 hrs. & 7.86 hrs. & 20.58 hrs. & 29.21 hrs. \\
\midrule
Holistic regression & Solve time & 238 sec. & 166 sec. & 663 sec. & 395 sec. \\
 & In-sample $R^2$ & 0.63 & 0.80 & 0.82 & 0.87 \\
 & In-sample MSE & 26.20 & 24.80 & 20.84 & 20.23 \\
 & In-sample MAE & 4.07 & 3.93 & 3.63 & 3.58 \\
\bottomrule\bottomrule
\end{tabular}
\end{table}

\section{Benchmark models and algorithms}

\subsection{Domain-based heuristic for LSNS} \label{app:synergy}

Typical approaches to subproblem selection for LSNS involve selecting or constructing a subproblem based on domain knowledge regarding where an existing solution is sub-optimal. We develop a domain-based heuristic to select a subproblem (i.e. 1-2 workstations, a continuous time block, $N_P$ pods, and $N_O$ orders) at each iteration which proceeds at each iteration of LSNS as follows:
\begin{enumerate}[label=(\arabic*)]
    \item Enumerate all possible workstation-time blocks. For even LSNS iterations, use interval time length $t_1$. For odd iterations, use $t_2 < t_1$ (even iterations select one workstation, while odd iterations select two).  
    \item From the set of possible workstation-time blocks, select the workstation with the lowest throughput. Break ties first by selecting the workstation-time block with the fewest orders open, then randomly.
    \item On odd iterations, add a second workstation by selecting the block that has the lowest throughput during the selected time period from the previous step (breaking ties in the same way).
    \item Select pods that are already visiting the workstation(s) during the selected time block, and let $N_P^{\text{curr}}$ be the number of pods added in this step.
    \item Select $\max(N_{P} - N_P^{\text{curr}} - N_P^{\text{res}},0)$ additional pods that are close to the workstation(s). Note: $N_P^{\text{res}}$ pods are reserved for the final step.
    \item Select orders that are already visiting the workstation(s) during the selected time block, and let $N_O^{\text{curr}}$ be the number of orders added in this step.
    \item Select $\max(N_{O} - N_O^{\text{curr}} ,0)$ additional orders that contain the most overlapping items with the pods assigned to the subproblem so far.
    \item Select $\min(N_P^{\text{res}}, N_P - N_P^{\text{curr}})$ additional pods that contain inventory required to complete the selected orders, prioritizing pods close to the workstation(s) and that have the most required items.
\end{enumerate}

\subsection{Decomposition heuristics} \label{app:decomp}

To benchmark our learn-then-optimize large-scale neighborhood search algorithm, we compare to two decomposition methods that break the full problem into two or three sequential decision problems.  

\subsubsection{Two-stage decomposition}

The first stage makes order-workstation and order-item-pod assignments, while minimizing the total number of pods used.
There are two binary decision variables, $z_{mw}$ which takes values one if order $m$ is assigned to workstation $w$, and $x_{imp}$ which takes value one if item $i$ from order $m$ is assigned to be picked from pod $p$. Auxiliary variable $y_{wp}$, used in the objective function, takes values one if pod $p$ is assigned to visit workstation $w$. The constraints enforce order assignment to a single workstation \eqref{2decomp_mw}, order-item assignment to a single pod \eqref{2decomp_imp}, pod inventory \eqref{2decomp_inv}, and consistency between the $\bx, \by, \bz$ variables \eqref{2decomp_cons}. Constraints \eqref{2decomp_ordermax} enforce that the number of orders assigned to each workstation is roughly balanced, with the total number of items to be picked at each between lower and upper bounds, $\underline{I}$ and $\overline{I}$.

\begin{align}
    \displaystyle \min \quad & \sum_{w \in \calW} \sum_{p \in \calP} y_{wp} \\
    \text{s.t.} \quad & \sum_{w \in \calW} z_{mw} = 1 \quad \forall m \in \calM \label{2decomp_mw} \\
    & \sum_{p \in \calP_i} x_{imp} = 1 \quad \forall m \in \calM, i \in \calI_m \label{2decomp_imp} \\
    & \sum_{m \in \calM_i} x_{imp} \leq u_{ip} \quad \forall i \in \calI, p \in \calP_i \label{2decomp_inv}\\
    & y_{wp} \geq x_{imp} + z_{mw} - 1 \quad \forall w \in \calW, m \in \calM, i \in \calI_m, p \in \calP_i \label{2decomp_cons} \\
    & \underline{I} \leq \sum_{m \in \calM} |\calI_m| \cdot  z_{mw} \leq \overline{I} \quad \forall  w \in \calW \label{2decomp_ordermax}\\
    &\displaystyle \text{Domain of definition: $\bx,\ \by,\ \bz$ binary} \nonumber 
\end{align}

The second stage takes as input the order-workstation and order-item-pod assignments, then determines workstation schedules that satisfy pod flow conservation, workstation order capacity, and congestion constraints. Binary decision variables include $x_{imt}$, which takes value one if item $i$ from order $m$ is picked at time $t$; $y_{pa}$, which takes value one if pod $p$ is assigned to arc $a$; and $v_{mt}$ which is one if order $m$ is open at time $t$. Auxiliary variables $f_{mt}$ and $g_{mt}$, along with constraints \eqref{2decomp_v}-\eqref{2decomp_g}, ensure consistency between $\bx$ and $\bv$. Further constraints enforce maximum open orders \eqref{2decomp_cap}, workstation throughput \eqref{2decomp_tpt}, pod flow balance \eqref{2decomp_pinit}-\eqref{2decomp_pflow}, maximum intersection congestion \eqref{2decomp_traff}, and consistency between the $\bx$ and $\by$ variables \eqref{2decomp_link}-\eqref{2decomp_nondec}. The objective function \eqref{2decomp_obj} maximizes the total throughput by the end of the time horizon.

\begin{align} 
\max  & \displaystyle \sum_{m \in \calM} \sum_{i \in \calI_m} x_{imT} \label{2decomp_obj} \\
    	\text{s.t   } &\displaystyle \sum_{m \in \calM(w)} v_{mt} \leq C_w \quad \forall w \in \calW, t \in \calT \label{2decomp_cap} \\
    	&\displaystyle \sum_{m \in \calM(w)} \sum_{i \in \calI_m} \delta^{\text{item}} (x_{imt} - x_{im,t-1}) + \sum_{p \in \calP} \sum_{a \in \calA^+_{N(w,t)}} \delta^{\text{pod}} y_{pa} \leq \Delta \quad \forall w \in \calW, t \in \calT \label{2decomp_tpt} \\
    	&\displaystyle  \sum_{a \in \calA^+_{N_p^{0}}} y_{pa} = 1 \label{2decomp_pinit} \\
    	&\displaystyle  \sum_{a \in \calA^-_{n}} y_{pa} - \sum_{a \in \calA^+_{n}} y_{pa} = 0 \quad \forall p \in \calP, n \in \calN \setminus (\calN_{end} \cup \{ N_p^{0}\}) \label{2decomp_pflow}  \\
    	&\displaystyle \sum_{p \in \calP} \sum_{a \in \calA} q_{ajt} y_{pa} \leq Q_{x} \quad \forall x \in \mathcal{X}, t \in \calT \label{2decomp_traff} \\
    	&\displaystyle (x_{imt} -  x_{im,t-1}) \leq  \sum_{a \in \calA_{\text{tr}} \cap \calA^+_{N(w,t)}} y_{pa} \quad \forall w \in \calW, m \in \calM(w), i \in \calI_m, p \in \calP_i, t \in \calT \label{2decomp_link} \\
    	&\displaystyle x_{imt} \geq x_{im,t-1} \quad \forall m \in \calM, i \in \calI_m, t \in \calT \label{2decomp_nondec} \\
    	&\displaystyle  v_{mt} = f_{mt} - g_{mt} \quad \forall m \in \calM, t \in \calT  \label{2decomp_v} \\
    	&\displaystyle  f_{mt} \geq x_{imt} \quad \forall m \in \calM, i \in \calI_m, p \in \calP_i, t \in \calT   \\
    	&\displaystyle  g_{mt} \leq x_{imt} \quad \forall m \in \calM, i \in \calI_m, t \in \calT \label{2decomp_g} \\
    	&\displaystyle \text{Domain of definition: $\bx,\ \by,\ \bff,\ \bg,\ \bv$ binary}   \label{2decomp_bds}
\end{align} 

\subsubsection{Three-stage decomposition}

The three-stage decomposition further breaks down the order-workstation and order-item-pod assignments into two sequential problems. The first stage makes order-workstation assignments while maximizing pod overlap opportunities. Binary decision variable $z_{mw}$ takes value one if order $m$ is assigned to workstation $w$, while auxiliary variable $x_{mnw}$ takes value one if both order $m$ and order $n$ are assigned to workstation $w$. The objective \eqref{3decomp_obj} maximizes the pod overlap between orders assigned to the same workstation, using a pre-processed parameter $\sigma_{mn}$ which represents the total number of pods that contain an item from order $m$ and order $n$. Constraints enforce order assignment to one workstation \eqref{3decomp_mw}, minimum and maximum items assigned to be picked at each workstation \eqref{3decomp_mi}, and consistency between $\bx$ and $\bz$ variables \eqref{3decomp_cons}.
\begin{align}
    \displaystyle \max \quad & \sum_{w \in \calW} \sum_{m \in \calM}\sum_{n \in \calM \setminus \{m\} } \sigma_{mn} x_{mnw} \label{3decomp_obj}\\
    \text{s.t.} \quad & \sum_{w in \calW} z_{mw} = 1 \quad \forall m \in \calM \label{3decomp_mw}\\
    & \underline{I} \leq \sum_{m \in \calM} |\calI_m| \cdot  z_{mw} \leq \overline{I} \quad \forall  w \in \calW \label{3decomp_mi} \\
    & x_{mnw} \leq \frac{1}{2} \left( z_{mw} +  z_{nw} \right) \quad \forall m \in \calM, n \in \calM \setminus \{m\}, w \in \calW \label{3decomp_cons} \\
    &\displaystyle \text{Domain of definition: $\bx,\ \bz$ binary} \nonumber 
\end{align}
The second stage assigns pods to serve orders and items within each workstation. Binary variables $x_{imp}$ take value one if item $i$ from order $m$ is assigned to pod $p$, while auxiliary binary variables $y_{wp}$ take value one if pod $p$ is assigned to go to workstation $w$. The objective \eqref{3decomp_obj2} minimizes the number of pods required to serve all orders. 
This problem can be decoupled by workstation, each solved in parallel. Constraints enforce pod assignment to each item \eqref{3decomp_imp}, pod inventory \eqref{3decomp_inv}, and consistency between $\bx$ and $\by$ variables \eqref{3decomp_xy}.
\begin{align}
    \displaystyle \min \quad & \sum_{w \in \calW} \sum_{p \in \calP} y_{wp} \label{3decomp_obj2} \\
    \text{s.t.} \quad & \sum_{p \in \calP_i} x_{imp} = 1 \quad \forall m \in \calM, i \in \calI_m \label{3decomp_imp}\\
    & \sum_{m \in \calM_i} x_{imp} \leq u_{ip} \quad \forall i \in \calI, p \in \calP_i \label{3decomp_inv} \\
    & y_{wp} \geq x_{imp} \quad \forall w \in \calW, p \in \calP_i, m \in \calM(w) \label{3decomp_xy} \\
    &\displaystyle \text{Domain of definition: $\bx,\ \by$ binary} \nonumber 
\end{align}
The third stage is the same as the final stage in the two-decomposition, making workstation scheduling and pod flow assignments, described fully in Equations~\eqref{2decomp_obj}-\eqref{2decomp_bds}.

\subsection{Downstream operations: path-finding and scheduling} \label{app:routing}

To assess the benefits of the congestion-aware TDS-CW solution, we evaluate assignments made by our model compared to a congestion-blind model by passing the assignments to a downstream routing model, presented below.

The model takes as inputs derived directly from the solution of the TDS-CW: $\calM^\prime$ the set of assigned orders, $w(m)$ the workstation assignment of order $m$, and $o(m)$ the assigned start time. To complete order $m$ in the path-finding and scheduling model, all pods $p \in \calP_m$ (the set assigned to fulfill order $m$ in the TDS-CW) must be delivered to the workstation by a deadline $t_{pm}^{\text{plan}}$ (based on the planned delivery time from the TDS-CW). The model finds a path for each pod from its storage location to the workstation and back, while enforcing congestion constraints at each intersection. Binary variable $y_{pa}$ takes value one if pod $p$ is assigned to arc $a$. Unlike all previous formulations, the arc $a$ is now associated with a specific path of intersections through the warehouse, rather than a fractional congestion proxy. The objective \eqref{route_obj} minimizes the total pod delay, defined as the time between the scheduled pod pick time from the task assignment model $t_{pm}^{\text{plan}}$ and the actual pick time in the routing model. Constraints \eqref{route_pinit}-\eqref{route_pflow} enforce initial location and flow conservation for each pod. Constraints \eqref{route_assign} enforce that a pod $p$ that is assigned to serve order $m$ must visit and depart from the order's assigned workstation $w(m)$ sometime after the order's assigned start time $o(m)$. Finally, constraints \eqref{route_traff} enforce intersection capacity, with binary parameter $q^\prime_{a \ell t}$ representing whether arc $a$ passes through intersection $\ell$ at time $t$.

\begin{align}
    \displaystyle \min \quad &\sum_{m \in \calM^\prime} \sum_{p \in \calP_m} \sum_{a \in \calA_{mp}} (t_a - t^{\text{plan}}_{pm} \cdot y_{pa}) \label{route_obj} \\
    \text{s.t.} \quad & \sum_{a \in \calA^+_{N_p^0}} y_{pa} = 1 \quad \forall p \in \calP  \label{route_pinit} \\
    & \sum_{a \in \calA^-_{n}} y_{pa} - \sum_{a \in \calA^+_{n}} y_{pa} = 0 \quad \forall p \in \calP, n \in \calN \setminus \{N_p^0 \}  \label{route_pflow} \\
    & \sum_{a \in \calA^+_{N(w(m), t)}} \sum_{t=o(m)}^{T} y_{pa} \geq 1 \quad \forall m \in \calM^\prime, p \in \calP_m  \label{route_assign} \\
    & \sum_{a \in \calA} \sum_{p \in \calP} q^\prime_{a \ell t} y_{pa} \leq Q_\ell  \quad \forall \ell \in \calJ, t \in \calT  \label{route_traff} \\
    & y_{pa} \in \{0,1\} \quad \forall p \in \calP, a \in \calA \nonumber
\end{align}

\section{Extensions and additional analyses}
\label{sec:extensions}

\subsection{Choice of time discretization in time-space networks}
\label{subsec:discretization}

In our model, a coarse discretization renders the congestion proxy less accurate and the constraint on the maximum orders open more restrictive. On the other hand, the flexibility of order-item-pod pick scheduling at the workstation becomes more restrictive with a more granular discretization; indeed, the TDS-CW allows multiple picks from the same pod if all items can be scheduled within the same time period, so a granular discretization artificially restricts the number and combinations of items that can be picked.

To investigate the impact of time discretization on the model's computational complexity and the quality of its solution, we train five machine learning models with five different discretizations ($\Delta=20,30,50,60,120$ seconds), and solve 10 new instances with each discretization. Results are reported in Table~\ref{tab:timedisc}. For these experiments, we use a 20 minute planning horizon (instead of 15) so that the horizon is evenly divisible by all discretizations.

\begin{table} %%
\caption{Impact of time discretization (medium instance, tight congestion, 20-minute horizon).}
\label{tab:timedisc}
\centering\footnotesize\renewcommand{\arraystretch}{1}
\begin{tabular}{lccccccc} \toprule
 &  &  \multicolumn{2}{c}{Pick scheduling} & \multicolumn{3}{c}{Orders} & Congestion \\ \cmidrule(lr){3-4}\cmidrule(lr){5-7}\cmidrule(lr){8-8}
 & Objective & & Items picked & Orders & Single & Avg open time & Distance \\
$\Delta$ & (throughput) & Util \% & per pod &  completed &  item \% & per item (s) & per pick \\
 \midrule
20 & 341 & 75\% & 1.65 & 179 & 55\% & 33 & 51.1 \\
\textbf{30} & \textbf{396} & \textbf{83\%} & \textbf{1.82} & \textbf{197} & \textbf{53\%} & \textbf{38} & \textbf{41.2} \\
50 & 382 & 75\% & 2.15 & 196 & 56\% & 48 & 38.0 \\
60 & 360 & 68\% & 2.33 & 190 & 56\% & 56 & 33.1 \\
120 & 251 & 44\% & 2.60 & 156 & 65\% & 98 & 36.6 \\
\bottomrule
\end{tabular}
%\begin{tablenotes}\linespread{1}
%\item 
%\end{tablenotes}
\end{table}

As expected, coarser discretizations achieve more picks per pod given the additional flexibility of longer time periods. However, when $\Delta > 30$ seconds, the total throughput decreases in spite of improved pod efficiency. This is because the restriction on open orders (applied over each time period) prevents the solution from achieving high utilization. This is evidenced by longer open times for orders and an increasing reliance on single-item orders (which do not count towards the order capacity). The results suggest that $\Delta=30$ strikes a good balance, with the best throughput.

\subsection{Calibration of parameters in the learn-then-optimize algorithm}
\label{subsec:subproblem}

Our learn-then-optimize approach relies on several key parameters that govern the size of the subproblem at each iteration. In general, the subproblem should be as large as possible to maximize improvements while retaining tractability (in our case, solving in a few seconds). To illustrate how we selected the subproblem size, we report subproblem solution times for many values of the size parameters. The size of the subproblem can be described by the number of workstation-time blocks ($N_M \times N_T$) and the number of pods ($N_P$) and orders ($N_M$) available for assignment. For each number of blocks, we consider ``low variety'' and ``high variety'' settings to vary $N_M$ and $N_P$. Table~\ref{tab:spsize} displays the results for subproblems ranging from 8 to 48 blocks, using randomized order and pod selection for 30 LSNS iterations. 

\begin{table} %%
\caption{Solve time and solution quality for various subproblem sizes, with the two selected for our learn-then-optimize experiments in \textbf{bold}.}
\label{tab:spsize}
\centering\footnotesize\renewcommand{\arraystretch}{1}
\begin{tabular}{ccclccccc} \toprule
% &  & \multicolumn{4}{c}{Tight Congestion} & \multicolumn{4}{c}{Moderate Congestion}  & \multicolumn{4}{c}{No Congestion}  \\
% \cmidrule(lr){3-6}\cmidrule(lr){7-10}\cmidrule(lr){11-14} 
%Instance & Method & \multicolumn{2}{c}{Objective} & Util & CPU & \multicolumn{2}{c}{Objective} & Util & CPU & \multicolumn{2}{c}{Objective} & Util & CPU \\
\multicolumn{4}{c}{Subproblem size} & \multicolumn{3}{c}{CPU per subproblem \emph{(s)}} & \multicolumn{2}{c}{Solution} \\ \cmidrule(lr){1-4}\cmidrule(lr){5-7}\cmidrule(lr){8-9} 
Time window & Stations & Blocks & Orders and Pods & Total & Solve & Build & Util \% & Picks per pod \\
\midrule
4 min. & 1 & 8 & Low variety & 1.05 & 1.03 & 0.03 & 33\% & 1.15  \\
 &  &  & High variety & 1.55 & 1.50 & 0.06 & 41\% & 1.25 \\
 \midrule
\textbf{6 min.} & \textbf{1} & \textbf{12} & \textbf{Low variety} & \textbf{2.04} & \textbf{1.96} & \textbf{0.09} & \textbf{41\%} & \textbf{1.21} \\
 &  &  & High variety & 3.48 & 3.30 & 0.18 & 42\% & 1.21 \\
 \cmidrule(lr){2-9}
 & \textbf{2} & \textbf{24} & \textbf{Low variety} & \textbf{8.60} & \textbf{2.21} & \textbf{6.38} & \textbf{79\%} & \textbf{1.37} \\
 &  &  & High variety & 52.14 & 3.69 & 48.45 & 83\% & 1.62 \\
  \cmidrule(lr){2-9}
 & 3 & 36 & Low variety & 20.54 & 3.18 & 17.37 & 86\% & 1.51 \\
 &  &  & High variety & 60.47 & 4.35 & 56.13 & 86\% & 1.51 \\
 \midrule
8 min. & 2 & 32 & Low variety & 34.74 & 4.45 & 30.3 & 80\% & 1.58 \\
 &  &  & High variety & 145.52 & 5.96 & 139.56 & 90\% & 1.76 \\
  \cmidrule(lr){2-9}
 & 3 & 48 & Low variety & 107.62 & 6.09 & 101.52 & 90\% & 1.66 \\
 &  &  & High variety & 140.65 & 8.41 & 132.25 & 92\% & 1.77 \\
\bottomrule
\end{tabular}
\begin{tablenotes}\linespread{1}
\item All solutions obtained via randomized LSNS for a pure comparison.
\item Subproblem solution times averaged over 30 LSNS iterations.
\end{tablenotes}
\end{table}

As expected, solution quality improves as the size of the subproblem increases. For the largest subproblems with 48 workstation-time blocks, the solutions achieve 90-92\% utilization and 1.66-1.77 items picked per pod after 30 LSNS iterations, even using randomized subproblem selection. This is because each subproblem is roughly 25\% of the size of the full instance. This comes with an obvious cost in terms of computational time, as each LSNS iteration takes up to two minutes. To keep the solution time down to a few seconds, we selected subproblems with 12-24 blocks for our LSNS algorithms. Furthermore, we find that at this size, increasing the order/pod variety increases solution times but hardly improves solution quality. For that reason, we elect to keep the variety low ($N_M=20, N_P=28$) and rely on the learned subproblem selection problem to choose only highly promising pods and orders. These results underscore the benefits of our learn-then-optimize approach to focus on the most promising pods and orders while retaining small subproblems at each iteration.

\subsection{Flexible pod storage locations}
\label{subsec:flexible}

The model formulated in the paper forces pods to return to their original storage location upon visiting a workstation. In practice, Amazon can allow pods to be returned to any storage location from time to time. In fact, our model readily extends to reflect different policies around pod storage, by defining time-space arcs accordingly in the time-space network. Figure~\ref{fig:tsnstoragelocs} shows time-space networks for two pod storage policies: one where pods must return to their original storage location (``fixed'' policy, Figure~\ref{subfig:tsn_onlystorage}), and one where pods can return to any location with available storage capacity (``flexible'' policy, Figure~\ref{subfig:tsn_anystorage}). In the former case, the model comprises $\calO(\calW \times \calT)$ arcs for each pod, with the only possible movements being between the pods home location and a workstation. In the latter case, pods start from a single location, but are then unrestricted, leading to $\calO(\calS \times \calW \times \calT)$ arcs. This lends additional flexibility in terms of congestion management, but comes at the obvious price of a larger decision space. In addition, the latter policy necessitates a new constraint on the number of pods stored at each storage location at any time:
\begin{align}
    \sum_{p \in \calP} y_{p,A(N(s,t),N(s,t+1))} \leq D \quad \forall s\in\calS, t\in 1,\ldots,T-1
\end{align}
where $D$ be the maximum number of pods per location ($D=24$ in our setup). This constraint ensures storage space availability by enforcing that the number of pods idling at each location never exceeds the capacity. 

\begin{figure} %%
    \centering
    \subfloat[Pod must return to original storage location]{\label{subfig:tsn_onlystorage}
            \includegraphics[width=7.8cm]{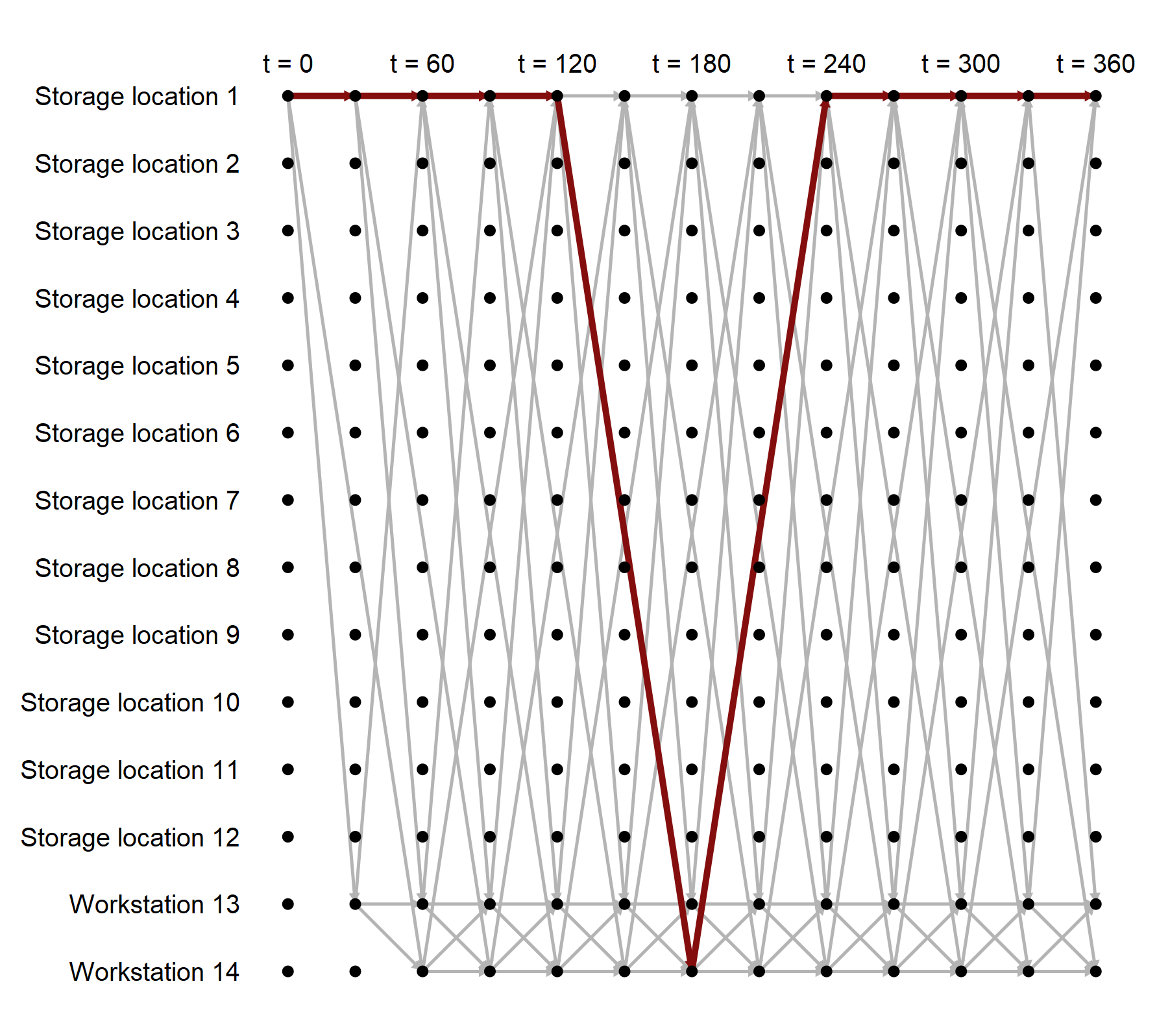}
        }\hspace{0.2 cm}
    \subfloat[Pod can be returned to any storage location]{\label{subfig:tsn_anystorage}
            \includegraphics[width=7.8cm]{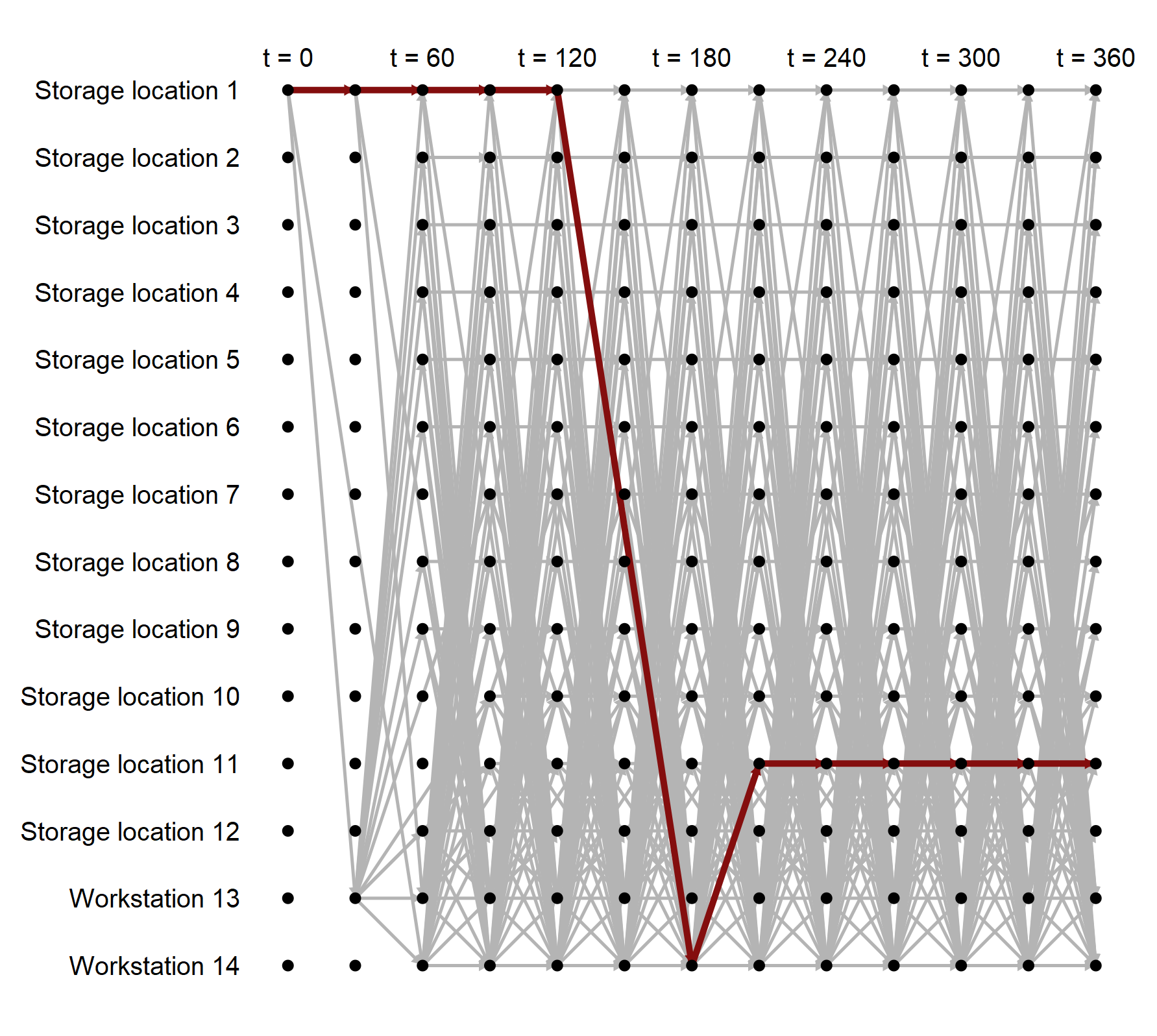}
        }
    \caption{Time-space network arcs (gray) and selected path (red) for one pod stored in location 1, under the fixed and flexible storage policies.}
    \label{fig:tsnstoragelocs}
%\vspace{-12pt}
\end{figure}

To determine the practical and computational impact of storage flexibility, we conducted experiments under both policies on small and medium instances under tight congestion. Under the flexible policy, we imposed two different time limits for re-optimization of the subproblem: 5 seconds (to be competitive with the fixed policy) and 30 seconds (to investigate the full benefits of the added flexibility). Results are averaged over 10 instances, and reported in Table~\ref{tab:anystorageloc}.

\begin{table} %%
\caption{Solution comparison with fixed vs. flexible location policies.}
\label{tab:anystorageloc}
\centering\footnotesize\renewcommand{\arraystretch}{1}
\begin{tabular}{llcccccc} \toprule
\midrule
  & Return & Subproblem & Avg total & Objective & Util & Items picked & Avg distance\\
 Instance &  location & time limit \scriptsize{(s)} & CPU \scriptsize{(s)} & (throughput) &  \% & per pod & per pick (m)\\
  \midrule
 Small & Fixed & 5 & 366 & 199 & 75\% & 2.42 & 22.1 \\
 \cmidrule(lr){2-8}
 & Flexible & 5 & 483 & 163 & 62\% & 2.38 & 23.9 \\
 &  & 30 & 601 & 835 & 77\% & 2.43 & 19.9 \\
   \midrule
Medium & Fixed & 5 & 502 & 328 & 90\% & 1.91 & 37.2 \\
  \cmidrule(lr){2-8}
 & Flexible & 5 & 536 & 272 & 74\% & 1.89 & 49.5 \\
 &  & 30 & 1,344 & 342 & 93\% & 1.94 & 35.2 \\
\bottomrule
\end{tabular}
%\begin{tablenotes}\linespread{1}
%\item
%\end{tablenotes}
\end{table}

We observe that with a subproblem time limit of 30 seconds, the flexibility of storage location leads to minor improvements in terms of throughput (2-4\% increase) and congestion management (5-10\% shorter trips). However, this results in significantly longer computational times, increasing from 8 minutes to 22 minutes for medium instances. When the subproblem time limit is set to 5 seconds per LSNS iteration, the computational times are competitive but there is a considerable drop in solution quality with flexible storage locations. In fact, the time-limited algorithm for this policy leads to 17-18\% \emph{lower} throughput compared to the fixed policy. The results underscore that adding storage location flexibility can lead to efficiency gains, although the added computational requirements can outweigh the benefits of flexibility.

\subsection{Impact of multi-stop trips} 
\label{subsec:multistop}

The TDS-CW allows multi-stop pod trips, where a pod departs from its storage location, visits multiple workstations, before returning to its original location. A question remains on the impact of such multi-stop trips on the efficiency of operations. Figure~\ref{fig:stationtostationarcs} compares the pod time-space networks when pods are allowed to visit one workstation at a time (Figure~\ref{subfig:singlestoptrips}) or to visit multiple workstations prior to returning to its storage location (Figure~\ref{subfig:multistoptrip}). Such flexibility can reduce travel times and alleviate congestion in the central part of the warehouse, at the cost of extra congestion around the workstations. Note that the number of arcs added to each time-space network is $\calO(\calW^2 \cdot \calT)$, compared to $\calO(\calW \cdot \calT)$ existing arcs for each pod. This change does not hinder scalability as long as the number of workstations remains small (e.g. $\calW = 3$ in our largest sized instance).
 
\begin{figure} %%
    \centering
    \subfloat[Pod path using only single stop trips]{\label{subfig:singlestoptrips}
            \includegraphics[width=7.8cm]{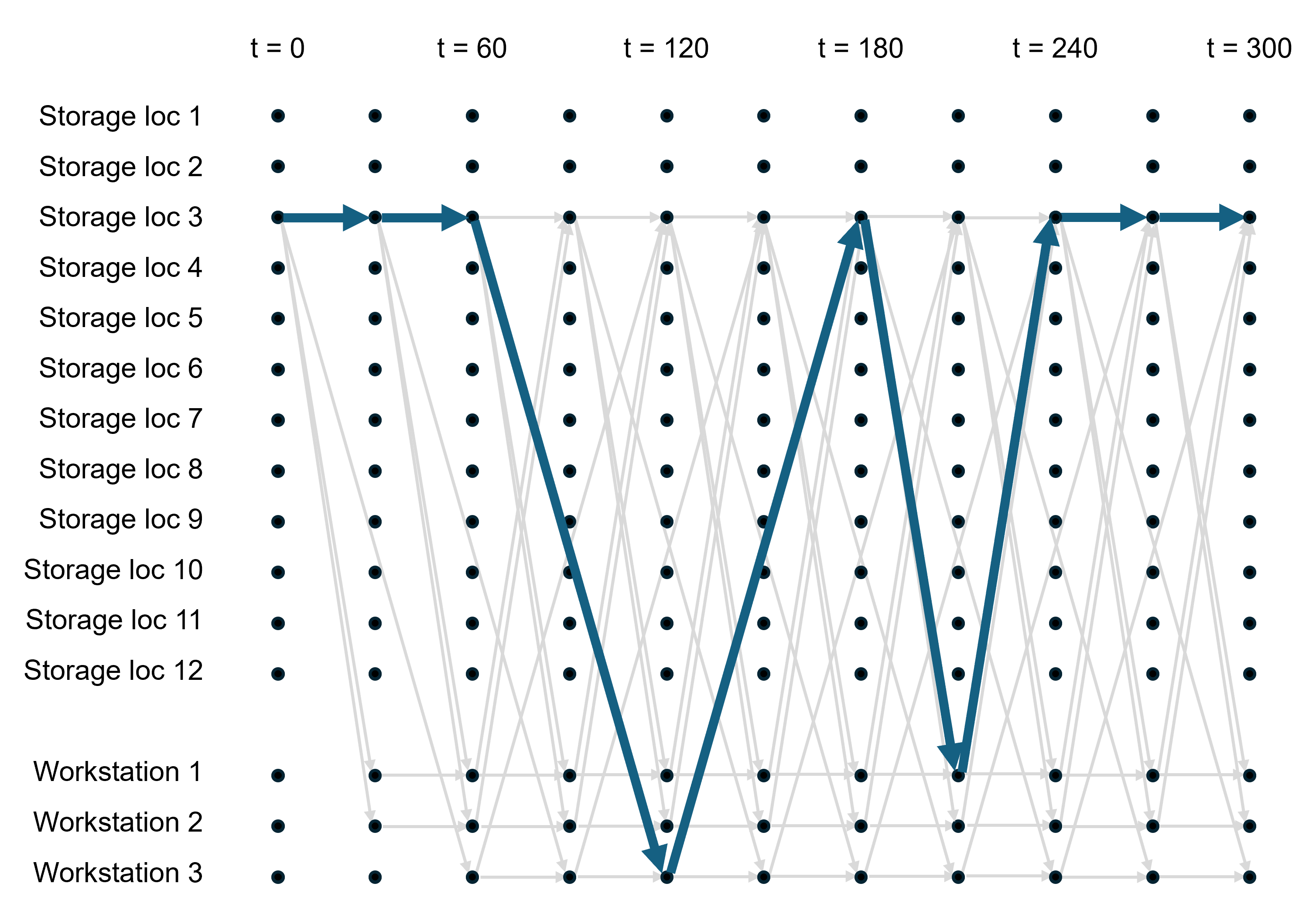}
        }\hspace{0.2 cm}
    \subfloat[Pod path for a multi-stop trip]{\label{subfig:multistoptrip}
            \includegraphics[width=7.8cm]{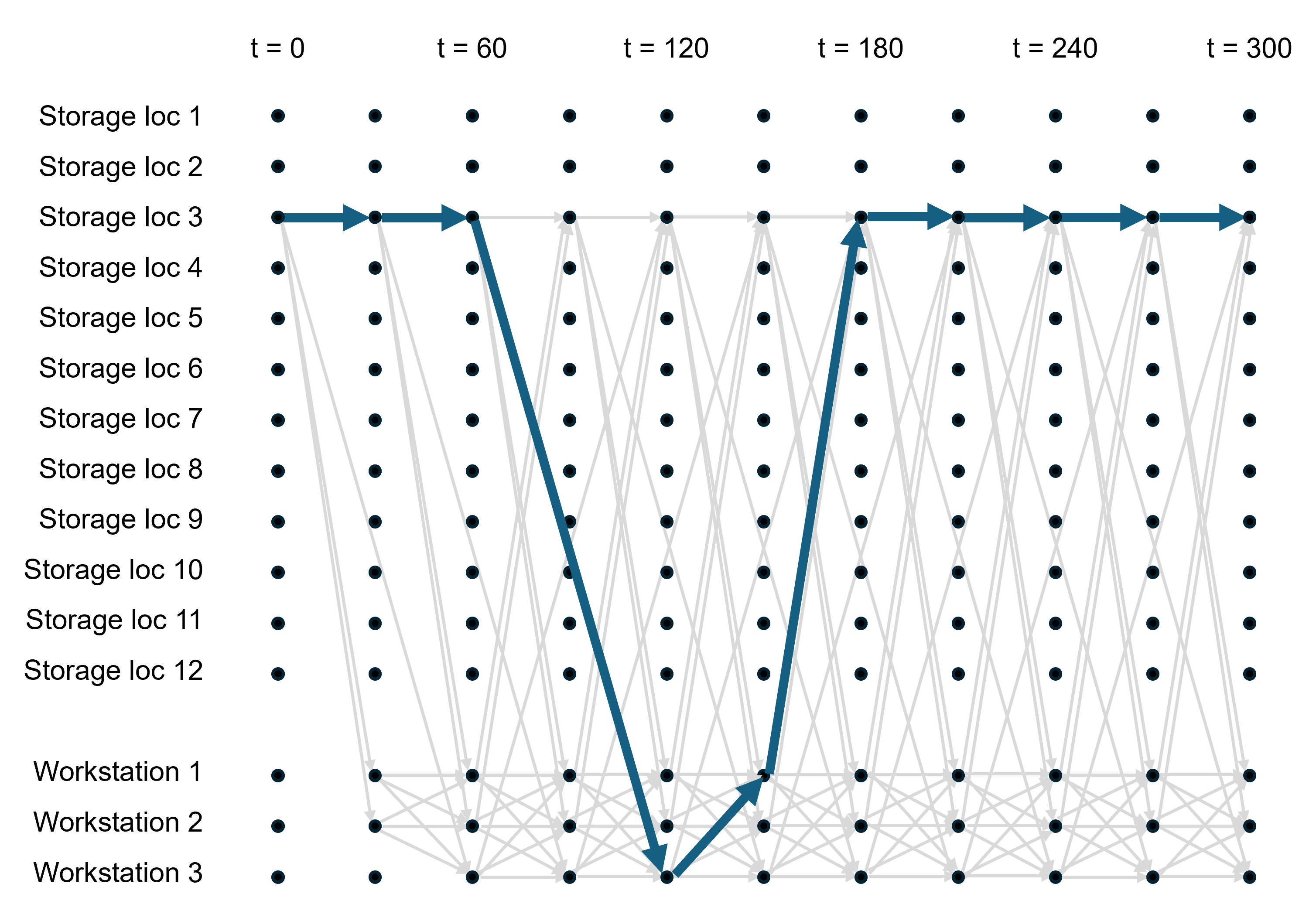}
        }
    \caption{A comparison of time-space network arcs (gray) and a selected path (blue) for one pod stored at location 3 under a single-stop trip and a multi-stop trip regime.}
    \label{fig:stationtostationarcs}
%\vspace{-12pt}
\end{figure}

We compare the impact of multi-stop trips in Table~\ref{tab:multistop}. On the one hand, allowing multi-stop trips improves throughput, by up to 6\% in medium-sized instances, due to faster travel between workstations and less congestion. On the other hand, the average number of stops per trip in the multi-stop trip regime is only 1.08-1.09, indicating that the solution predominantly relies on single-stop trips. This result stems from the fact that the most efficient use of any pod (to avoid congestion and increase throughput) is to pick all useful items at a single workstation via a single-stop trip. Thus, while multi-stop trips are an indicator of good \emph{trip efficiency} (a multi-stop trip generates less congestion than multiple single-stop trips), they also represent poor \emph{stop efficiency} (by requiring multiple stops to pick all requested items). Stop efficiency is captured by our ``items picked per pod'' metric (or ``picks per stop''). In other words, multi-stop trips are useful for improving travel patterns when a pod is needed in two or more workstations, but the main driver of throughput improvement lies in seeking synergistic order-workstation and item-pod assignments to allow more picks per stop, only requiring multi-stop trips when order-workstation re-assignment is not possible. These results further reinforce the insights generated in the paper on the importance of coordinating warehousing operations to allow more picks per pod. (Figure~\ref{fig:podpicks})

\begin{table} %%
\caption{Comparison of formulations with and without multi-stop trips.}
\label{tab:multistop}
\centering\footnotesize\renewcommand{\arraystretch}{1}
\begin{tabular}{llcccccc} \toprule
% &  & \multicolumn{4}{c}{Tight Congestion} & \multicolumn{4}{c}{Moderate Congestion}  & \multicolumn{4}{c}{No Congestion}  \\
% \cmidrule(lr){3-6}\cmidrule(lr){7-10}\cmidrule(lr){11-14} 
%Instance & Method & \multicolumn{2}{c}{Objective} & Util & CPU & \multicolumn{2}{c}{Objective} & Util & CPU & \multicolumn{2}{c}{Objective} & Util & CPU \\
 &  &  &  & \multicolumn{2}{c}{Percent of pods making...}  & \multicolumn{2}{c}{Item picks} \\ \cmidrule(lr){5-6}\cmidrule(lr){7-8} 
Instance & Multi-stop trips? & Throughput & Stops per trip & Multiple stops & Multiple trips & Per stop & Per trip \\
\midrule
Medium & No & 269 & 1.00 & 3.4\% & 3.4\% & 1.74 & 1.74 \\
 & Yes & 286 & 1.09 & 4.0\% & 2.6\% & 1.76 & 1.93 \\
 \midrule
Full & No & 2,300 & 1.00 & 3.2\% & 3.2\% & 1.94 & 1.94 \\
 & Yes & 2,306 & 1.08 & 3.6\% & 2.7\% & 1.91 & 2.05 \\
\bottomrule
\end{tabular}
%\begin{tablenotes}\linespread{1}
%\item 
%\end{tablenotes}
\end{table}

\end{APPENDICES}
\end{document}

%% file: abbrev.tex
\def\R{\mathbb{R}}

\def\E{\mathbb{E}}

\def\calA{\mathcal{A}}
\def\calB{\mathcal{B}}
\def\calC{\mathcal{C}}

\def\calE{\mathcal{E}}
\def\calF{\mathcal{F}}
\def\calG{\mathcal{G}}
\def\calH{\mathcal{H}}
\def\calI{\mathcal{I}}
\def\calJ{\mathcal{J}}

\def\calL{\mathcal{L}}
\def\calM{\mathcal{M}}
\def\calN{\mathcal{N}}
\def\calO{\mathcal{O}}
\def\calP{\mathcal{P}}

\def\calR{\mathcal{R}}
\def\calS{\mathcal{S}}
\def\calT{\mathcal{T}}

\def\calW{\mathcal{W}}

\def\calZ{\mathcal{Z}}

\def\bff{\boldsymbol{f}}
\def\bg{\boldsymbol{g}}

\def\bi{\boldsymbol{i}}

\def\bq{\boldsymbol{q}}
\def\br{\boldsymbol{r}}

\def\bv{\boldsymbol{v}}
\def\bw{\boldsymbol{w}}
\def\bx{\boldsymbol{x}}
\def\by{\boldsymbol{y}}
\def\bz{\boldsymbol{z}}

\def\st{\text{s.t.}}

\definecolor{mygreen}    {RGB}{0,90,0}
\definecolor{myblue}     {RGB}{0,51,140}
\definecolor{myorange}   {RGB}{238,118,0}
\definecolor{myred}      {RGB}{126,0,0}
\definecolor{mygray}     {RGB}{100,100,105}
\definecolor{mygrayblue} {RGB}{0,128,128}
\definecolor{mygraygreen}{RGB}{128,128,0}
\definecolor{DarkPurple}     {RGB}{142, 36, 170}
\definecolor{LightPurple}    {RGB}{57, 130, 7}